\def\year{2021}\relax
\documentclass[letterpaper]{article} 
\usepackage{aaai21}  
\usepackage{times}  
\usepackage{helvet} 
\usepackage{courier}  
\usepackage[hyphens]{url}  
\usepackage{graphicx} 
\usepackage{graphicx}  
\usepackage{natbib}  
\usepackage{caption} 

\usepackage{amsfonts}       
\usepackage{amsmath}
\usepackage{comment}
\usepackage{subfigure}
\usepackage{booktabs}       
\usepackage{algorithm}
\usepackage{algpseudocode}
\usepackage{color}
\usepackage{hyperref}       
\usepackage{todonotes}
\usepackage{wrapfig}

\usepackage[symbol]{footmisc}

\newtheorem{myLemma}{Lemma}
\newtheorem{myDefinition}{Definition}

\title{Foresee then Evaluate: Decomposing Value Estimation \\ with Latent Future Prediction}

\author{

    Hongyao Tang\textsuperscript{\rm 1}\thanks{Work partially done as an intern at Noah's Ark Lab, Huawei.},
    Jianye Hao\textsuperscript{\rm 1,2}\thanks{Corresponding author.},
    Guangyong Chen\textsuperscript{\rm 3}, 
    Pengfei Chen\textsuperscript{\rm 4}, 
    Chen Chen\textsuperscript{\rm 2}, \\ 
    Yaodong Yang\textsuperscript{\rm 2}, 
    Luo Zhang\textsuperscript{\rm 1}, 
    Wulong Liu\textsuperscript{\rm 2}, 
    Zhaopeng Meng\textsuperscript{\rm 1}
}
\affiliations{

    \textsuperscript{\rm 1}College of Intelligence and Computing, Tianjin University,
    \textsuperscript{\rm 2}Noah's Ark Lab, Huawei \\
    \textsuperscript{\rm 3}Shenzhen Institutes of Advanced Technology, Chinese Academy of Sciences, 
    \textsuperscript{\rm 4}The Chinese University of Hong Kong

    \{bluecontra,luozhang,mengzp\}@tju.edu.cn, 
    gy.chen@siat.ac.cn,
    pfchen@cse.cuhk.edu.hk,
    \{haojianye,chenchen9,yang.yaodong,liuwulong\}@huawei.com

}

\begin{document}
\linespread{0.97}
\maketitle

\begin{abstract}
  Value function is the central notion of Reinforcement Learning (RL).
  Value estimation, especially with function approximation, can be challenging since it involves the stochasticity of environmental dynamics and reward signals that can be sparse and delayed in some cases.
  A typical model-free RL algorithm usually estimates the values of a policy by Temporal Difference (TD) or Monte Carlo (MC) algorithms directly from rewards, without explicitly taking dynamics into consideration. 
  In this paper, we propose Value Decomposition with Future Prediction (VDFP), providing an explicit two-step understanding of the value estimation process: 1) first foresee the latent future, 2) and then evaluate it.
  We analytically decompose the value function into a latent future dynamics part and a policy-independent trajectory return part,
  inducing a way to model latent dynamics and returns separately in value estimation.
  Further, we derive a practical deep RL algorithm, consisting of a convolutional model to learn compact trajectory representation from past experiences, a conditional variational auto-encoder to predict the latent future dynamics and a convex return model that evaluates trajectory representation.
  In experiments, 
  we empirically demonstrate the effectiveness of our approach for both off-policy and on-policy RL in several OpenAI Gym continuous control tasks as well as a few challenging variants with delayed reward.
\end{abstract}

\section{Introduction}
Reinforcement learning (RL) is a promising approach to obtain the optimal policy in sequential decision-making problems.
One of the most appealing characteristics of RL is that policy can be learned in a model-free fashion, without the access to environment models.
Value functions play an important role in model-free RL \cite{Sutton1988ReinforcementLA}, which are usually used to derive a policy implicitly in value-based methods \cite{Mnih2015DQN} or guide the policy updates in policy-based methods \cite{Schulman2015TRPO,Silver2014DPG}.
With deep neural networks, value functions can be well approximated even for continuous state and action space, making it practical for model-free RL to deal with more challenging tasks \cite{Lillicrap2015DDPG,Mnih2015DQN,Silver2016Go,vinyals2019grandmaster,HafnerLB020Dream,schreck2019retrosyn}.

Value functions define the expected cumulative rewards (i.e., returns) of a policy,
indicating how a state or taking an action under a state could be beneficial when performing the policy.
A value function is usually estimated directly from rewards through Monte Carlo (MC) or Temporal Difference (TD) algorithms \cite{Sutton1988ReinforcementLA},
without explicitly dealing with the entanglement of reward signals and environmental dynamics.
However, learning with such entanglement can be challenging in practical problems 
due to the complex interplay between highly stochastic environmental dynamics and noisy and even delayed rewards.
Besides, it is apparent that the dynamics information is not well considered and utilized 
during the learning process 
of value functions. 
Intuitively, human beings usually not only learn the reward feedback of their behaviors but also understand the dynamics of the environment impacted by their polices.
To evaluate a policy, it can commonly be the case of the following two steps: first \emph{foresee} how the environment could change afterwards, and then \emph{evaluate} how beneficial the dynamics could be.
Similar ideas called \emph{Prospective Brain} are also studied in cognitive behavior and neuroscience \cite{Atance2001EpisodicFT,Schacter2007RememberingTP,Schacter2007TheCN}.
Therefore, we argue that it can be important to disentangle the dynamics and returns of value function for better value estimation and then ensure effective policy improvement.

Following the above inspiration, in this paper
we consider that value estimation can be explicitly conducted in a two-step way:
1) an agent first predicts the following environmental dynamics regarding a specific policy in a latent representation space, 
2) then it evaluates the value of the predicted latent future.
Accordingly, 
we look into the value function and 
re-write it in a composite form of:
1) a reward-independent predictive dynamics function, which defines the expected representation of future state-action trajectory;
and 2) a policy-independent trajectory return function that maps any trajectory (representation) to its discounted cumulative reward.
This provides a new decomposed view of value function and induces a trajectory-based way to disentangle the dynamics and returns accordingly, namely Value Decomposition with Future Prediction (\textbf{VDFP}).
VDFP allows the decoupled two parts to be modeled separately and may alleviate the complexity of taking them as a whole. 
Further, we propose a practical implementation of VDFP,
consisting of a convolutional model to learn latent trajectory representation, a conditional variational dynamics model to predict and a convex trajectory return model to evaluate.
At last, we derive practical algorithms for both off-policy and on-policy model-free Deep RL by replacing conventional value estimation with VDFP.

Key contributions of this work are summarized as follows.
\begin{itemize}
    \item 
    We propose an explicit decomposition of value function (i.e., VDFP),
    in a form of the composition between future dynamics prediction and trajectory return estimation.
    It allows value estimation to be performed in a decoupling fashion flexibly and effectively.
    \item 
    We propose a conditional Variational Auto-Encoder (VAE) \cite{Higgins2017Beta,Kingma2013AEVB} to model the underlying distribution of future trajectory and then use it for prediction in a latent representation space.
    \item
    Our algorithms derived from VDFP outperforms their counterparts with conventional value estimation for both off-policy and on-policy RL in continuous control tasks.
    Moreover, VDFP shows significant effectiveness and robustness under challenging delayed reward settings.
\end{itemize}
For reproducibility, 
we conduct experiments on commonly adopted OpenAI gym continuous control tasks \cite{Brockman2016Gym,Todorov2012MuJoCo} and perform ablation studies for each contribution. 
Source codes are available at \url{https://github.com/bluecontra/AAAI2021-VDFP}.

\section{Related Work}
Thinking about the future has been considered as an integral component of human cognition \cite{Atance2001EpisodicFT,Schacter2007TheCN}.
In neuroscience, the concept of the prospective brain
\cite{Schacter2007RememberingTP} indicates that a crucial function of the brain is to 
integrate information past experiences and to construct mental simulations about possible future events.
One related work in RL that adopts the idea of future prediction is \cite{Dosovitskiy2017DFP}, in which a supervised model is trained to predict the residuals of goal-related measurements at a set of temporal offsets in the future. 
With a manually designed goal vector, actions are chosen through maximizing the predicted outcomes.
Dynamics prediction models are also studied in model-based RL 
to learn the dynamics model of the environment for synthetic experience generation or planning \cite{Atkeson1997Robot,Sutton1991Dyna}.
SimPLe \cite{Kaiser2019MBRL} and Dreamer \cite{HafnerLB020Dream} are proposed to learn one-step predictive world models that are used to train a policy within the simulated environment. 
Multi-steps and long-term future are also modeled in \cite{Hafner2019Learning,Ke2019LearningD} with recurrent variational dynamics models,
after which actions are chosen through online planning, e.g., Model-Predictive Control (MPC). 
In contrast to seek for a world model, 
in this paper we care about imperfect latent future prediction underneath the model-free value estimation process,
which are closer to human nature in our opinion.

Most model-free deep RL algorithms approximate value functions with deep neural networks to generalize value estimates in large state and action space, e.g., DDPG \cite{Lillicrap2015DDPG}, Proximal Policy Optimization (PPO) \cite{Schulman2017PPO}, and Advantage Actor-Critic (A2C) \cite{Mnih2016AC}.
A well-approximated value function can induce effective policy improvement in Generalized Policy Iteration (GPI) \cite{Sutton1988ReinforcementLA}.
Value functions are usually learned from rewards through TD or MC algorithms,
without explicitly considering the dynamics.
In this paper, we propose a decomposed value estimation algorithm that explicitly models the underlying dynamics.

A similar idea that decompose the value estimation is
the Successor Representation (SR) for TD learning \cite{Dayan1993ImprovingGF}.
SR assumes that the immediate reward function is a dot-product of state representation and a weight vector, and then the value function can be factored into the expected representation of state occupancy by dividing the weight vector.
Thereafter, Deep Successor Representation (DSR) \cite{Kulkarni2016DSR} extends the idea of SR based on Deep Q-Network (DQN) \cite{Mnih2015DQN}. 
The SR function is updated with TD backups and the weight vector is approximated from states and immediate rewards in collected experiences.
The idea of SR is also further developed in transfer learning 
\cite{Barreto17SF,Barreto18SRTransfer}.
In this paper, we derive the composite function form of value function from the perspective of trajectory-based future prediction, and we show that SR can be viewed as a special linear case of our derivation.
In contrast to using TD to learn SR function, we use a conditional VAE to model the latent distribution of trajectory representation.
Besides, we focus on the trajectory return instead of immediate reward, which can be more effective and robust for practical reward settings (e.g., sparse and delayed rewards).
In a nutshell, our work differs from SR in both start points and concrete algorithms.

\section{Background}
Consider a Markov Decision Process (MDP) $\left< \mathcal{S}, \mathcal{A}, \mathcal{P}, \mathcal{R}, \rho_0, \gamma, T\right>$,
defined with a state set $\mathcal{S}$, an action set $\mathcal{A}$, transition function $\mathcal{P}: \mathcal{S} \times \mathcal{A} \times \mathcal{S} \rightarrow \mathbb{R}_{\in [0,1]}$, reward function $\mathcal{R}: \mathcal{S} \times \mathcal{A} \rightarrow \mathbb{R}$, initial state distribution $\rho_0: \mathcal{S} \rightarrow \mathbb{R}_{\in [0,1]}$, discounted factor $\gamma \in [0,1]$, and finite horizon $T$. 
The agent interacts with the MDP at discrete timesteps by performing its policy $\pi: \mathcal{S} \rightarrow \mathcal{A}$, 
generating a trajectory of states and actions,
$\tau_{0:T} = \left(s_0, a_0, \dots, s_T , a_T \right)$, where $s_0 \sim \rho_0(s_0)$, $a_t \sim \pi(s_t)$ and $s_{t+1} \sim \mathcal{P}(s_{t+1}|s_t, a_t)$.
An RL agent's objective is to 
maximize the expected discounted cumulative reward:
$J(\pi) = \mathbb{E} \left[\sum_{t=0}^{T}\gamma^{t} r_t| \pi \right]$ where $r_t = \mathcal{R}(s_t,a_t)$.

In RL, 
state-action value function $Q$ is defined as the expected cumulative discounted reward for selecting action $a$ in state $s$, then following a policy $\pi$ afterwards:
\begin{equation}
\label{eqation:1}
    Q^{\pi}(s,a) = \mathbb{E} \left[\sum_{t=0}^{T}\gamma^{t} r_t|s_0=s, a_0=a; \pi \right]. 
\end{equation}
State value function
$V^{\pi}(s) = \mathbb{E} \left[\sum_{t=0}^{T}\gamma^{t} r_t|s_0=s; \pi \right]$
defines the value of states under policy $\pi$ similarly.

For continuous control, a parameterized policy $\pi_{\theta}$, with parameters $\theta$, can be updated by taking the gradient of the objective $\nabla_{\theta} J(\pi_{\theta})$. 
In actor-critic methods, a deterministic policy (actor) can be updated with the deterministic policy gradient (DPG) theorem \cite{Silver2014DPG}:
\begin{equation}
\label{eqation:2}
   \nabla_{\theta} J(\pi_{\theta}) = \mathbb{E}_{s \sim \rho^{\pi}} \left[ \nabla_{\theta} \pi_{\theta}(s) \nabla_{a} Q^{\pi}(s,a)|_{a=\pi_{\theta}(s)}\right],
\end{equation}
where $\rho^{\pi}$ is the discounted state distribution under policy $\pi$. 
Deep Deterministic Policy Gradient (DDPG) \cite{Lillicrap2015DDPG} is one of the most representative RL algorithm for continuous policy learning.
The critic network ($Q$-function) is approximated with off-policy TD learning and the policy is updated through the Chain Rule in Equation \ref{eqation:2}.

\section{Value Decomposition of Future Prediction}
\label{section:model}

Value estimation faces the coupling of environmental dynamics and reward signals.
It can be challenging to obtain accurate value functions in complex problems with stochastic dynamics and sparse or delayed reward.
In this section, we look into value functions and propose a way to decompose the dynamics and returns from the perspective of latent future trajectory prediction.

We first consider a trajectory representation function $f$ that maps any trajectory to its latent representation, i.e.,  $m_{t:t+k} = f (\tau_{t:t+k})$ for $\tau_{t:t+k} = \left(s_t, a_t, ..., s_{t+k}, a_{t+k} \right)$ with $k \ge 0$.
We then define the trajectory return function $U$ and the predictive dynamics function $P$ as follows:


\begin{myDefinition}
\label{definition:U}
The trajectory return function $U$ defines the cumulative discounted reward of any trajectory $\tau_{t:t+k}$ with the representation $ m_{t:t+k} = f (\tau_{t:t+k})$:
\begin{equation}
\label{eqation:3}
   U \big( f(\tau_{t:t+k}) \big) = U ( m_{t:t+k} ) = \sum_{t^{\prime}=t}^{t+k}\gamma^{t^{\prime} - t} r_{t^{\prime}}.
\end{equation}
\end{myDefinition}
Since $U$ does not depend on a particular policy, it can be viewed as a partial model of the environment that evaluates the overall utility of a trajectory. 

\begin{myDefinition}
\label{definition:P}
Given the representation function $f$, the predictive dynamics function $P$ denotes the expected representation of the future trajectory for performing action $a \in \mathcal{A}$ in state $s \in \mathcal{S}$, then following a policy $\pi$:
\begin{equation}
\label{eqation:4}
\begin{aligned}
   P^{\pi}( s, a ) & = \mathbb{E} \big[ f(\tau_{0:T}) | s_0=s, a_0=a;\pi \big] \\ 
   & = \mathbb{E} \big[ m_{0:T} | s_0=s, a_0=a;\pi \big].
\end{aligned}
\end{equation}
\end{myDefinition}
Similar to the definition of $Q$-function, $P$ is associated to policy $\pi$, except for the expectation imposed on the trajectory representation.
It predicts how the states and actions would evolve afterwards, which is independent with reward.

With above definitions, we derive the following lemma:
\begin{myLemma}
\label{lemma:lower_bound}
Given a policy $\pi$, the following lower bound of the $Q$-function holds for all $s \in \mathcal{S}$ and $a \in \mathcal{A}$, when function $U$ is convex:
\begin{equation}
\label{equation:lower_bound}
    Q^{\pi}(s,a) \ge U \big(P^{\pi}(s,a) \big).
\end{equation}
The equality is strictly established when $U$ is linear.
\end{myLemma}
The proof can be obtained with \emph{Jensen's Inequality}
Complete proof can be found in Supplementary Material \ref{supp:lem1_proof}.
Similar conclusion can also be obtained for state value function $V$ and we focus on $Q$-function in the rest of the paper.

Lemma \ref{lemma:lower_bound} provides a lower-bound approximation of the $Q$-function as a composite function of $U$ and $P$.
When $U$ is a linear function, 
the equality guarantees that we can also obtain the optimal policy through optimizing the composite function.
Note that a linear $U$ actually does not limit the representation power of the whole composite function,
since the input of $U$, i.e., 
trajectory representation,
can be non-linear.
Moreover, we suggest that Successor Representation (SR) \cite{Dayan1993ImprovingGF,Kulkarni2016DSR} can be considered as a special linear case when $U$ is a weight vector and $P$ represents the expected discounted successor state occupancy. 
For the case that $U$ is a convex function (e.g., a single fully connected layer with ReLU activation), 
Lemma \ref{lemma:lower_bound} indicates a lower-bound optimization of the $Q$-function by maximizing the composite function. 
There is no theoretical guarantee for the optimality of the learned policy in such cases, although we also found comparable results for convex functions in our experiments (see Ablation \ref{section:ablation}).

The above modeling induces an understanding that the $Q$-function takes an explicit two-step estimation: 
1) it first predicts the expected future dynamics under the policy in terms of latent trajectory representation (i.e., $P$), 
2) then evaluates the utility of latent prediction (i.e., $U$).
This coincides our intuition that humans not only learn from rewards but also understand the dynamics.
This provides us a way to decompose the value estimation process by dealing with $P$ and $U$ separately, namely Value Decomposition of Future Prediction (\textbf{VDFP}).
The decoupling of environmental dynamics and returns reduces the complexity of value estimation and provides flexibility in training and using the decomposed two parts.
With a compact trajectory representation, prediction and evaluation of future dynamics can be efficiently carried out in a low-dimensional latent space.
Moreover, VDFP draws a connection between model-free RL and model-based RL since the composite function in Lemma \ref{lemma:lower_bound} indicates an evidence of learning partial or imperfect environment models during model-free value estimation.

\begin{figure*}[t]
\centering
\includegraphics[width=0.80\textwidth]{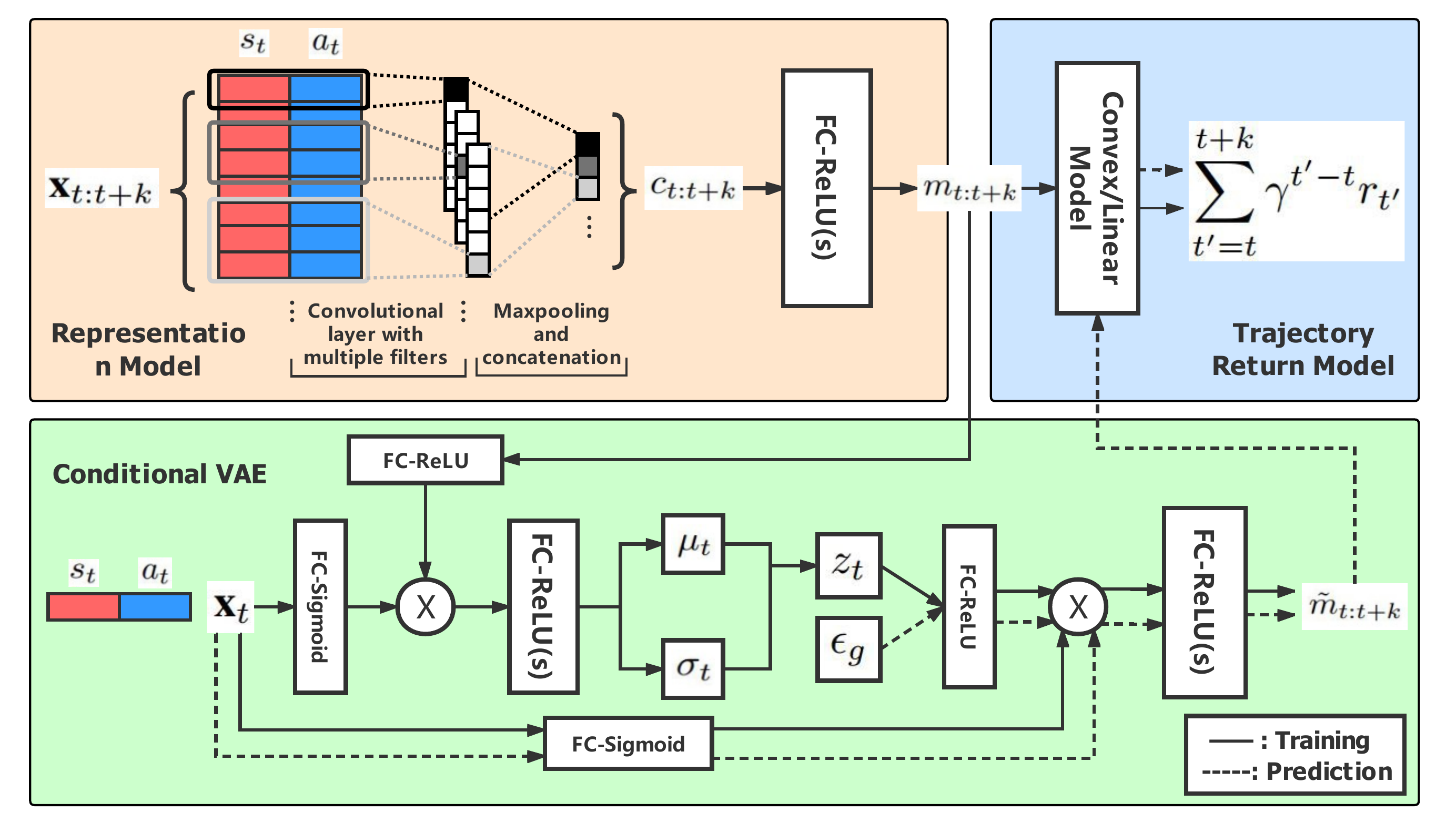}

\caption{The overall network structure of our models consists of: representation model (\emph{orange}), trajectory return model (\emph{blue}) and conditional VAE (\emph{green}).
We abbreviate the Fully-Connected layers as FC (with certain activation) and use $\otimes$ to denote the element-wise product operation.
The solid lines illustrate the flow of training process and the dashed lines illustrate the two-step prediction of decomposed value estimation 
with from the generative decoder ($P$) to the trajectory return model ($U$).
}
\label{figure:1}
\end{figure*}

Finally, when value estimation is conducted with the composite function approximation in Lemma \ref{lemma:lower_bound}, 
we can obtain the value-decomposed deterministic policy gradient $\nabla_{\theta} \tilde{J}(\pi_{\theta})$ by extending Equation \ref{eqation:2} with the Chain Rule:
\begin{equation}
\label{eqation:7}
\begin{aligned}
   \nabla_{\theta} \tilde{J}(\pi_{\theta}) = \mathbb{E}_{s \sim \rho^{\pi}} \big[ \nabla_{\theta} \pi_{\theta}(s) 
   & \nabla_{a} P^{\pi}(s,a)|_{a=\pi_{\theta}(s)} \\
   \cdot & \nabla_{m} U(m)|_{m=P^{\pi}(s,a)}
   \big].
\end{aligned}
\end{equation}

\section{Deep Reinforcement Learning with VDFP}
\label{section:approach}
In this section, 
we implement VDFP proposed in previous section with modern deep learning techniques and then derive a practical model-free Deep RL algorithm from it. 

\subsection{State-Action Trajectory Representation}
\label{section:representation}
The first thing to consider is the representation of state-action trajectory.
To derive a practical algorithm, an effective and compact representation function is necessary because: 1) the trajectories may have variant length, 2) and there may be irrelevant features in states and actions which might hinder the estimation of the cumulative discounted reward of the trajectory.
We propose using Convolutional Neural Networks (CNNs) to learn a representation model $f^{\rm CNN}$ of the state-action trajectory, 
similar to the use for sentence representation in \cite{Kim2014CNN}.
In our experiments, we found that this way achieves faster training and better performance than the popular sequential model LSTM \cite{Hochreiter1997LSTM} (see Ablations  \ref{section:ablation}).

An illustration of $f^{\rm CNN}$ is shown in the orange part of Figure \ref{figure:1}.
Let $x_t \in \mathbb{R}^l$ be the $l$-dimensional feature vector of a state-action pair $(s_t,a_t)$. 
A trajectory $\tau_{t:t+k}$ (padded if necessary) is represented as $\textbf{x}_{t:t+k} = x_t \oplus x_{t+1} \oplus \dots \oplus x_{t+k}$, where $\oplus$ is the concatenation operator.
A feature $c^i_{t:t+k}$ of trajectory $\tau_{t:t+k}$ can be generated
via a convolution operation which involves a 
filter $\textbf{w}^i \in \mathbb{R}^{h^i \times l}$ that applied to a window of $h^i$ state-action pairs, and then a max-pooling:
\begin{equation}
\label{eqation:8}
\begin{aligned}
    & c^i_{t:t+k} = \max \{c^i_t, c^i_{t+1} \dots, c^i_{t + k - h^i + 1}\}, \\
    & {\rm where} \ c^i_j = {\rm ReLU}(\textbf{w}^i \cdot \textbf{x}_{j:j+h^i-1} + b).
\end{aligned}
\end{equation}
We apply multiple filters ${\{\textbf{w}^i}\}_{i=1}^{n}$ on trajectory $\tau_{t:t+k}$ similarly to generate the $n$-dimensional feature vector 
$c_{t:t+k}$, 
then obtain the trajectory representation $m_{t:t+k}$ after feeding $c_{t:t+k}$ through a Multi-Layer Perceptron (MLP): 
\begin{equation}
\label{eqation:9}
\begin{aligned}
    c_{t:t+k} = & {\ \rm concat}( c^1_{t:t+k},  c^2_{t:t+k}, \dots, c^n_{t:t+k}), \\
    m_{t:t+k} = & f^{\rm CNN} (\tau_{t:t+k}) = {\rm MLP}(c_{t:t+k}) .
\end{aligned}
\end{equation}

Through sliding multiple convolutional filters of different scales along the trajectory, 
effective features can be extracted with max-pooling thereafter.
Intuitively, it is like to scan the trajectories and remember the most relevant parts.

\subsection{Trajectory Return Model}
\label{section:return}
With a compact trajectory representation, we consider the trajectory return function $U$ that maps a trajectory representation into its discounted cumulative rewards.
As discussed in Section \ref{section:model}, $Q$-function can be exactly estimated by the composite function when $U$ is linear without decreasing representation power.
Therefore,
we use a linear $U$, as shown in the blue part of Figure \ref{figure:1}, for following demonstration.

The representation model $f^{\rm CNN}$ and return model $U^{\rm Linear}$ can be trained together with respect to their parameters $\omega$, by minimizing the mean square error of trajectory returns with mini-batch samples from the experience buffer $\mathcal{D}$:
\begin{equation}
\label{eqation:10}
\begin{aligned}
    \mathcal{L}^{\rm Ret}(\omega) = \mathbb{E}_{\tau \sim \mathcal{D}} \Big[ \big( & U^{\rm Linear}(f^{\rm CNN}(\tau_{t:t+k})) - \sum_{t^{\prime}=t}^{t+k}\gamma^{t^{\prime} - t} r_{t^{\prime}} \big)^2 \Big].
\end{aligned}
\end{equation}

Generally, we can use any convex functions for the trajectory return function $U$. 
For the simplest case, one can use a single fully-connected layer with ReLU or Leaky-ReLU activation.
More sophisticated models like Input Convex Neural Networks (ICNNs) \cite{Amos17ICNN,Chen19OptimalControl} can also be considered. 
The results for using such convex models can be seen in Ablation (\ref{section:ablation}).

\subsection{Conditional Variational Dynamics Model}
\label{section:conditional-VAE}
One key component of VDFP is the predictive dynamics function $P$.
It is non-trivial to model the expected representation of future trajectory since it reflects the long-term interplay between stochastic environment transition and agent's policy.
The most straightforward way to implement the predictive dynamics function is to use a MLP $P^{\rm MLP}$ that takes the state and action as input and predicts the expected representation of future trajectory.
However, such a deterministic model performs poorly since it is not able to capture the stochasticity of possible future trajectories.
In this paper, we propose a conditional Variational Auto-Encoder (VAE) to capture the underlying distribution of future trajectory representation conditioned on the state and action,
achieving significant improvement over $P^{\rm MLP}$ in our experiments.
We also provide more insights and experiments on the choice of conditional VAE in Supplementary Material \ref{supp:VAE_discussion}.

The conditional VAE consists of two parts, 
an encoder network $q_{\phi}(z_t|m_{t:t+k}, s_t, a_t)$ and a decoder network $p_{\varphi}(m_{t:t+k}|z_t, s_t, a_t)$ with variational parameters $\phi$ and generative parameters $\varphi$ respectively.
With a chosen prior, generally using the multivariate Gaussian distribution $\mathcal{N}(0,I)$, the encoder approximates the conditional posteriors of latent variable $z_t$ from trajectory representation instances, producing a Gaussian distribution with mean $\mu_t$ and standard deviation $\sigma_t$.
The decoder takes a given latent variable as input and generates a representation of future trajectory $\tilde{m}_{t:t+k}$ conditioned on the state-action pair.
The structure of the conditional VAE is illustrated in the green part of Figure \ref{figure:1}.
Specially, we use an element-wise product operation to emphasize an explicit relation between the condition stream and trajectory representation stream.

\begin{figure*}[t]
\centering
\subfigure[LunarLander-v2]{
\includegraphics[width=0.24\textwidth]{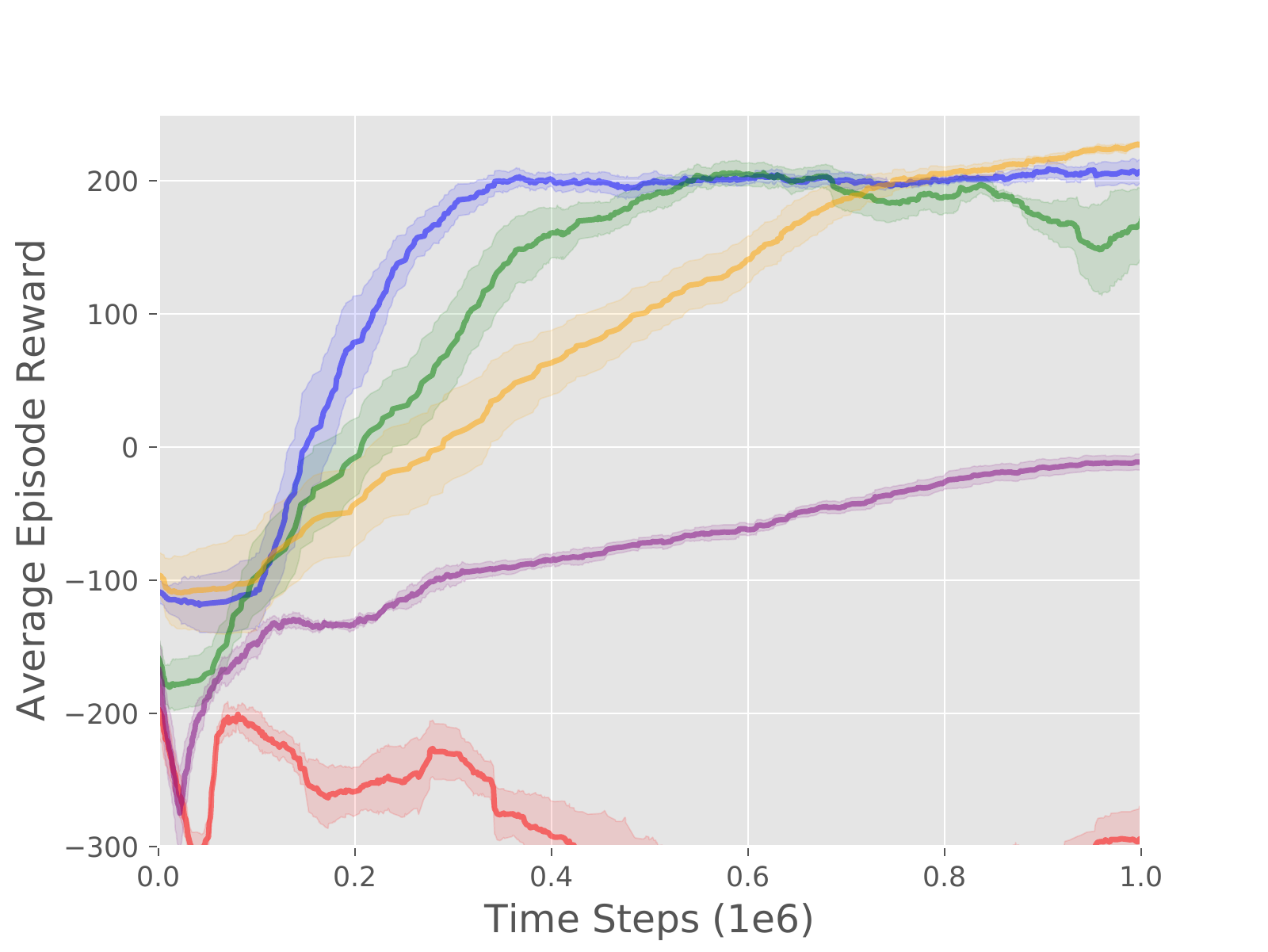}
}
\hspace{-0.35cm}
\subfigure[InvertedDoublePendulum-v1]{
\includegraphics[width=0.24\textwidth]{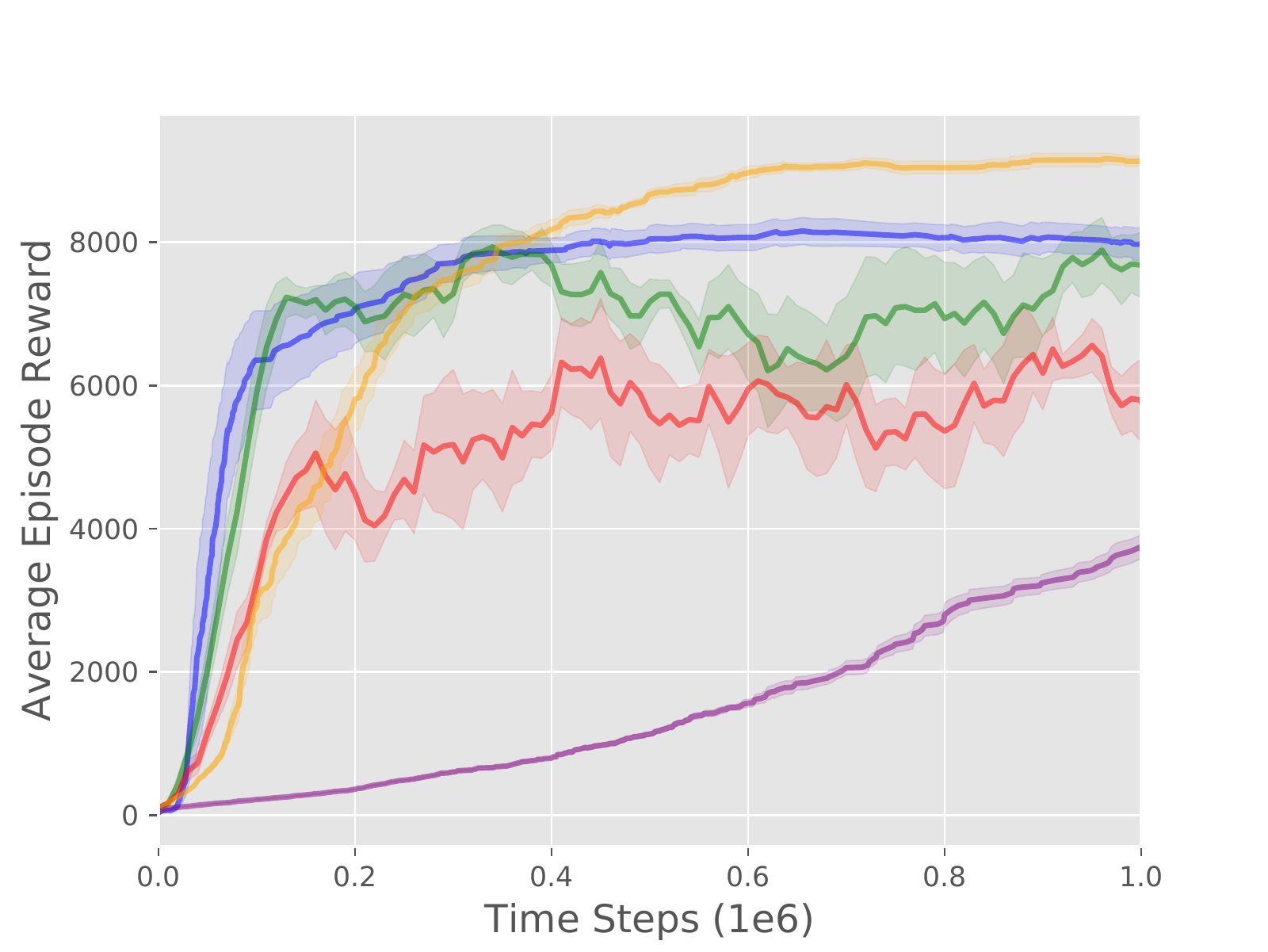}
}
\hspace{-0.35cm}
\subfigure[HalfCheetah-v1]{
\includegraphics[width=0.24\textwidth]{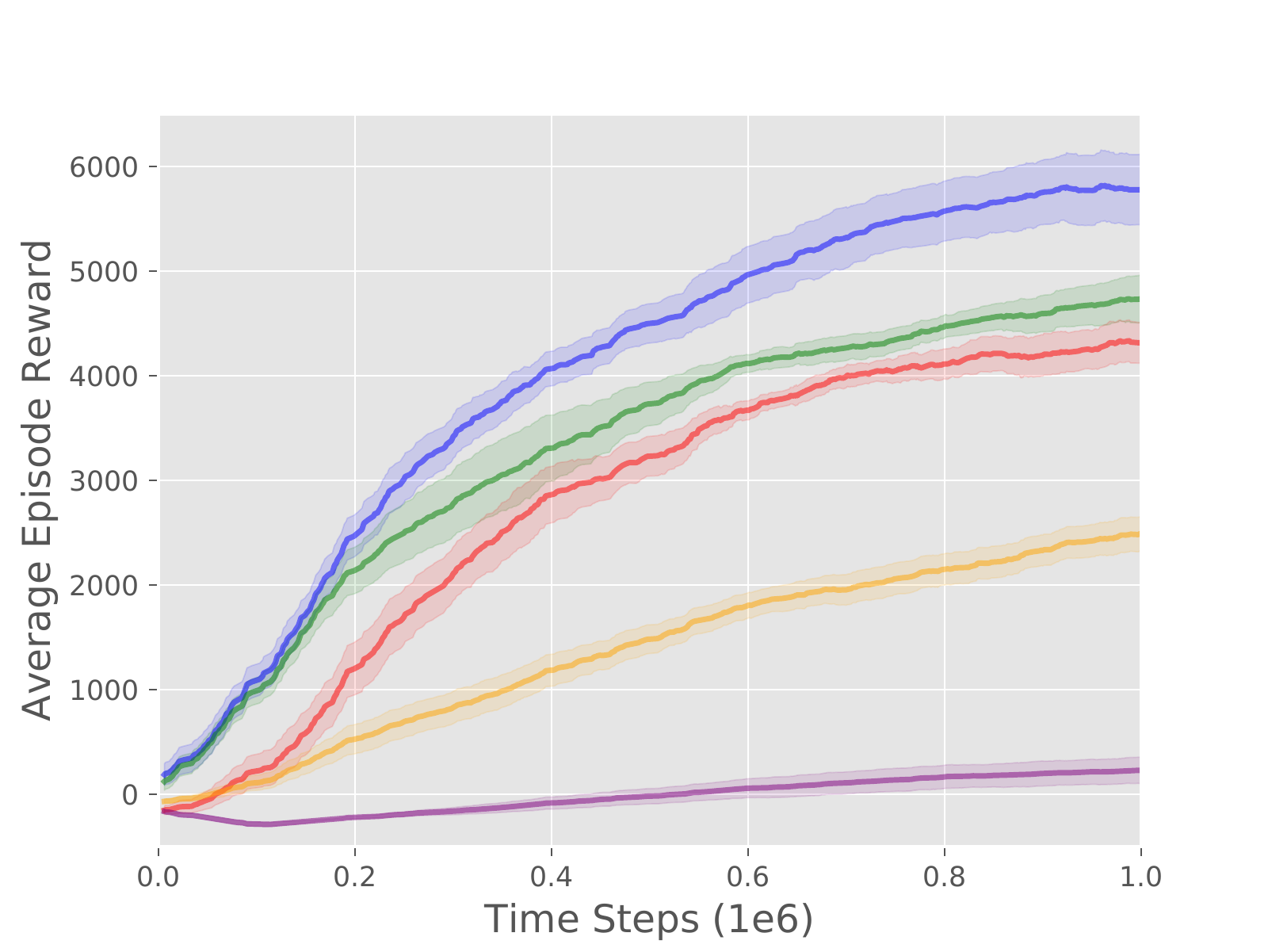}
}
\hspace{-0.35cm}
\subfigure[Walker2d-v1]{
\includegraphics[width=0.24\textwidth]{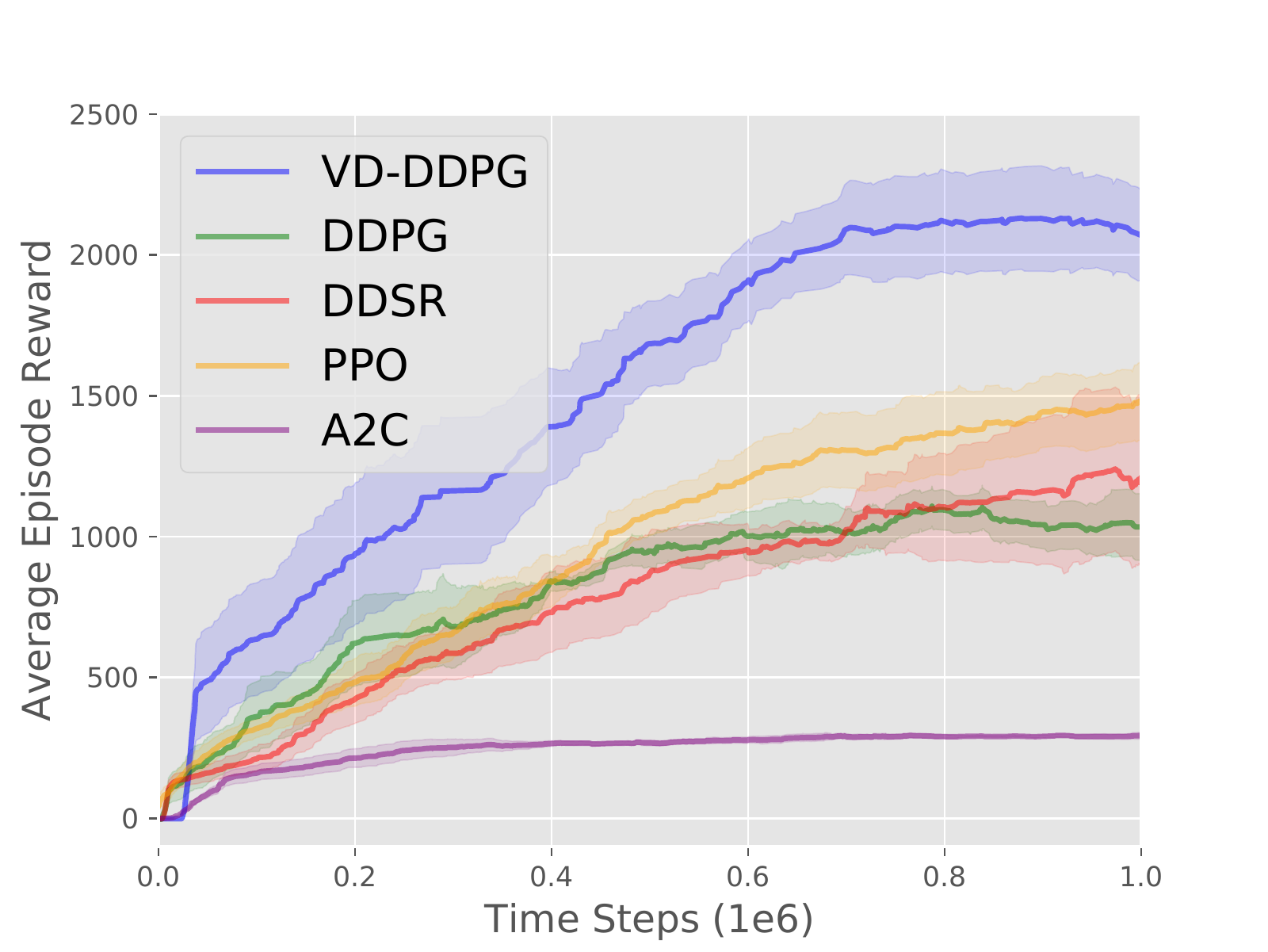}
}

\caption{Learning curves of algorithms in several OpenAI Gym continuous tasks.
Results are average episode reward ($\sum_{t=0}^{T}r_t$) over recent 100 episodes.
The shaded region denotes half a standard deviation of average evaluation over 5 trials. }
\label{figure:2}
\end{figure*}

During the training process of the conditional VAE, 
the encoder infers a latent distribution $\mathcal{N}(\mu_t, \sigma_t)$ of the trajectory representation $m_{t:t+k}$ conditioned on state $s_t$ and action $a_t$. 
A latent variable then is sampled with re-parameterization trick, i.e., $z_t = \mu_t + \sigma_t \cdot \mathcal{N}(0, I)$, which is taken as part of input by the decoder to reconstruct the trajectory representation. 
This process models the underlying stochasticity of potential future trajectories.
We train the conditional VAE with respect to the variational lower bound \cite{Kingma2013AEVB}, in a form of the reconstruction loss along with a KL divergence regularization term:
\begin{equation}
\label{eqation:11}
\begin{aligned}
    \mathcal{L}^{\rm VAE}(\phi, \varphi) = 
    & \mathbb{E}_{ \tau_{t:t+k} \sim \mathcal{D}} \Big[ \| m_{t:t+k} 
    - \tilde{m}_{t:t+k} \|_{2}^{2} \\
    & + \beta D_{\rm KL} \big( \mathcal{N}(\mu_t,\sigma_t) \| \mathcal{N}(0,I) \big) \Big],
\end{aligned}
\end{equation}
where $m_{t:t+k}$ is obtained from the representation model (Equation \ref{eqation:9}). 
We use a weight $\beta > 1$ to encourage VAE to discover disentangled latent factors for better inference quality, which is also known as a $\beta$-VAE \cite{Burgess2018Understanding,Higgins2017Beta}. 
See Ablation (Section \ref{section:ablation}) for the influence of different values of $\beta$.

After the training via instance-to-instance reconstruction, we can use the well-trained decoder to generate the instances of trajectory representation with latent variables sampled from the prior $\mathcal{N}(0,I)$.
However, this differs from the expected representation we expect for the predictive dynamics model (Equation \ref{eqation:4}).
To narrow down this discrepancy,
we propose using a clipped generative noise during the generation process,
which allows us to obtain high-quality prediction of expected future trajectories.
Finally, the predictive dynamics model $P^{\rm VAE}$ can be viewed as the generation process of the decoder with a clipped generative noise $\epsilon_g$:
\begin{equation}
\label{eqation:12}
\begin{aligned}
    \tilde{m}_{t:t+k} & = P^{\rm VAE}(s, a, \epsilon_{g}), \\
    {\rm where} \ \epsilon_{g} \sim {\rm Clip} & \big( \mathcal{N}(0,I), -c, c \big) \ {\rm and} \ c \ge 0.
\end{aligned}
\end{equation}
Empirically, we found such a clipping can achieve similar results to Monte Carlo sampling but with only one sample.
A further discussion on clip value $c$ is in Ablation \ref{section:ablation}.


\subsection{Overall Algorithm}
\label{subsection:algorithm}
With the three modules introduced above, now we derive deep RL algorithms from VDFP.
We build our algorithm on off-policy RL algorithm DDPG, by replacing the conventional critic ($Q$-function) with a decomposed one, namely Value Decomposed DDPG (\textbf{VD-DDPG}).
Accordingly, the actor (policy) is updated through value decomposed deterministic policy gradient (Equation \ref{eqation:7}) with respect to the VDFP critic.
Note our algorithm does not use target networks for both the actor and critic
since no TD target is calculated here.
The complete algorithm of VD-DDPG can be found in Supplementary Material \ref{supp:VDDDPG} Algorithm 1.

For on-policy RL, we also equip PPO with VDFP for the value estimation of $V$-function and obtain Value Decomposed PPO (\textbf{VD-PPO}). 
Details and empirical results are omitted 
and can be found in Supplementary Material \ref{supp:VDPPO}.

\section{Experiments}
\label{section:experiments}
We conduct our experiments on serveral representative continuous control tasks in OpenAI gym \cite{Brockman2016Gym,Todorov2012MuJoCo}.
We make no modifications to the original environments or reward functions (except the delay reward modification in Section \ref{delay-reward-section}).

\subsection{Evaluation}
\label{section:evaluation}

To evaluate the effectiveness of VDFP, 
we focus on four continuous control tasks: one Box2D task LunarLander-v2 and three MuJoCo InvertedDoublePendulum-v1, HalfCheetah-v1 and Walker2d-v1 \cite{Fujimoto2018TD3,Fujimoto2018BCQ,Schulman2015TRPO,Schulman2017PPO,Peng19AWR}.
We compare our derived approach VD-DDPG with DDPG \cite{Lillicrap2015DDPG}, PPO \cite{Schulman2017PPO}, and A2C \cite{Mnih2016AC}, as well as the Deterministic DSR (DDSR), a variant of DSR \cite{Kulkarni2016DSR} for continuous control.
Since DSR is originally proposed for discrete action problems, we implement DDSR based on DDPG according to the author's codes for DSR on GitHub.
VD-DDPG, DDSR and DDPG only differs in the implementation of the critic network to ensure a fair comparison.
For PPO and A2C, we adopt non-paralleled implementation and use 
GAE \cite{Schulman2016GAE} with $\lambda = 0.95$ for stable policy gradient.
Each task is run for 1 million timesteps and the results are reported over 5 random seeds of Gym simulator and network initialization.

\begin{table*}
  \caption{
  Performance of algorithms in HalfCheetah-v1 and Walker2d-v1 under the first delayed reward setting.
  Different delay steps ($d$) are listed from left to right.
  Results are max Average Episode Reward over 5 trials of 1 million timesteps.
  Downarrow ($\downarrow$) represents the percentage of performance decrease compared with the results under common setting in Figure \ref{figure:2} (i.e., $d$ = 0).
  }
  \label{table:delay_reward}
  \centering
  \scalebox{0.80}{
  \begin{tabular}{c|cccc|cccc}
    \toprule 
    & \multicolumn{4}{c}{HalfCheetah-v1} & \multicolumn{3}{c}{Walker2d-v1} \\
    \cmidrule(r){2-5} \cmidrule(r){6-9}
    Algorithm & $d$ = 16 & $d$ = 32 & $d$ = 64 & $d$ = 128 & $d$ = 16 & $d$ = 32 & $d$ = 64 & $d$ = 128\\
    \midrule
    VD-DDPG 
    & \textbf{4823} (17\%$\downarrow$) 
    & \textbf{4677} (\textbf{20\%}$\downarrow$) 
    & \textbf{3667} (\textbf{37\%}$\downarrow$) 
    & \textbf{2479} (\textbf{57\%}$\downarrow$) 
    & \textbf{1822} (14\%$\downarrow$) 
    & \textbf{1824} (14\%$\downarrow$) 
    & \textbf{1628} (24\%$\downarrow$) 
    & \textbf{570} (73\%$\downarrow$)\\
    DDPG
    & 2675 (43\%$\downarrow$) 
    & 1546 (67\%$\downarrow$) 
    & 1176 (75\%$\downarrow$) 
    & 909 (81\%$\downarrow$) 
    & 653 (41\%$\downarrow$) 
    & 583 (47\%$\downarrow$) 
    & 404 (64\%$\downarrow$) 
    & 325 (71\%$\downarrow$) \\
    DDSR 
    & 1337 (69\%$\downarrow$) 
    & 860 (80\%$\downarrow$) 
    & 832 (81\%$\downarrow$) 
    & -1 (100\%$\downarrow$) 
    & 249 (80\%$\downarrow$)
    & 212 (83\%$\downarrow$)
    & 146 (88\%$\downarrow$) 
    & 125 (90\%$\downarrow$) \\
    PPO 
    & 1894 (24\%$\downarrow$) 
    & 1721 (31\%$\downarrow$) 
    & 982 (61\%$\downarrow$) 
    & 689 (72\%$\downarrow$) 
    & 629 (39\%$\downarrow$) 
    & 351 (76\%$\downarrow$) 
    & 274 (81\%$\downarrow$) 
    & 210 (86\%$\downarrow$) \\
    A2C 
    & 207 (\textbf{9\%}$\downarrow$) 
    & 111 (51\%$\downarrow$) 
    & -35 (116\%$\downarrow$) 
    & -144 (163\%$\downarrow$) 
    & 287 (\textbf{2\%}$\downarrow$) 
    & 281 (\textbf{4\%}$\downarrow$) 
    & 276 (\textbf{6\%}$\downarrow$) 
    & 212 (\textbf{27\%}$\downarrow$) \\
    \bottomrule
  \end{tabular}
  }
\end{table*}

For VD-DDPG, we set the KL weight $\beta=$ 1000 as in \cite{Burgess2018Understanding} and the clip value $c$ as 0.2.
A max trajectory length
of 256 is used expect using 64 already ensures a good performance for HalfCheetah-v1.
An exploration noise sampled from $\mathcal{N}(0,0.1)$ \cite{Fujimoto2018TD3} is added to each action selected by the deterministic policy of DDPG, DDSR and VD-DDPG.
The discounted factor is 0.99 and we use Adam Optimizer \cite{KingmaB14Adam} for all algorithms.
Exact experimental details of algorithms are provided in Supplementary Material \ref{supp:experimental}.

The learning curves of algorithms are shown in Figure \ref{figure:2}.
We can observe that VD-DDPG matches or outperforms other algorithms in both final performance and learning speed across all four tasks.
Especially, VD-DDPG shows a clear margin over its off-policy counterparts, i.e., DDPG and DDSR.
This indicates that VDFP brings better value estimation and thus more effective policy improvement. 
Similar results are also found in the comparison between VD-PPO and PPO (see Supplementary Material \ref{supp:VDPPO}).

\subsection{Delayed Reward}
\label{delay-reward-section}
One potential advantage of VDFP is to alleviate the difficulty of value estimation especially in complex scenarios.
We further demonstrate the significant effectiveness and robustness of VDFP under delayed reward settings.
We consider two representative delayed reward settings in real-world scenarios:
1) multi-step accumulated rewards are given at sparse time intervals; 
2) each one-step reward is delayed for certain timesteps. 
To simulate above two settings, we make a simple modification to MuJoCo tasks respectively:
1) deliver $d$-step accumulated reward every $d$ timesteps;
2) delay the immediate reward of each step by $d$ steps.
With the same experimental setups in Section \ref{section:evaluation}, we evaluate the algorithms on HalfCheetah-v1 and Walker2d-v1 under different delayed reward settings, with a delay step $d$ from 16 to 128.
Table \ref{table:delay_reward} plots the results under the first delayed reward setting and similar results are also observed in the second delayed reward setting. 

As the increase of delay step $d$, all algorithms gradually degenerate in comparison with Figure \ref{figure:2} (i.e., $d = 0$).
DDSR can hardly learn effective policies under such delayed reward settings due to the failure of its immediate reward model even with a relatively small delay step (e.g., $d$ = 16).
VD-DDPG consistently outperforms others under all settings, in both learning speed and final performance (2x to 4x than DDPG).
Besides, VD-DDPG shows good robustness with delay step $d \le 64$.
As mentioned in Section \ref{section:model}, we suggest that the reason
can be two-fold:
1) with VDFP, it can always learn the dynamics 
effectively from states and actions, 
no matter how rewards are delayed;
2) the trajectory return model is robust with delayed reward since it approximates the overall utilities instead of one-step rewards.

\begin{table*}[t]
  \caption{Ablation of VDFP across each contribution in HalfCheetah-v1. 
  Results are max Average Episode Reward over 5 trials of 1 million timesteps.
  $\pm$ corresponds to half a standard deviation.
  LReLU/Linear denotes a single fully-connected layer with Leaky-ReLU/None activation.
  Note that Operator, KL Weight and Clip Value are not applicable (`--') for the MLP architecture.
  }
  \label{table:ablation}
  \centering
  \scalebox{0.68}{
  \begin{tabular}{cc|cc|cc|c|cc|c}
    \toprule
    \multicolumn{2}{c}{Representation} &  \multicolumn{2}{c}{Architecture} & \multicolumn{2}{c}{Operator}                 \\
    \cmidrule(r){1-2} \cmidrule(r){3-4} \cmidrule(r){5-6}
    CNN     & LSTM & VAE & MLP & Elem.-Wise Prod. & Concat. & Return Model & KL Weight ($\beta$) & Clip Value ($c$) & Results\\
    \midrule
    \checkmark & & \checkmark & & \checkmark & & Linear & 1000 & 0.2 & \textbf{5818.60} $\pm$ \ \textbf{336.25}\\
    \midrule
        & \checkmark & \checkmark & & \checkmark & & Linear & 1000 & 0.2 & 5197.03 $\pm$ \ 156.52\\
    \checkmark & & & \checkmark & -- & -- & Linear & -- & -- & 2029.00 $\pm$ \ 486.11\\
    \checkmark & & \checkmark & & & \checkmark & Linear & 1000 & 0.2 & 4541.71 $\pm$ \ 104.22\\
    \midrule
    \checkmark & & \checkmark & & \checkmark & & Convex (LReLu w/ $\alpha$=0.2) & 1000 & 0.2 & 4781.18 $\pm$ \ 440.50\\
    \checkmark & & \checkmark & & \checkmark & & Convex (LReLu w/ $\alpha$=0.5) & 1000 & 0.2 & 4985.66 $\pm$ \ 348.08\\
    \checkmark & & \checkmark & & \checkmark & & Convex (ICNN \cite{Amos17ICNN}) & 1000 & 0.2 & 5646.12 $\pm$ \ 328.61\\
    \checkmark & & \checkmark & & \checkmark & & Convex (NE-ICNN \cite{Chen19OptimalControl}) & 1000 & 0.2 & 5310.23 $\pm$ \ 198.35\\
    \midrule
    \checkmark & & \checkmark & & \checkmark & & Linear & 100 & 0.2 & 4794.96 $\pm$ \ 370.02\\
    \checkmark & & \checkmark & & \checkmark & & Linear & 10 & 0.2 & 3933.33 $\pm$ \ 361.82\\
    \midrule
    \checkmark & & \checkmark & & \checkmark & & Linear & 1000 & $\infty$ & 4752.84 $\pm$ \ 328.75\\
    \checkmark & & \checkmark & & \checkmark & & Linear & 1000 & 0.0 & 5712.55 $\pm$ \ 233.74\\
    \bottomrule
  \end{tabular}
  }
\end{table*}

\subsection{Ablation}
\label{section:ablation}

We analyze the contribution of each component of VDFP:
1) CNN (v.s. LSTM) for trajectory representation model (Representation); 
2) conditional VAE (v.s. MLP) for predictive dynamics model (Architecture);
3) element-wise product (v.s. concatenation) for conditional encoding process (Operator);
and 4) model choices (linear v.s. convex)
for trajectory return function (Return). 
We interpolate the slope $\alpha$ for negative input of a single fully-connected layer with Leaky-ReLU activation from 1 (linear) to 0 (ReLU-activated).
Specially, we also consider two more advanced convex models, i.e., Input Convex Neural Network (ICNN) \cite{Amos17ICNN} and Negation-Extened ICNN (NE-ICNN) \cite{Chen19OptimalControl}.
We use the same experimental setups in Section \ref{section:evaluation}. 
The results are presented in Table \ref{table:ablation} and complete curves are in Supplementary Material \ref{supp:complete_curves}.

First, we can observe that CNN achieves better performance than LSTM.
Additionally, CNN also shows lower training losses and takes less practical training time (almost 8x faster) in our experiments.
indicating CNN is able to extract effective features.
Second, the significance of conditional VAE is demonstrated by its superior performance over MLP.
This supports our analysis in Section \ref{section:conditional-VAE}. 
Conditional VAE can well capture the trajectory distribution via the instance-to-instance reconstruction and then obtain the expected representation during the generation process.
Third, element-wise product shows an apparent improvement over concatenation.
We hypothesize that the explicit relation between the condition and representation imposed by element-wise product, forces the conditional VAE to learn more effective hidden features.
Lastly, adopting linear layer for trajectory return model slightly outperforms the case of using convex models.
This is due to the equality between the composite function approximation and the $Q$-function (Lemma \ref{lemma:lower_bound}) that ensured by the linear layer.
ICNN and NE-ICNN are comparable with the linear case and better than Leaky-ReLU layers due to their stronger representation ability.

Moreover, we analyse the influence of weight $\beta$ for KL loss term (Equation \ref{eqation:11}).
The results are consistent to the studies of $\beta$-VAE \cite{Burgess2018Understanding,Higgins2017Beta}:
larger $\beta$ applies stronger emphasis on VAE to discover disentangled latent factors, resulting in better inference performance.
For clip value $c$ in prediction process (Equation \ref{eqation:12}), clipping achieves superior performance than not clipping ($c = \infty$) since this narrows down the discrepancy between prediction instance and expected representation of future trajectory.
Besides, using the mean values of latent variables ($c = 0.0$) does not necessarily ensure the generation of the expected representation, since the decoder network is non-linear.
Considering the non-linearity and imperfect approximation of neural networks, 
setting $c$ to a small value ($c = 0.2$) empirically performs slightly better.

\section{Discussion and Conclusion}
\textbf{Discussion.}
In VDFP, we train a variational predictive dynamics model (conditional VAE) from sampled trajectories.
To some extent, it can be viewed as a Monte Carlo (MC) learning \cite{Sutton1988ReinforcementLA} at the perspective of trajectory representation.
MC is known in traditional RL studies to suffer from high variance and is almost dominated by TD in deep RL.
A recent work \cite{AmiranashviliDK18TDor} provides an evidence that finite MC can be an effective alternative of TD in deep RL problems.
In our paper, the idea of decomposed MC estimation reflected in VDFP may direct a possible way to address the variance by separately modeling dynamics and returns, 
i.e., a new variational MC approach which is applicable in deep RL.

Another thing is that the conditional VAE is trained in an off-policy fashion for VD-DDPG.
It is supposed to be flawed since trajectories collected by old policies may not able to represent the future under current policy.
However, we do not observe apparent adverse effects of using off-policy training in our experiments, and explicitly introducing several off-policy correction approaches shows no apparent benefits.
Similar results are also found in DFP \cite{Dosovitskiy2017DFP} and D4PG \cite{Barth-MaronHBDH18D4PG}.
It may also be explained as in 
AWR \cite{Peng19AWR}
which optimizes policy with regard to advantages over an averaged baseline value function that trained from the off-policy data collected by different policies.
Analogically, an off-policy trained VAE can be considered to model the average distribution of trajectory representations under recent policies.

VDFP allows flexible uses of decomposed models.
We believe it can be further integrated with such as pre-training either part from offline trajectory data, reusing from and transfer to different tasks.
Besides, recent advances in Representation Learning \cite{Zhang20Bisim,Srinivas20CURL,Schwarzer20Momen} can also be 
incorporated.
See Supplementary Material \ref{supp:further_discussion} for further discussions.
We expect to further investigate the problems discussed above in the future.


\textbf{Conclusion.}
we present an explicit two-step understanding of value estimation from the perspective of trajectory-based latent future prediction.
We propose VDFP which allows value estimation to be conducted effectively and flexibly in different problems.
Our experiments demonstrate the effectiveness of VDFP for both off-policy and on-policy RL 
especially in delayed reward settings.
For future work, it is worthwhile to investigate the problems discussed above.

\section*{Acknowledgments}
The work is supported by the National Natural Science Foundation of China (Grant Nos.: 61702362, U1836214, U1813204), Special Program of Artificial Intelligence and Special Program of Artificial Intelligence of Tianjin Municipal Science and Technology Commission (No.: 56917ZXRGGX00150), Tianjin Natural Science Fund (No.: 19JCYBJC16300), Research on Data Platform Technology Based on Automotive Electronic Identification System.

\bibliography{aaai21_vdfp}

\clearpage

\onecolumn
\appendix

\section{Missing Proof and Formulation}
\label{supp:lem1_proof}
\subsection{Complete Proof for Lemma \ref{lemma:lower_bound}}
With the definitions of the trajectory return function $U$ and the predictive dynamics function $P$, we derive the following lemma \ref{lemma:lower_bound}

\emph{Proof.}
Remind the definition of state-action value function $Q$:
\begin{equation}
\label{eqation:Q_function}
    Q^{\pi}(s,a) = \mathbb{E} \left[\sum_{t=0}^{T}\gamma^{t} r_t|s_0=s, a_0=a; \pi \right]. 
\end{equation}
Replace the summation in Equation \ref{eqation:Q_function} with the trajectory return function $U$,
\begin{equation}
\label{eqation:Q_function_with_U}
    Q^{\pi}(s,a) = \mathbb{E} \left[U(m_{0:T})|s_0=s, a_0=a; \pi \right]. 
\end{equation}
Exchange the expectation and the function $U$ at the RHS of Equation \ref{eqation:Q_function_with_U}, and with \emph{Jensen's Inequality} we obtain,
\begin{equation}
\begin{aligned}
    Q^{\pi}(s,a) & = \mathbb{E} \big[U(m_{0:T})|s_0=s, a_0=a; \pi \big] \\
    & \ge U \big(\mathbb{E} \left[m_{0:T}|s_0=s, a_0=a; \pi \right] \big) \\
    & = U \big(P^{\pi}(s,a) \big) ,
\end{aligned}
\end{equation}
when function $U$ is convex. 
The equality can be strictly established when $U$ is linear.
A similar lemma can also be derived for the state value function $V$ case.

\subsection{Complete Formulation for the conditional VAE}
\label{supp:VAE_form}

A variational auto-encoder (VAE) is a generative model which aims to maximize the marginal log-likelihood $\log p(X) = \sum_{i=1}^N \log p(x_i)$ where $X = \{x_1, \dots ,x_N\}$, the dataset.
While computing the marginal likelihood is intractable in nature, it is a common choice to train the VAE through optimizing the variational lower bound:
\begin{equation}
    \log p(X) \ge \mathbb{E}_{q(X|z)}[\log p(X|z)] + D_{\rm KL} \big(q(z|X) \| p(z) \big),
\end{equation}
where $p(z)$ is chosen a prior, generally the multivariate normal distribution $\mathcal{N}(0,I)$.

In our paper, we use a conditional VAE to model the latent distribution of the representation of  future trajectory conditioned on the state and action, under certain policy.
Thus, the true parameters of the distribution, denoted as $\varphi^*$, maximize the conditional log-likelihood as follows:
\begin{equation}
    \varphi^* = \arg \max_{\varphi} \sum_{i=1}^N \log p^{\pi}_{\varphi}(m_{t:t+k}|s_t,a_t).
\end{equation}
The likelihood can be calculated with the prior distribution of latent variable $z$:
\begin{equation}
    p^{\pi}_{\varphi}(m_{t:t+k}|s_t,a_t) = \int p^{\pi}_{\varphi}(m_{t:t+k}|z_t, s_t, a_t) p^{\pi}_{\varphi}(z_t) {\rm d}z.
\end{equation}
Since the prior distribution of latent variable $z$ is not easy to compute, an approximation of posterior distribution $q^{\pi}_{\phi}(z_t|m_{t:t+k},s_t,a_t)$ with parameterized $\phi$ is introduced.

The variational lower bound of such a conditional VAE can be obtained as follows, superscripts and subscripts are omitted for clarity:
\begin{equation}
\begin{aligned}
        D_{\rm KL} & \big(q_{\phi}(z|m,s,a) \| p_{\varphi}(z|m,s,a) \big)  \\
        & = \int q_{\phi}(z|m,s,a) \log \frac{q_{\phi}(z|m,s,a)}{p_{\varphi}(z|m,s,a)} {\rm d}z \\
        & = \int q_{\phi}(z|m,s,a) \log \frac{q_{\phi}(z|m,s,a) p_{\varphi}(m|s,a)}{p_{\varphi}(z,m|s,a)} {\rm d}z \\
        & = \int q_{\phi}(z|m,s,a) \big[ \log p_{\varphi}(m|s,a) +  \log \frac{q_{\phi}(z|m,s,a) }{p_{\varphi}(z,m|s,a)} \big]{\rm d}z \\
        & = \log p_{\varphi}(m|s,a) + \int q_{\phi}(z|m,s,a)  \log \frac{q_{\phi}(z|m,s,a) }{p_{\varphi}(z,m|s,a)} {\rm d}z \\
        & = \log p_{\varphi}(m|s,a) + \int q_{\phi}(z|m,s,a)  \log \frac{q_{\phi}(z|m,s,a) }{p_{\varphi}(m|z,s,a) p_{\varphi}(z|s,a)} {\rm d}z \\
        & = \log p_{\varphi}(m|s,a) + \int q_{\phi}(z|m,s,a) \big[ \log \frac{q_{\phi}(z|m,s,a) }{p_{\varphi}(z|s,a)} - \log p_{\varphi}(m|z,s,a) \big] {\rm d}z \\
        & = \log p_{\varphi}(m|s,a) + D_{\rm KL} \big( q_{\phi}(z|m,s,a) \| p_{\varphi}(z|s,a) \big) - \mathbb{E}_{z \sim q_{\phi}(z|m,s,a)} \log p_{\varphi}(m|z,s,a). \\
\end{aligned}
\end{equation}
Re-arrange the above equation,
\begin{equation}
\begin{aligned}
    \log p_{\varphi}(m|s,a) - & D_{\rm KL} \big(q_{\phi}(z|m,s,a) \| p_{\varphi}(z|m,s,a) \big) \\
    & =  \mathbb{E}_{z \sim q_{\phi}(z|m,s,a)} \log p_{\varphi}(m|z,s,a)
    - D_{\rm KL} \big( q_{\phi}(z|m,s,a) \| p_{\varphi}(z|s,a) \big). \\
\end{aligned}
\end{equation}
Since the KL divergence is none-negative, we obtain the variational lower bound for the conditional VAE:
\begin{equation}
    \log p_{\varphi}(m|s,a) \ge \log p_{\varphi}(m|s,a) - D_{\rm KL} \big(q_{\phi}(z|m,s,a) \| p_{\varphi}(z|m,s,a) \big) = - \mathcal{L}^{\rm VAE},
\end{equation}
Thus, we can obtain the optimal parameters through optimizing the equation below:
\begin{equation}
\label{equation:elbo}
\begin{aligned}
    \mathcal{L}^{\rm VAE} = - \mathbb{E}_{z \sim q_{\phi}(z|m,s,a)} \log p_{\varphi}(m|z,s,a) &
    + D_{\rm KL} \big( q_{\phi}(z|m,s,a) \| p_{\varphi}(z|s,a) \big), \\
    \phi^*, \varphi^* = & \arg \min_{\phi,\varphi} \mathcal{L}^{\rm VAE}.
\end{aligned}
\end{equation}

$q_{\phi}(z|m,s,a)$ is the conditional variational encoder and the $\log p_{\varphi}(m|z,s,a)$ is the conditional variational decoder. 

In our paper, we implement the encoder and decoder with deep neural networks.
The encoder takes the trajectory representation and state-action pair as input and output a Gaussian distribution with mean $\mu_t$ and standard deviant $\sigma_t$, from which a latent variable is sampled and then feed into the decoder for the reconstruction of the trajectory representation.
Thus, we train the conditional VAE with respect to the variational lower bound (Equation \ref{equation:elbo}), in the following form:
\begin{equation}
    \mathcal{L}^{\rm VAE}(\phi, \varphi) = \mathbb{E}_{\tau_{t:t+k} \sim \mathcal{D}} \Big[ \|m_{t:t+k} - \tilde{m}_{t:t+k}\|_2^2
    + D_{\rm KL} \big( \mathcal{N}(\mu_t,\sigma_t) \| \mathcal{N}(0,I) \big) \Big].
\end{equation}

\section{Discussion for Conditional VAE and MLP}
\label{supp:VAE_discussion}

What the predictive dynamics function $P$ exactly does is a prediction problem from input $x$ (state-action pair) to true target $y$ (expected embedding of future trajectory) with stochastic samples of target in the dataset $\mathcal{D}$:
\begin{equation}
\begin{aligned}
    \mathcal{D} = & \{x_i, \hat{y_i}\}_{i=0}^k , \\
    {\rm where} \ \hat{y_i} \sim Dist(& y_i) \ {\rm and} \ y_i = f(x_i, h).
\end{aligned}
\end{equation}
$f$ is the true function that maps the input $x$ to target $y$, and $h$ is the hidden factor (policy $\pi$ in our problem).

For an MLP, it approximates function $f$ over $\mathcal{D}$ directly, without considering the hidden factor $h$ nor the underlying distribution $Dist(y)$.
Thus, an MLP can do poor approximation under such circumstance.
In contrast, in our paper we train a conditional VAE to recover the underlying distribution $Dist(y)$ via sample-to-sample reconstruction, and then clip the generative noise to get the approximation of true target $y$.

Actually, different from common supervised learning problem with a fixed target $f$, in our RL problem, it is a moving target approximation problem since the policy (the hidden factor $h$) can change, which can be more challenging.
Therefore, we suggest that conditional VAE can better capture the latent variables of the stochastic and moving targets) than MLP, which is significant to our approach.

See Figure \ref{subfig:vae_mlp} in Supplementary Material \ref{supp:complete_curves} for the empirical comparison between VAE and MLP.
In addition to MLP, we provide the results of using an MLP with random Gaussian noise as input of $P$ instead of VAE trained latent,
showing a better performance than MLP w/o random noise but still much worse than using VAE.

\section{Deep Reinforcement Learning with Value Decomposition with Future Prediction}

\subsection{Value Decomposed Deep Deterministic Policy Gradient}
\label{supp:VDDDPG}

We propose our first Deep RL algorithm, i.e., Value Decomposed DDPG (\textbf{VD-DDPG}), through replacing the conventional value function of DDPG (the critic, $Q$-function) with a decomposed value function according to VDFP.
VD-DDPG is an off-policy algorithm with deterministic policy as DDPG but with a two-step decoupling value estimation in contrast.
The complete algorithm of VD-DDPG is shown in Algorithm \ref{algorithm:vd-ddpg}.
The predictive model (conditional VAE) is trained with mini-batch samples of state-action trajectories in Line 8-9.
The trajectory representation model and trajectory return model is trained together in Line 14-17.
The policy is trained with the forward predicted objective obtained by future prediction and then return evaluation in Line 10-11 (see the dashed lines in Figure \ref{figure:1}).   

Empirical results for VD-DDPG are provided in Section \ref{section:experiments}.

\begin{algorithm}[t]
  \caption{Value Decomposed Deep Deterministic Policy Gradient (VD-DDPG) algorithm}
  \begin{algorithmic}[1]
    \State Initialize actor 
    with random parameters $\theta$, and buffer $D$ 
    \State Initialize representation model and trajectory return model with random parameters $\omega$
    \State Initialize conditional VAE with random parameters $\phi$, $\varphi$
	\For{episode $= 1, 2, \dots$}
	    \For{$t = 1, 2, \dots, T$}
	        \State Observe state $s_t$ and select action $a_t = \pi_{\theta}(s_t) + \epsilon_{\rm e}$, with exploration noise $\epsilon_{\rm e} \sim \mathcal{N}(0,\sigma)$
	        \State Execute action $a_t$ and obtain reward $r_t$
	        \State Uniformly sample a mini-batch of 
	        $N$ experience $\{ ( \tau_{t_i:T} ) \}_{i=1}^{N}$ from $\mathcal{D}$
	        \State Update $\phi$, $\varphi$ by minimizing $\mathcal{L}^{\rm VAE}(\phi, \varphi)$ (Equation \ref{eqation:11})
	        \State Predict the representation of future trajectory $\tilde{m} = P^{\rm VAE}(s,a,\epsilon_{g})|_{s=s_i,a=\pi_{\theta}(s_i)}$
	        \State Update $\theta$ with value-decomposed deterministic policy gradient (Equation \ref{eqation:7})
        \EndFor
        \State Store experiences $\{ (\tau_{t:T}, r_{t:T}) \}_{t=1}^{T}$ in $\mathcal{D}$
        \For{epoch $= 1, 2, \dots$, num\_epoch}
            \State Uniformly sample a mini-batch of $N$ experience $\{ (\tau_{t_i:T}, r_{t_i:T}) \}_{i=1}^{N}$ from $\mathcal{D}$
    	    \State Update representation model $f^{\rm CNN}$ and return model $U^{\rm Linear}$ by minimizing $\mathcal{L}^{\rm Ret}(\omega)$ (Equation \ref{eqation:10})
        \EndFor
	\EndFor
  \end{algorithmic}
\label{algorithm:vd-ddpg}
\end{algorithm}

\subsection{Value Decomposed Proximal Policy Optimization}
\label{supp:VDPPO}

Since VD-DDPG is an off-policy RL algorithm with deterministic policy, we also demonstrate the effectiveness of our decomposed model on on-policy RL algorithm with stochastic policy.
We provide Value-decomposed PPO (VDFP-PPO) algorithm by modifying the $V$-function of PPO algorithm according to VDFP.
In Section \ref{subsection:algorithm}, we omit our second proposed algorithm, i.e., Value Decomposed PPO (\textbf{VD-PPO}), due to space limitation.
The complete algorithm of VD-PPO is shown in Algorithm \ref{algorithm:vd-ppo}.

\begin{algorithm*}[ht]
  \caption{Value decomposed Proximal Policy Optimization (VD-PPO) algorithm}
  \begin{algorithmic}[1]
    \State Initialize a Gaussian policy $\pi$ with random parameters $\theta$
    \State Initialize representation model and trajectory return model with random parameters $\omega$
    \State Initialize conditional VAE with random parameters $\phi$, $\varphi$
    \State Initialize the return model buffer $\mathcal{D}$ and the trajectory batch buffer $\mathcal{B}$ as empty set \O
	\For{episode $= 1, 2, \dots$}
	    \For{$t = 1, 2, \dots, T$}
	        \State Observe state $s_t$ and select action $a_t \sim \pi_{\theta}(s_t)$
	        \State Execute action $a_t$ and obtain reward $r_t$
        \EndFor
        \State Add trajectories $\{ (\tau_{t:T}, r_{t:T}) \}_{t=1}^{T}$ in $\mathcal{D}$ and $\mathcal{B}$
        \For{epoch $= 1, 2, \dots$, \texttt{num\_epoch}}
            \State Uniformly sample a mini-batch of $N$ experience $\{ (\tau_{t_i:T}, r_{t_i:T}) \}_{i=1}^{N}$ from $\mathcal{D}$
    	    \State Update representation model $f^{\rm CNN}$ and return model $U^{\rm Linear}$ by minimizing $\mathcal{L}^{\rm Ret}(\omega)$
        \EndFor
        \If {episode \% \texttt{update\_freq} = 0}
            \For{epoch $= 1, 2, \dots$, \texttt{num\_epoch}}
    	        \State Update $\phi$, $\varphi$ by minimizing $\mathcal{L}^{\rm VAE}(\phi, \varphi)$ with trajectory batch $\mathcal{B}$
	            \State Predict state values $\hat{V}(s_i) = U\big(P^{\rm VAE}(s, \epsilon_{g})|_{s=s_i}\big)$ 
	            \State Compute advantage estimates $\hat{A}(s_i, a_i) = \sum^{T}_{k=i} \gamma^{k-i} r_i - \hat{V}(s_i)$
	            \State Update $\theta$ by optimizing the clipped surrogate objective as in vanilla PPO
            \EndFor
            \State Empty the trajectory batch buffer $\mathcal{B}$
        \EndIf
	\EndFor
  \end{algorithmic}
\label{algorithm:vd-ppo}
\end{algorithm*}

To evaluate the decomposed value estimation in on-policy RL and stochastic policy setting, we compare VD-PPO with PPO algorithm in several Mujoco tasks.
For a fair comparison, we adopt the vanilla implementation of PPO (without using GAE), from which we build the VD-PPO algorithm. 
VD-PPO and PPO use the same hyperparameter setting and most network structure design.
The only difference between PPO and VD-PPO is the implementation of value estimation.
More implementation details can be found in Supplementary Material \ref{supp:experimental}.

\begin{wrapfigure}{r}{0cm}
\begin{minipage}[t]{.37\linewidth}
\centering
\vspace{-0.5cm}
\includegraphics[width=1.\textwidth]{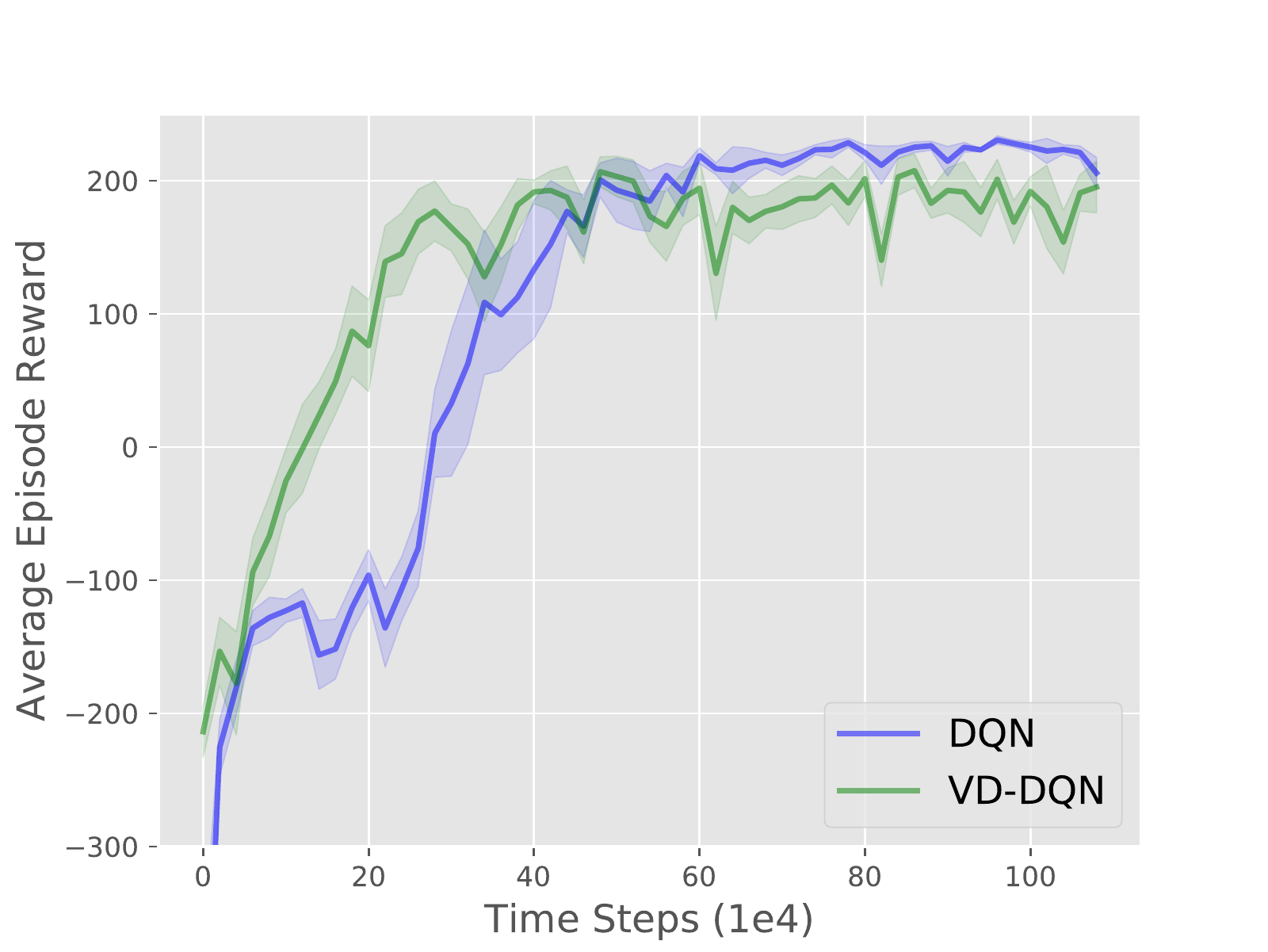}
\end{minipage}
\caption{
Learning curves of VD-DQN and DQN in LunarLander-v2.
Results are means and half a standard deviations over 5 trials. 
}
\vspace{-1.2cm}
\label{figure:vddqn}
\end{wrapfigure}

\begin{figure}
\centering
\subfigure[HalfCheetah-v1]{
\includegraphics[width=0.26\textwidth]{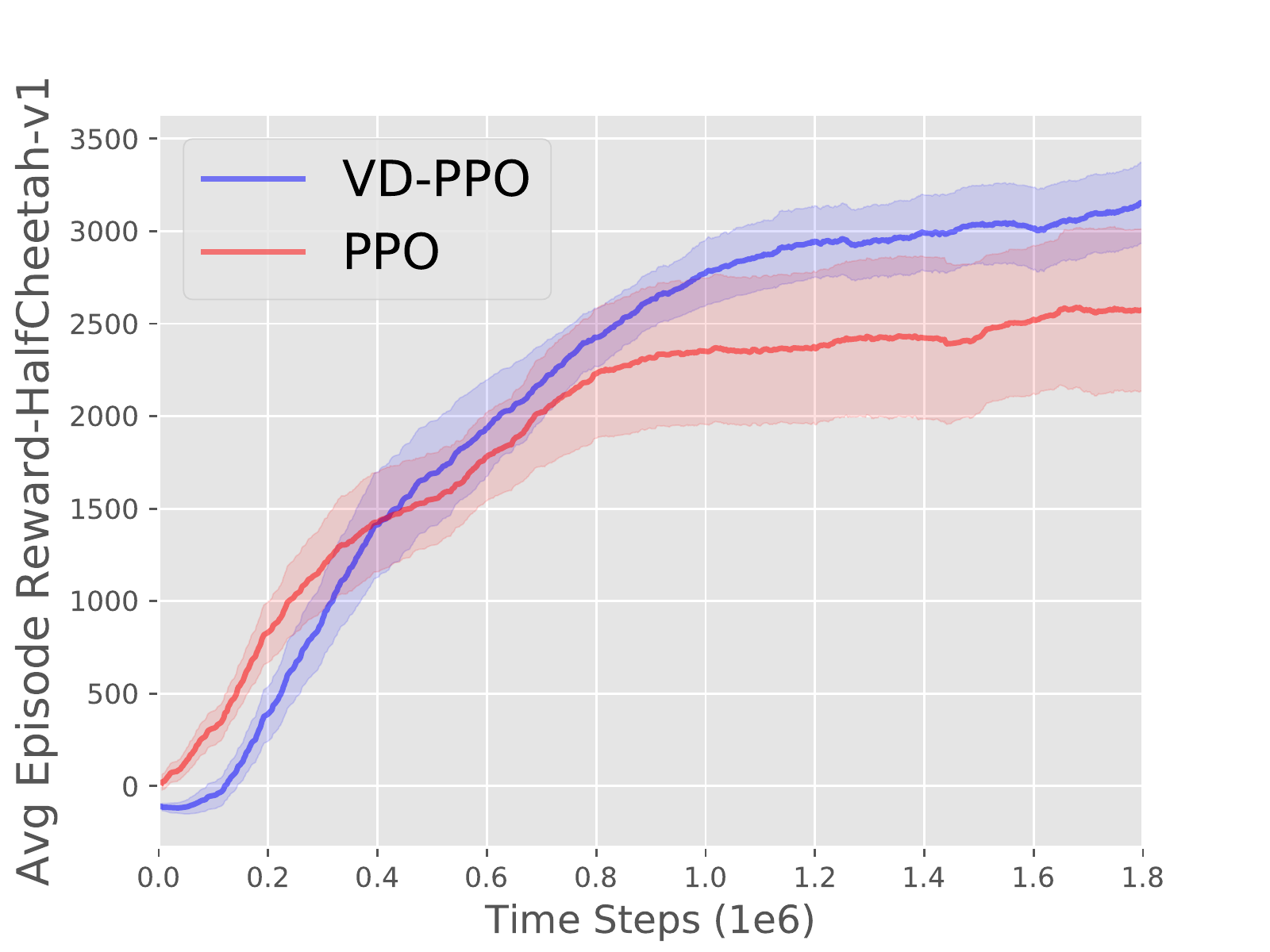}
}
\hspace{-0.7cm}
\subfigure[Walker2d-v1]{
\includegraphics[width=0.26\textwidth]{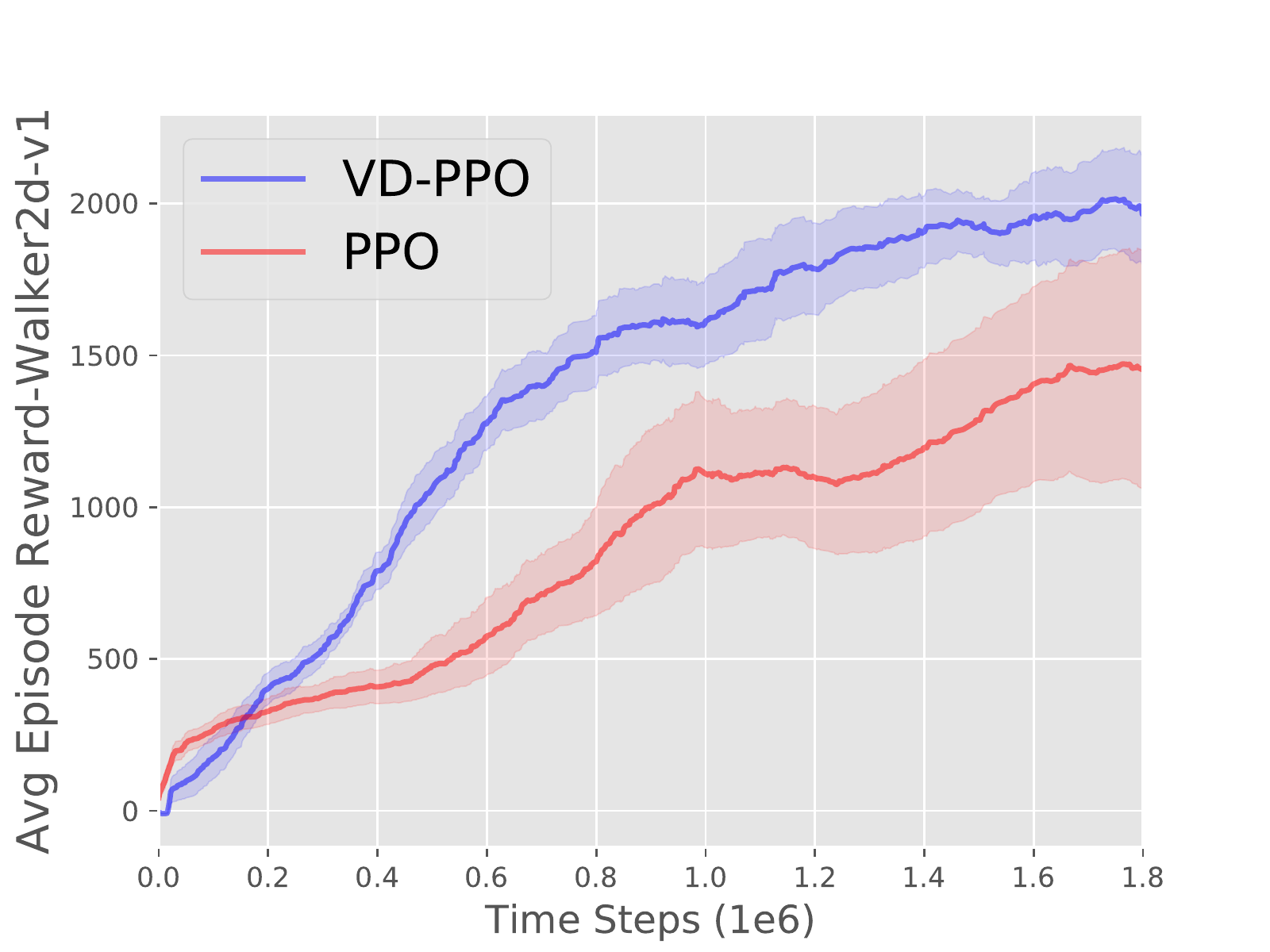}
}
\hspace{-0.7cm}
\subfigure[Hopper-v1]{
\includegraphics[width=0.26\textwidth]{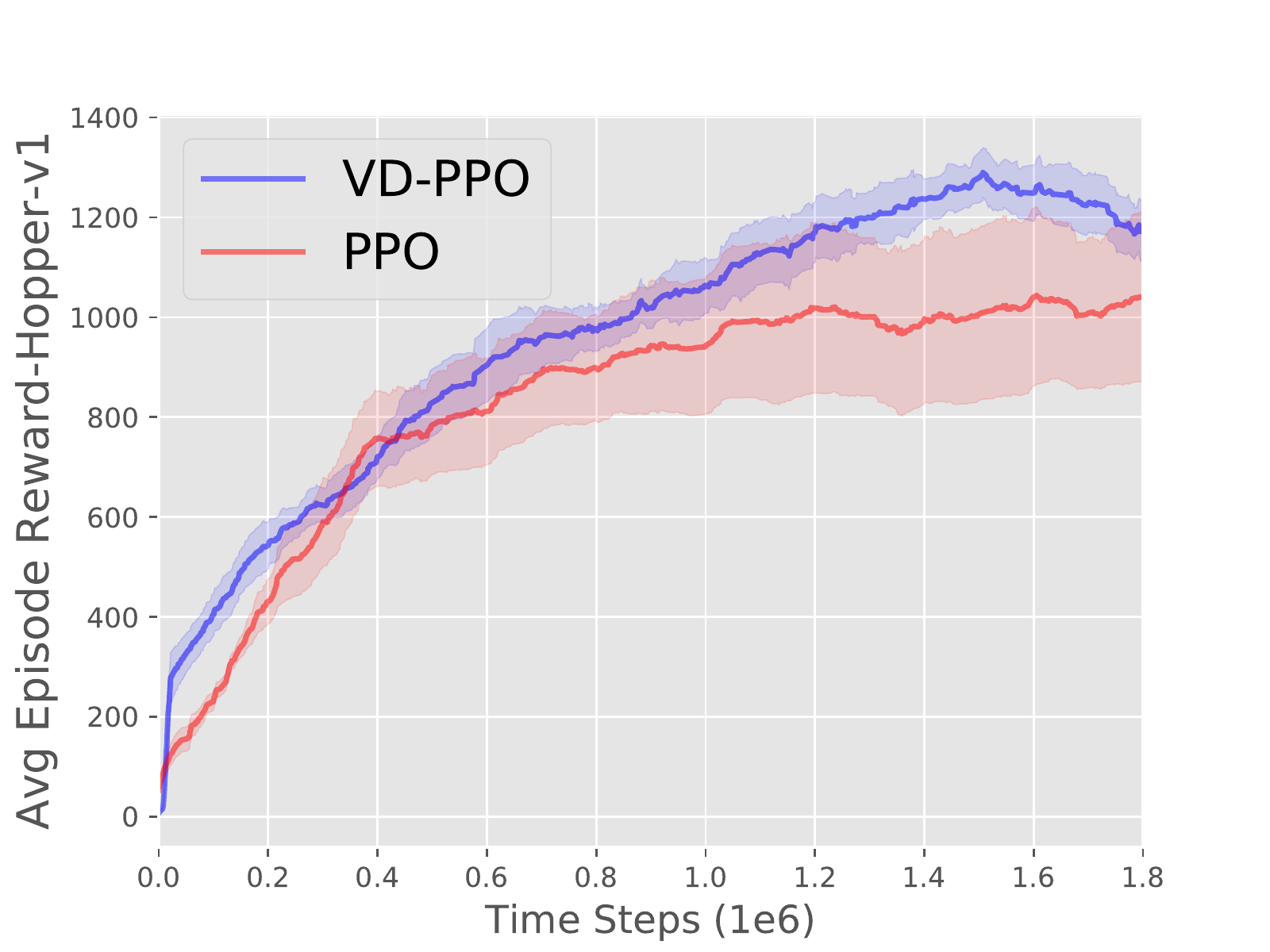}
}
\hspace{-0.7cm}
\subfigure[Ant-v1]{
\includegraphics[width=0.26\textwidth]{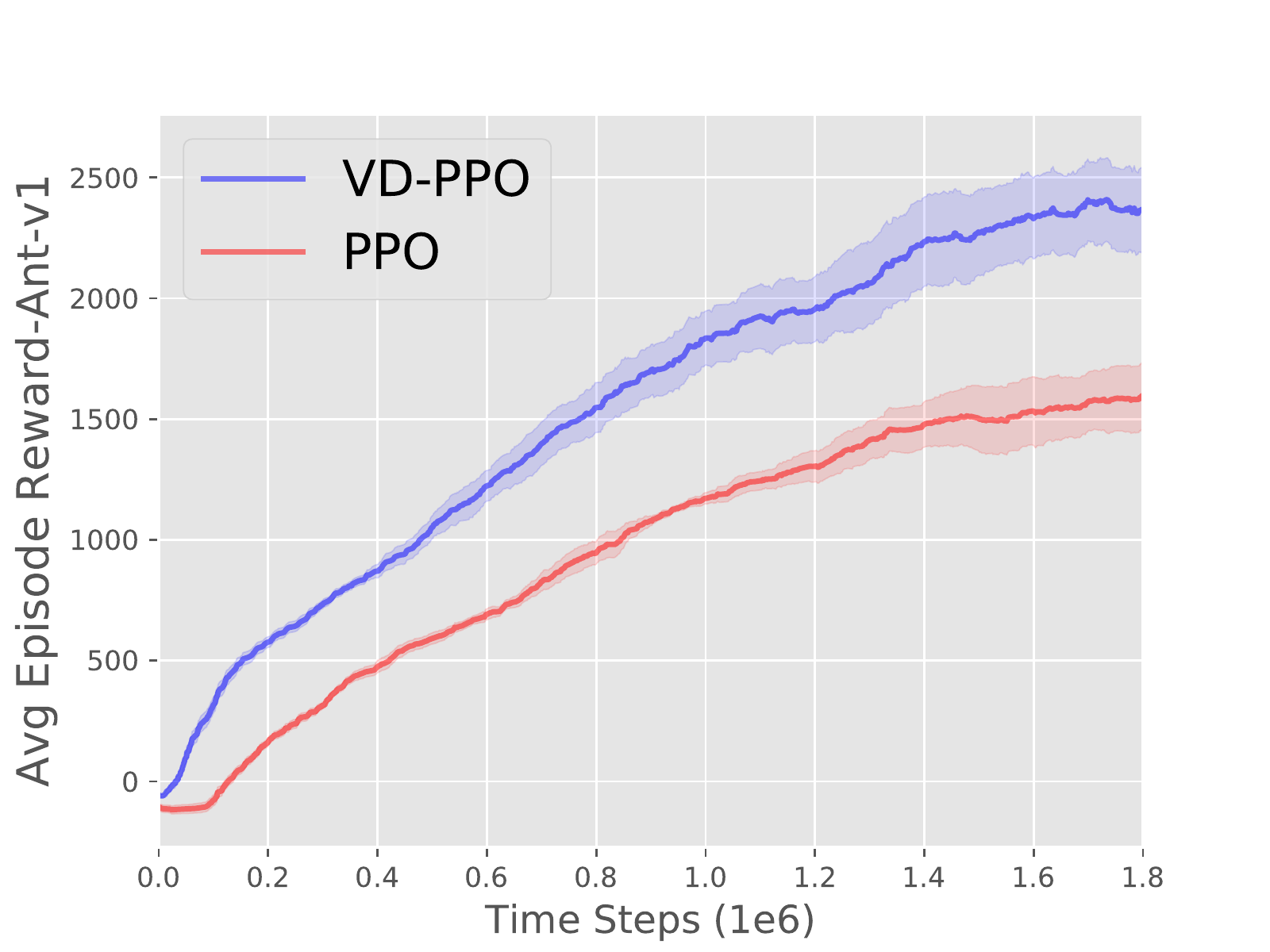}
}

\caption{Learning curves of VD-PPO and PPO (both w/o GAE) across four Mujoco tasks.
The shaded region denotes half a standard deviation of average evaluation over 5 trials. 
Results are smoothed over recent 100 episodes.}
\label{figure:vdfp-ppo}
\end{figure}

Figure \ref{figure:vdfp-ppo} shows the comparison results between VD-PPO and PPO across 4 mujoco tasks.
We can observe that VD-PPO outperforms PPO with a clear margin in all tasks.
This means that the decomposed value estimation can be more effective with explicit two separate models than the vanilla MC estimation which uses a coupling MLP model.

\subsection{Value Decomposed Deep $Q$-Network}
\label{supp:VDDQN}

Additionally, we implemented VDFP for the $Q$-network of DQN \cite{Mnih2015DQN}, called VD-DQN,
and conducted a comparison between VD-DQN v.s. DQN in LunarLander-v2, a discrete action version provided in OpenAI Gym
which has vector state input.
The results is shown in Figure \ref{figure:vddqn}.

VD-DQN shows about 2x faster learning than DQN with respect to time step and converges to the same level as DQN.
A further evaluation and development of VD-DQN in Atari is planned for future work.


\subsection{Some Further Discussions}
\label{supp:further_discussion}

\subsubsection{Decoupling Training in RL.}
Intuitively, the complexity of underlying form of the coupling value function can escalate in challenging problems. 
Thus, it increases the difficulty of approximation and the approximation error can induce learning problems, 
e.g., aggravate overestimation issue as analyzed in \cite{AnschelBS17AVGQ}.  

A reasonable decoupling can ease the learning difficulty (e.g., supervised trained trajectory return model in our paper). 
Similar ideas are also studied in SR \cite{Barreto17SF} and DFP \cite{Dosovitskiy2017DFP}; 
recently many works seek to covert RL problems to decoupled Supervised Learning problems for easier implementation and more stable training \cite{AbdolmalekiSTMH18MPO,Peng19AWR,Srivastava19UpsideDown}.
Moreover, the decoupling training in our paper are highly modular which allows flexible training choices and the potentials of transfer in multi-task problems.

\subsubsection{Policy Learning with VDFP Lower bound.}
Both SPG and DPG is gradient ascent with respect to the learning objective (expected return) and the difference of many algorithms lies at how to estimate or approximate expected returns. 
VDFP proposes a lower-bound approximation of expected return, which can be viewed as a pessimistic objective to some degree.
The relation between VDFP and other pessimistic objectives, like Clipped Double-Q \cite{Fujimoto2018TD3} and Maxmin-Q \cite{LanPFW20MAXMINQ} to address overestimation issue, deserves further study.

\subsubsection{Trajectory Representation in Larger State Space.}
It is a common problem for representation learning that large state space (or high-dimension input) increases the difficulty of learning. 
For pixel-input problems (e.g., Atari), we think additional CNNs are need for feature extraction. 
It is notable that the trajectory representation of VDFP is expected to contain necessary information for return prediction, 
instead of the details for state or trajectory reconstruction (as often considered in model-based RL), alleviating the difficulty of learning to some extent. 
Moreover, reconstruction-free unsupervised representation learning like Contrastive Learning \cite{Srinivas20CURL,Schwarzer20MPR,Fu20CMRL} is well compatible with VDFP in this case.

\section{Experimental Details}
\label{supp:experimental}

\subsection{Environment Setup}
We conduct our experiments on MuJoCo continuous control tasks in OpenAI gym.
We use the OpenAI gym with version 0.9.1, the mujoco-py with version 0.5.4 and the MuJoCo products with version MJPRO131.
Our codes are implemented with Python 3.6 and Tensorflow 1.8.
For more details of our code can refer to the \texttt{ReadME\_for\_Code.pdf} and \texttt{SourceCode.zip}.

\subsection{Network Structure}

Our DDPG and PPO are implemented with reference to \texttt{github.com/sfujim/TD3} (TD3 official source code) and \texttt{github.com/pat-coady/trpo} respectively.
For a fair comparison, DDSR and VD-DDPG are modified from our DDPG implementation which only differ in the critic, same for A2C and VD-PPO (see the additional experiments in Appendix C) from PPO.
    
As shown in Table \ref{table:ac_network}, we use a two-layer feed-forward neural network of 200 and 100 hidden units with ReLU activation (except for the output layer) for the actor network for all algorithms, and for the critic network for DDPG, PPO and A2C.
For PPO and A2C, the critic denotes the $V$-network.

\begin{table}[ht]
  \caption{Network structures for the actor network and the critic network ($Q$-network or $V$-network).
  }
  \label{table:ac_network}
  \centering
  \scalebox{1.0}{
  \begin{tabular}{ccc}
    \toprule
    Layer & Actor Network ($\pi(s)$) & Critic Network ($Q(s,a)$ or $V(s)$)\\
    \midrule
    Fully Connected & (state dim, 200) & (state dim, 200) \\
    Activation & ReLU & ReLU \\
    \midrule
    Fully Connected & (200, 100) & (action dim + 200, 100) or (200, 100) \\
    Activation & ReLU & ReLU \\
    \midrule
    Fully Connected & (100, action dim) & (100, 1) \\
    Activation & tanh & None \\
    \bottomrule
  \end{tabular}
  }
\end{table}

For DDSR, the factored critic (i.e., $Q$-function) consists of a representation network, a reconstruction network, a SR network and a linear reward vector, as described in the original paper of DSR.
The structure of the factored critic of DDSR is shown in Table \ref{table:ddsr_critic}.

\begin{table}[ht]
  \caption{The Network structure for the factored critic network of DDSR, including a representation network, a reconstruction network, a SR network and a reward vector.
  }
  \label{table:ddsr_critic}
  \centering
  \scalebox{1.0}{
  \begin{tabular}{ccc}
    \toprule
    Network & Layer & Structure \\
    \midrule
    Representation Network & Fully Connected & (state dim, 200) \\
     & Activation & ReLU \\
     & Fully Connected & (200, 100) \\
     & Activation & ReLU \\
     & Fully Connected & (100, representation dim) \\
     & Activation & None \\
    \midrule
    Reconstruction Network & Fully Connected & (representation dim, 100) \\
     & Activation & ReLU \\
     & Fully Connected & (100, 200) \\
     & Activation & ReLU \\
     & Fully Connected & (200, state dim) \\
     & Activation & None \\
    \midrule
    SR Network & Fully Connected & (representation dim, 200) \\
     & Activation & ReLU \\
     & Fully Connected & (200, 100) \\
     & Activation & ReLU \\
     & Fully Connected & (100, representation dim) \\
     & Activation & None \\
    \midrule
    Linear Reward Vector & Fully Connected (not use bias) & (representation dim, 1) \\
     & Activation & None \\
    \bottomrule
  \end{tabular}
  }
\end{table}

For VD-DDPG, the decomposed critic (i.e., $Q$-function) consists of a convolutional representation network $f^{\rm CNN}$, a linear trajectory return network $U^{\rm Linear}$ and a conditional VAE (an encoder network and a decoder network).
The structure of the decomposed critic of VD-DDPG is shown in Table \ref{table:vdfp_critic}.

\begin{table}[ht]
  \caption{The Network structure for the factored critic network of VD-DDPG, including a convolutional representation network $f^{\rm CNN}$, a linear trajectory return network $U^{\rm Linear}$ and a conditional VAE (an encoder network and a decoder network).
  }
  \label{table:vdfp_critic}
  \centering
  \scalebox{1}{
  \begin{tabular}{ccc}
    \toprule
    Network & Layer (Name) & Sturcture \\
    \midrule
    Representation Network & Convolutional & filters with height $\in \{1,2,4,8,16,32,64\}$ \\
    $f^{\rm CNN}(\tau_{t:t+k})$ & & of numbers $ \{20, 20, 10, 10, 5, 5, 5\}$ \\
     & Activation & ReLU \\
     & Pooling (concat) & Maxpooling \& Concatenation \\
     & Fully Connected (highway) & (filter num, filter num) \\
     & Joint & Sigmoid(highway) $\cdot$ ReLU(highway)\\
     & & + (1 - Sigmoid(highway)) $\cdot$ ReLU(concat)\\
     & Dropout & Dropout(drop\_prob = 0.2) \\
     & Fully Connected & (filter num, representation dim) \\
     & Activation & None \\
    \midrule
    Conditional Encoder Network & Fully Connected (main) & (representation dim, 400) \\
    $q_{\phi}(z_t|m_{t:t+k}, s_t, a_t)$ & Fully Connected (encoding) & (state dim + action dim, 400) \\
     & Pairwise-Product & Sigmoid(encoding) $\cdot$ ReLU(main) \\
     & Fully Connected & (400, 200) \\
     & Activation & ReLU \\
     & Fully Connected (mean) & (200, $z$ dim) \\
     & Activation & None \\
     & Fully Connected (log\_std) & (200, $z$ dim) \\
     & Activation & None \\
    \midrule
    Conditional Decoder Network & Fully Connected (latent) & (z dim, 200) \\
    $p_{\varphi}(m_{t:t+k}| z_t, s_t, a_t)$ & Fully Connected (decoding) & (state dim + action dim, 200) \\
     & Pairwise-Product & Sigmoid(decoding) $\cdot$ ReLU(latent) \\
     & Fully Connected & (200, 400) \\
     & Activation & ReLU \\
     & Fully Connected (reconstruction) & (400, representation dim) \\
     & Activation & None \\
    \midrule
    Trajectory Return Network & Fully Connected & (representation dim, 1) \\
    $U^{\rm Linear}(m_{t:t+k})$ & Activation & None \\
    \bottomrule
  \end{tabular}
  }
\end{table}

In our ablation studies (Table \ref{table:ablation} and Figure \ref{figure:ablation}), for VDFP\_MLP, we also use a two-layer feed-forward neural network of 200 and 100 hidden units with ReLU activation (except for the output layer) for $P^{\rm MLP}$.
For VDFP\_LSTM, we use one LSTM layer with 100 units to replace the convolutional layer (along with maxpooling layer) as described in Table \ref{table:vdfp_critic}.
For VDFP\_Concat, we concatenate the state, action and the representation (or latent variable) rather than a pairwise-product structure.
For VDFP\_ReLU, we add an ReLU-activated fully-connected layer with 50 units in front of the linear layer for trajectory return model. 

\subsection{Hyperparameter}

For all our experiments, we use the raw observation and reward from the environment and no normalization or scaling are used.
No regularization is used for the actor and the critic in all algorithms.
Table \ref{table:hyperparameter} shows the common hyperparamters of algorithms used in all our experiments.
For VD-DDPG and DDSR, critic learning rate denotes the learning rate of the conditional VAE and the successor representation model respectively. 
Return (reward) model learning rate denotes the learning rate of the return model (along with the representation model) for VD-DDPG and the learning rate of the immediate reward vector for DDSR. 

\begin{table}[ht]
  \caption{A comparison of common hyperparameter choices of algorithms.
  We use `-' to denote the `not applicable' situation.
  }
  \label{table:hyperparameter}
  \centering
  \scalebox{1.0}{
  \begin{tabular}{c|c|cc|cc}
    \toprule
    Hyperparameter & VD-DDPG & DDSR & DDPG & PPO & A2C\\
    \midrule
    Actor Learning Rate & 2.5$\cdot$10$^{-4}$ & 2.5$\cdot$10$^{-4}$ & 10$^{-4}$ & 10$^{-4}$ & 10$^{-4}$ \\
    Critic (VAE, SR) Learning Rate & 10$^{-3}$ & 10$^{-3}$ & 10$^{-3}$ & 10$^{-3}$ & 10$^{-3}$ \\
    Return (Reward) Model Learning Rate & 5$\cdot$10$^{-4}$ & 5$\cdot$10$^{-4}$ & - & - & - \\
    \midrule
    Discount Factor & 0.99 & 0.99 & 0.99 & 0.99 & 0.99 \\
    Optimizer & Adam & Adam & Adam & Adam & Adam\\
    Target Update Rate & - & 10$^{-3}$ & 10$^{-3}$ & - & -\\
    Exploration Policy & $\mathcal{N}(0, 0.1)$ & $\mathcal{N}(0, 0.1)$ & $\mathcal{N}(0, 0.1)$ & None & None\\
    Batch Size & 64 & 64 & 64 & 256 & 256 \\
    Buffer Size & 10$^{5}$ & 10$^{5}$ & 10$^{5}$ & - & - \\
    Actor Epoch & - & - & - & 10 & 10 \\
    Critic Epoch & - & - & - & 10 & 10 \\
    \bottomrule
  \end{tabular}
  }
\end{table}

\subsection{Additional Implementation Details}

\subsubsection{Training setup.}
For DDPG, the actor network and the critic network is updated every 1 time step.
We implement the DDSR based on the DDPG algorithm, by replacing the original critic of DDPG with the factored $Q$-function as described in the DSR paper.
The actor, along with all the networks described in Table \ref{table:ddsr_critic} are updated every 1 time step.
The representation dimension is set to 100.
Before the training of DDPG and DDSR, we run 10000 time steps for experience collection, which are also counted in the total time steps.

For PPO and A2C, we use Generalized Advantage Estimation (GAE) \cite{Schulman2016GAE} with $\lambda = 0.95$ for stable policy gradient.
The clip range of PPO algorithm is set to 0.2.
The actor network and the critic network are updated every $2$ and $5$ episodes for HalfCheetah-v1 and Walker2d-v1 respectively, with the epoches and batch sizes described in Table \ref{table:hyperparameter}.

For VD-DDPG, we set the KL weight $\beta$ as 1000 and the clip value $c$ as 0.2.
The latent variable dimension ($z$ dim) is set to 50 and the representation dimension is set to 100.
We collect trajectories experiences in the first 5000 time steps and then pre-train the trajectory model (along with the representation model) and the conditional VAE for 15000 time steps, after which we start the training of the actor.
All the time steps above are counted into the total time steps for a fair comparison.
The trajectory return model (along with the representation model) is trained every 10 time steps for the pre-train process and is trained every 50 time steps in the rest of training, which already ensures a good performance in our experiments.
The actor network and the conditional VAE are trained every 1 time step.

\subsubsection{Trajectory length setup.}
We consider a max trajectory length $l$ for VD-DDPG in our experiments.
For example, a max length $l$ = 256 can be considered to correspond a discounted factor $\gamma = 0.99$ as $0.99^{256} \approx 0.076$.
In practice, for a max length $l > 64$, we add an additional fully-connected layer with ReLU activation before the convolutional representation model $f^{\rm CNN}$, to aggregate the trajectory into the length of 64.
This is used for the purpose of reducing the time cost and accelerating the training of the convolutional representation model as described in Table \ref{table:vdfp_critic}.
For example, we feed every 4 state-action pairs of a trajectory with length 256 into the aggregation layer to obtain an aggregated trajectory with the length 64, and then feed the aggregated trajectory to $f^{\rm CNN}$ for a trajectory representation.
Other ways to process long-trajectory with high efficiency can also be considered.

We train $U$ with trajectory segments from the full episode with varying length from 1 (the extreme case at the tail) to the max trajectory length (e.g., 256) we used. 
Actually, we once tried to add additional prediction of several $k$-step (e.g., $k$ = 8, 16, 32) accumulated reward 
during the training of $U$ and we did not find apparent difference in final results. 

\subsubsection{Attempts on LSTM variants.}
As LSTM is a popular choice for sequence, 
we first spent a lot of time evaluating and tuning LSTM, 
reverse LSTM (since steps near current time may need to be weighted more), 
GRU and MLP-Agg (i.e., transform each s-a pair with MLP and then perform aggregation among pairs like mean-reduce). 
We then tried CNN for sentence representation in \cite{Kim2014CNN} 
and found it empirically performs better and faster without too much tuning. 

\section{Complete Learning Curves and Additional Experiments}
\label{supp:complete_curves}

\subsection{Learning Curves for the Results in Ablation}
\begin{figure}[ht]
\centering
\subfigure[VAE v.s. MLP]{
\includegraphics[width=0.34\textwidth]{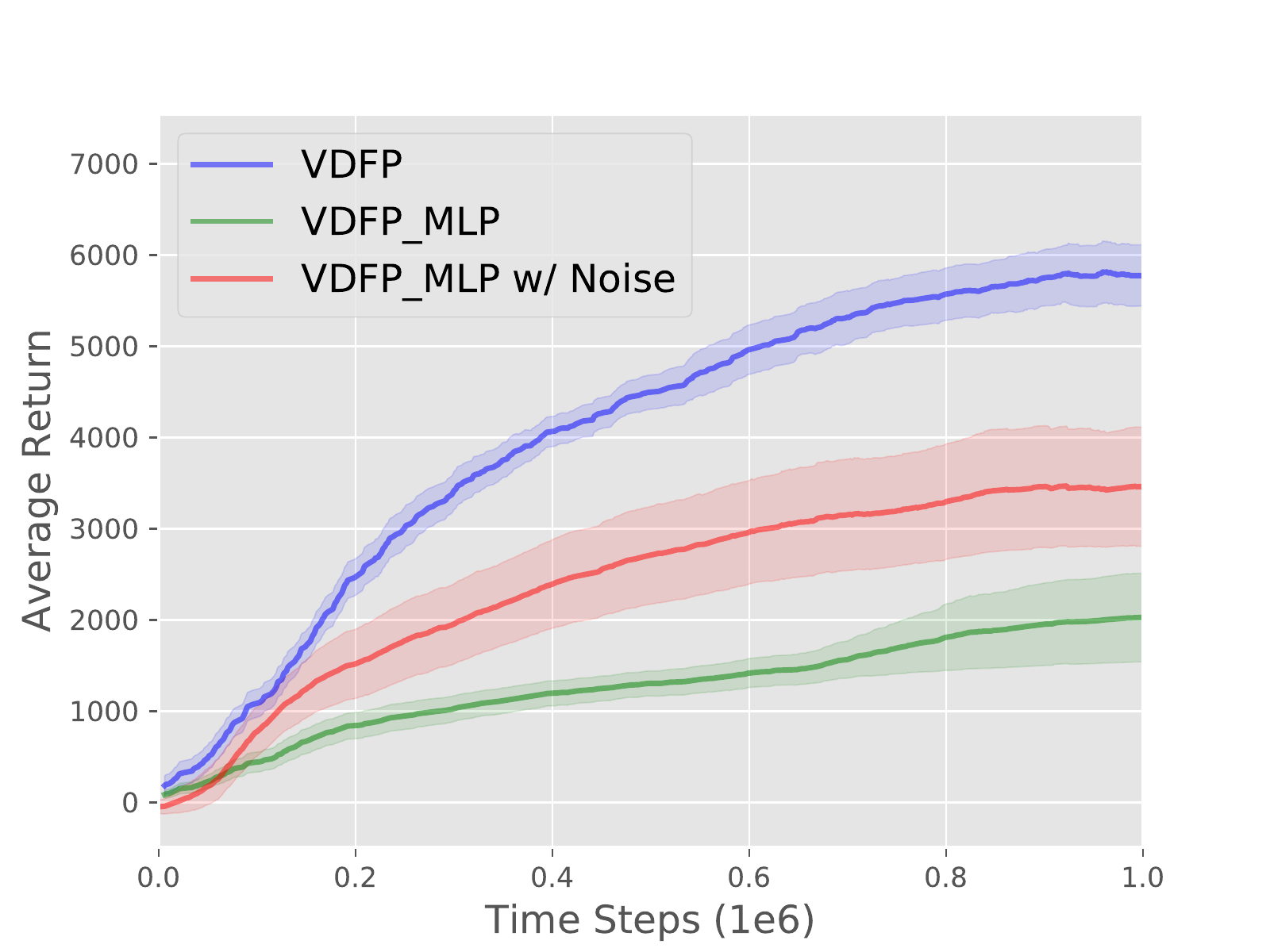}
}
\hspace{-0.8cm}
\subfigure[CNN v.s. LSTM]{
\includegraphics[width=0.34\textwidth]{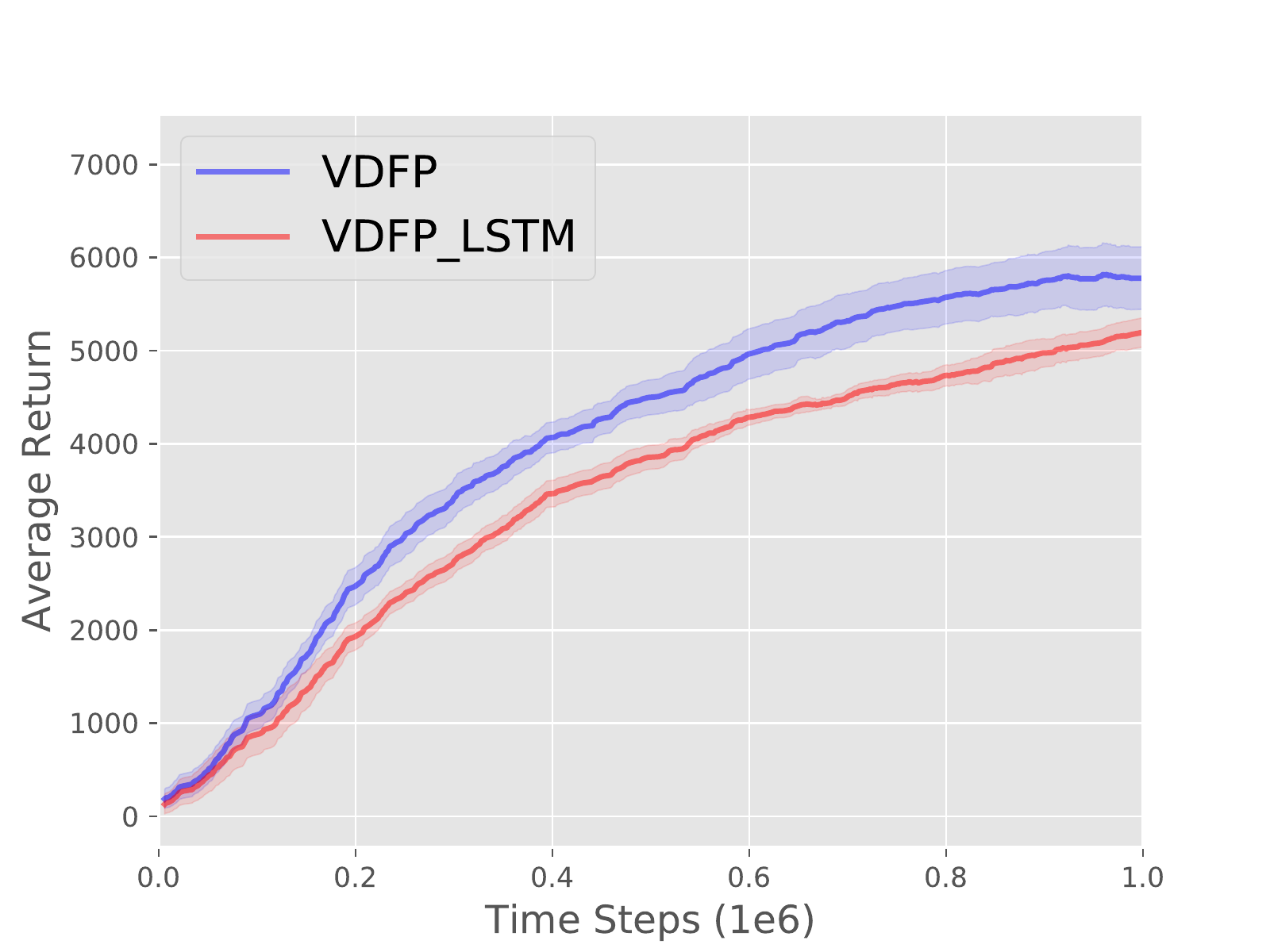}
}
\hspace{-0.8cm}
\subfigure[Pairwise-Prod. v.s. Concat.]{
\includegraphics[width=0.34\textwidth]{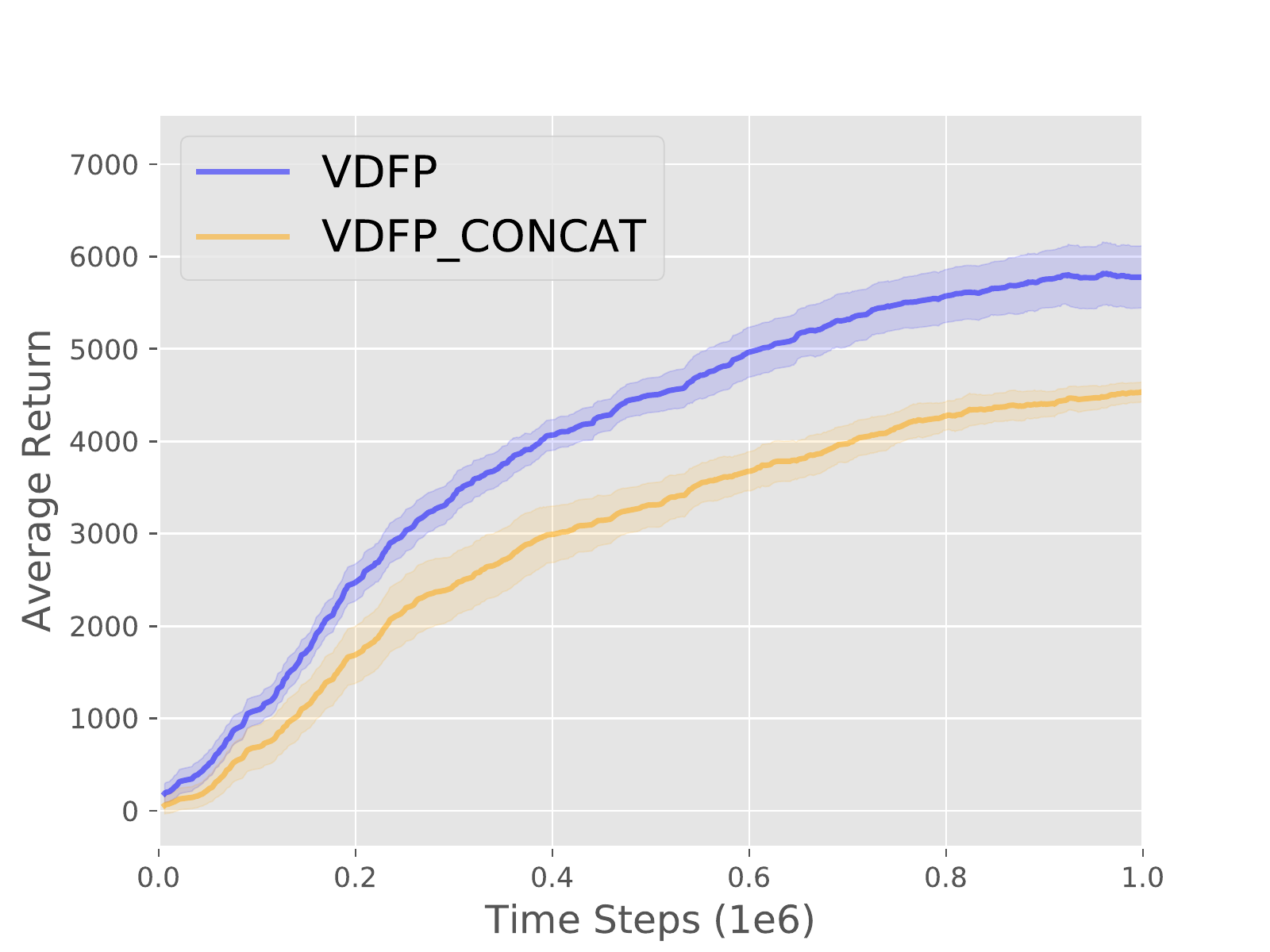}
}
\hspace{-1cm}
\subfigure[Linear v.s. ReLU]{
\includegraphics[width=0.25\textwidth]{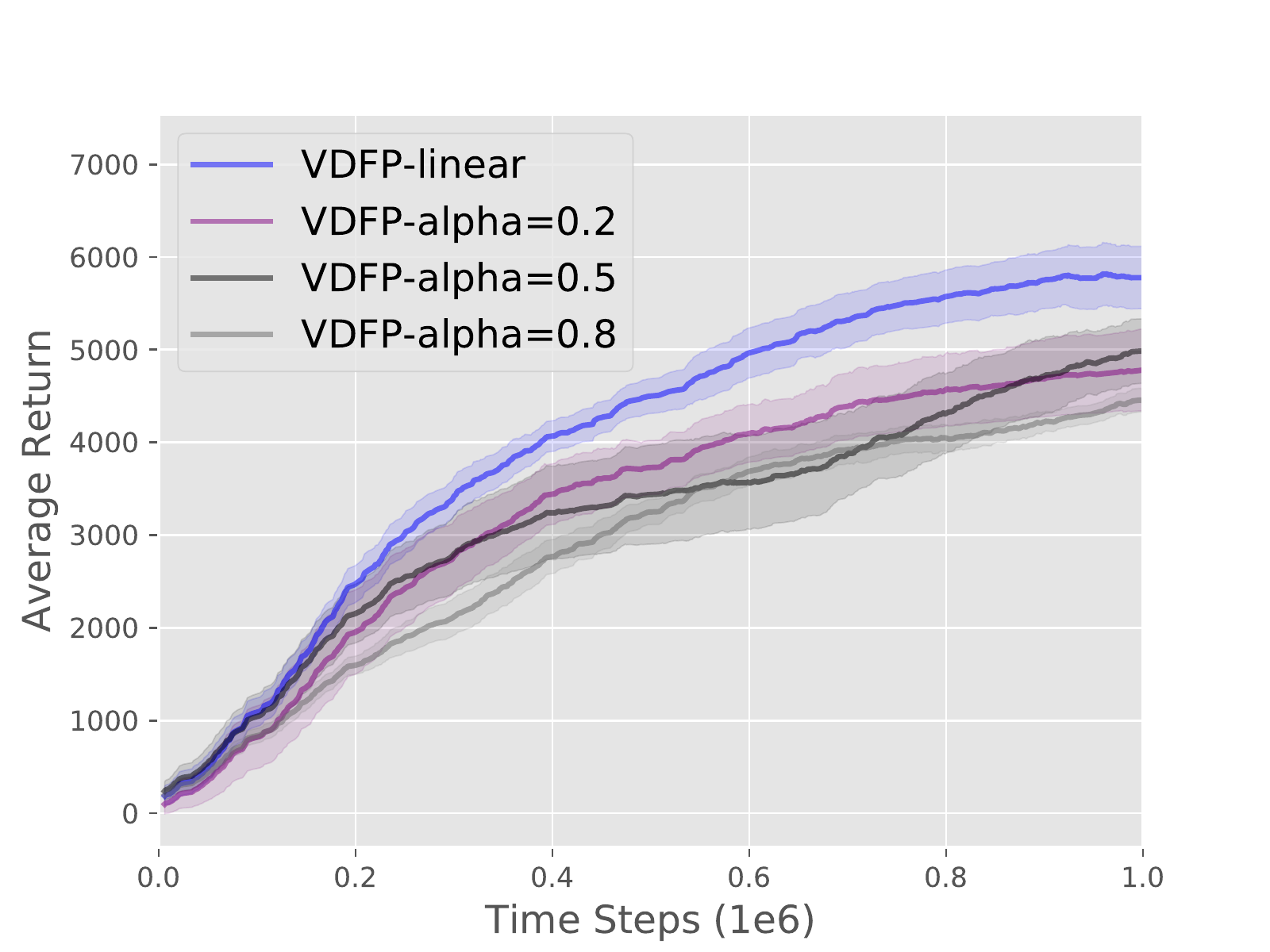}
}
\hspace{-0.6cm}
\subfigure[Linear v.s. Advanced Convex]{
\includegraphics[width=0.25\textwidth]{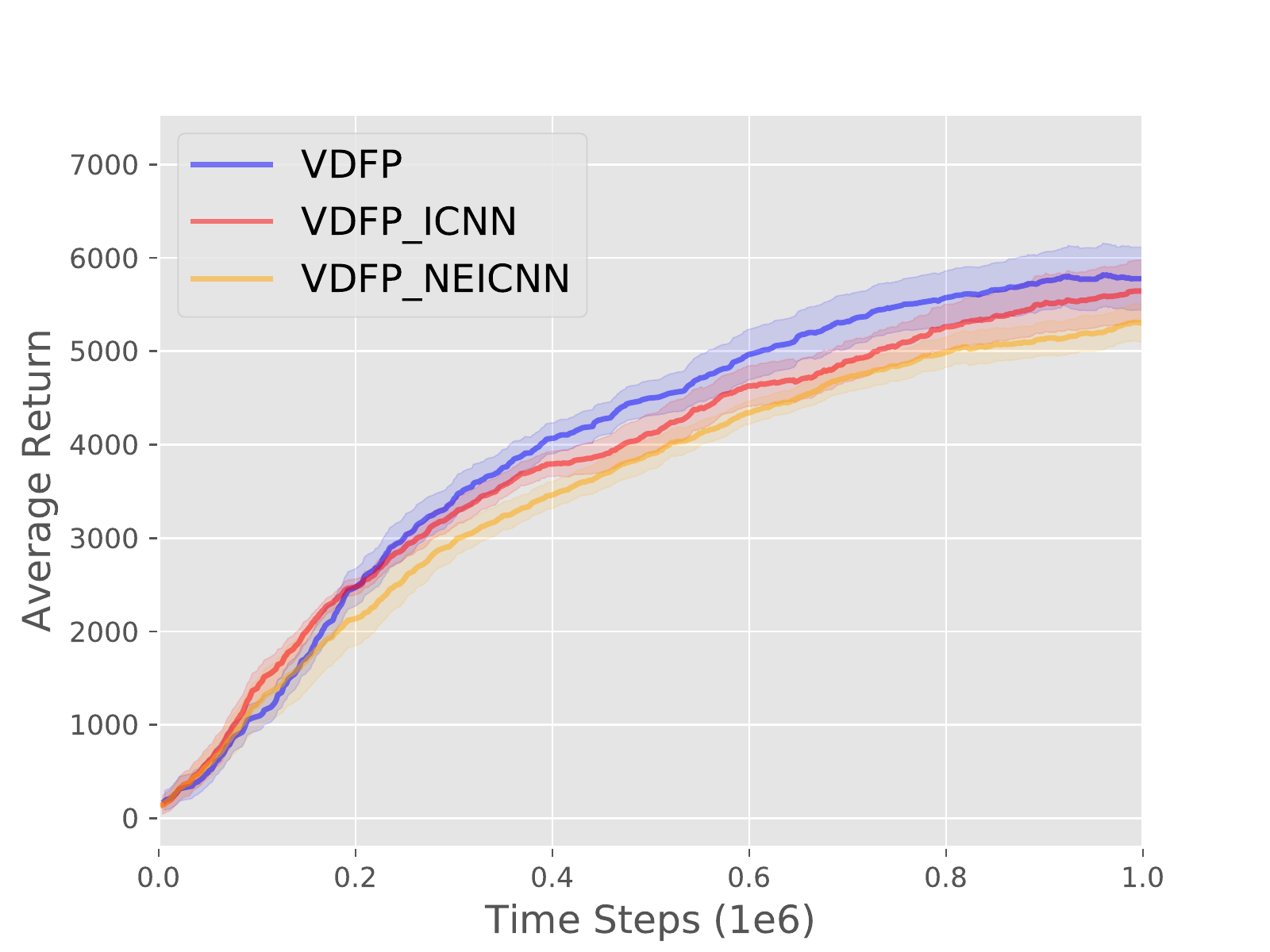}
}
\hspace{-0.6cm}
\subfigure[KL weights]{
\includegraphics[width=0.25\textwidth]{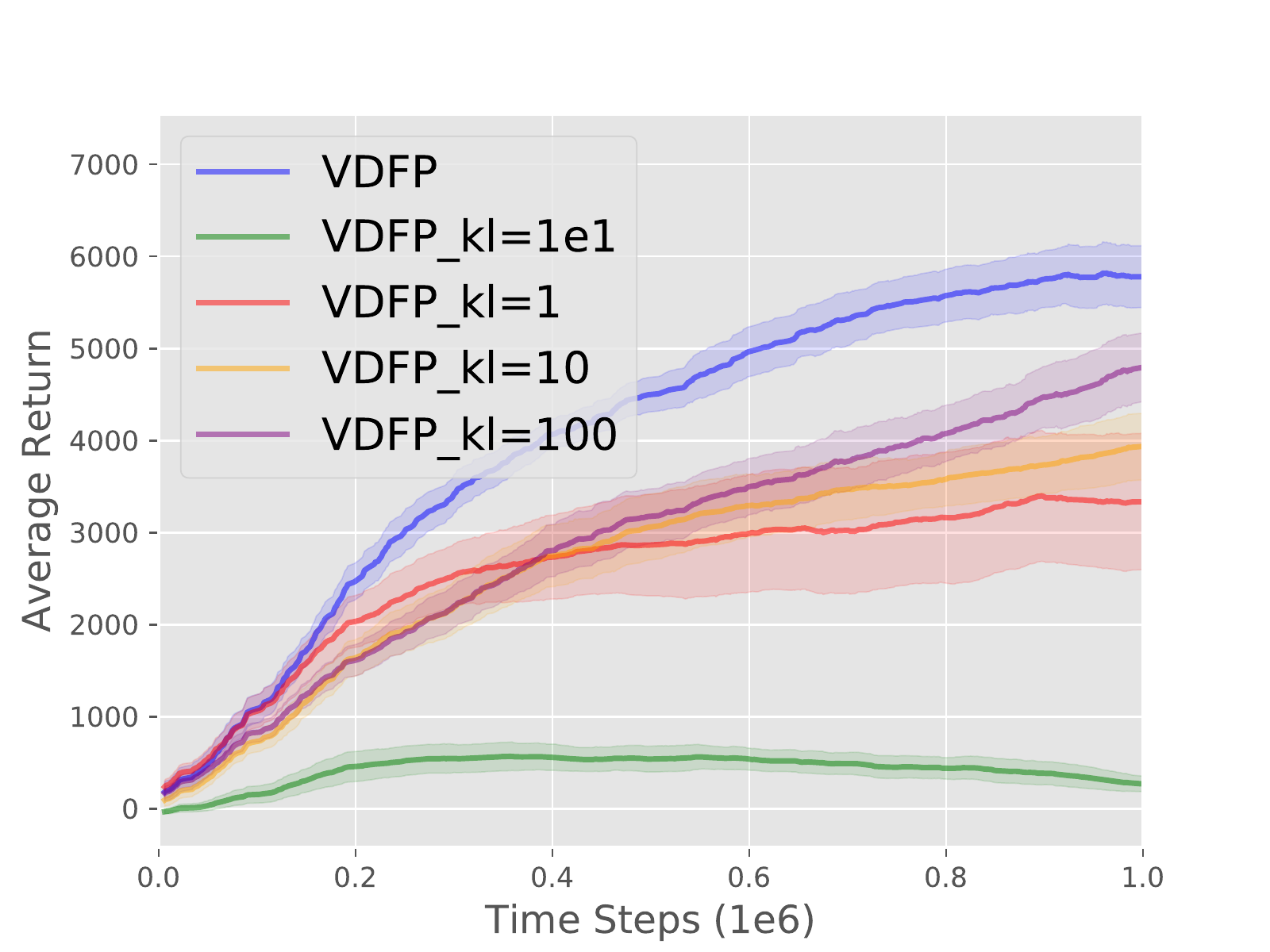}
}
\hspace{-0.6cm}
\subfigure[Clip value]{
\includegraphics[width=0.25\textwidth]{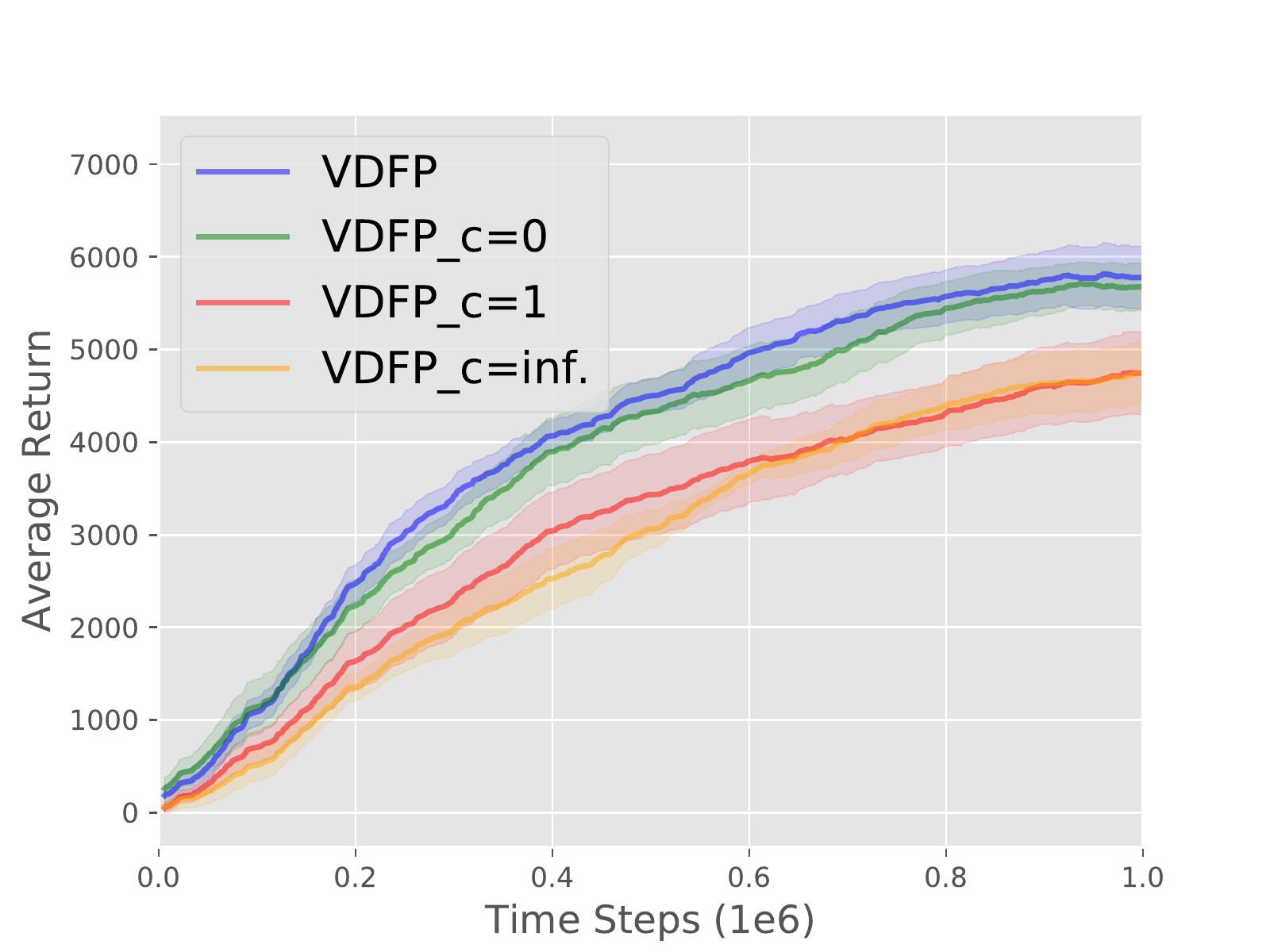}
}

\caption{Learning curves of ablation studies for VDFP (i.e., VAE + CNN + Pairwise-Product + linear layer + kl=1000 + c=0.2) in HalfCheetah-v1.
Here VDFP also equals to VD-DDPG.
The shaded region denotes half a standard deviation of average evaluation over 5 trials. 
Results are smoothed over recent 100 episodes.}
\label{figure:ablation}
\end{figure}

Figure \ref{figure:ablation} shows the learning curves of VDFP and its variants for ablation studies (Section \ref{section:ablation}), corresponding to the results in Table \ref{table:ablation}. 
VDFP equals to VD-DDPG in Figure \ref{figure:ablation}.

\subsection{Learning Curves for the Results in Delayed Reward Settings (Table \ref{table:delay_reward})}
We consider two representative delayed reward settings in real-world scenarios:
\begin{itemize}
\item multi-step accumulated rewards are given at sparse time steps;
\item each one-step reward is delayed for certain time steps.
\end{itemize}
To simulate above two settings, we make a simple modification to MuJoCo tasks respectively:
\begin{itemize}
\item deliver $d$-step accumulated reward every $d$ time steps or at the end of an episode;
\item delay the immediate reward of each step by $d$ steps and compensate at the end of episode.
\end{itemize}

\begin{figure}
\centering
\hspace{-0.1cm}
\subfigure[delay step $d$ = 0]{
\includegraphics[width=0.21\textwidth]{appendix_figs/supp-Evaluation-HalfCheetah-v1.pdf}
}
\hspace{-0.6cm}
\subfigure[delay step $d$ = 16]{
\includegraphics[width=0.21\textwidth]{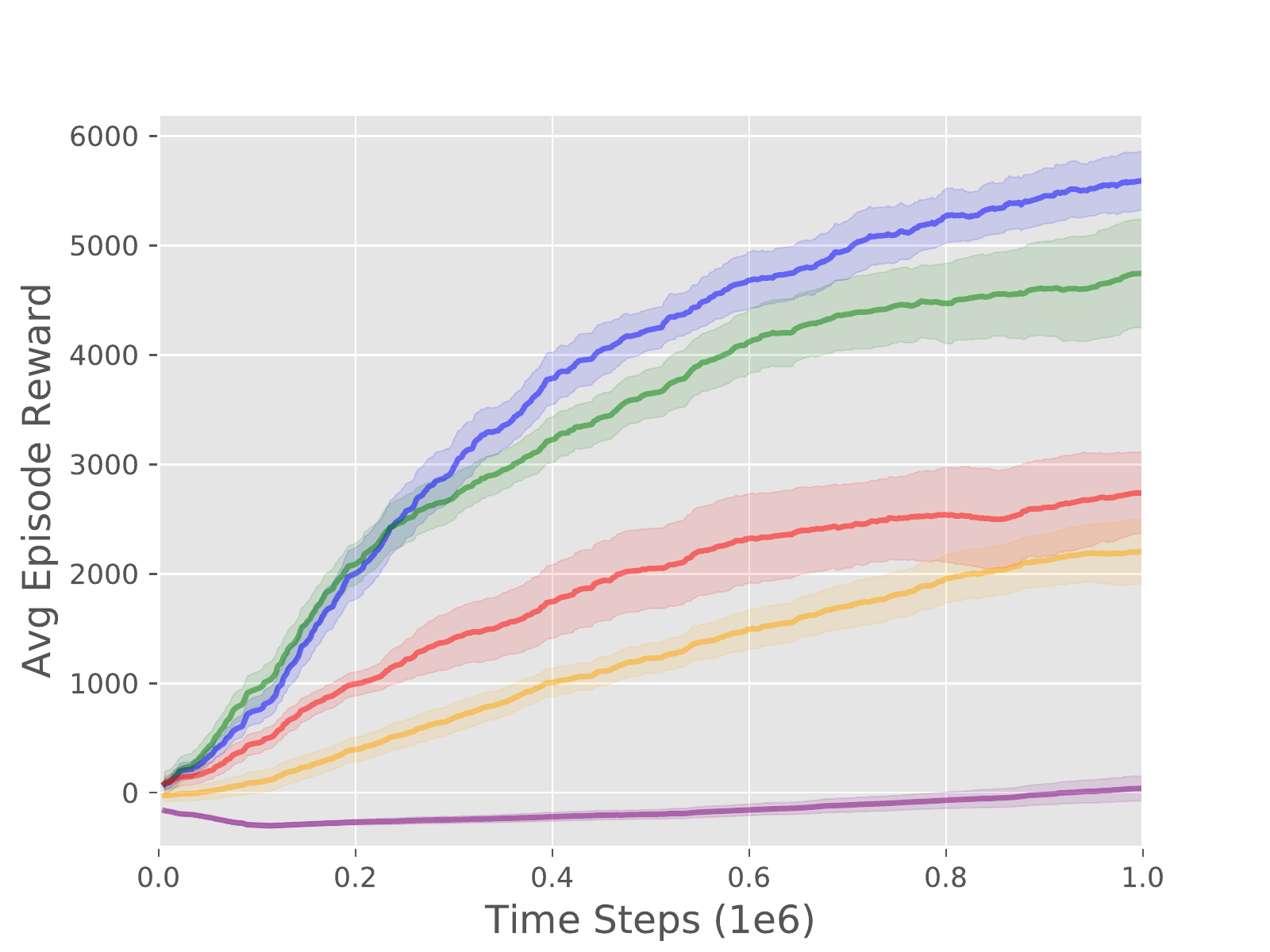}
}
\hspace{-0.6cm}
\subfigure[delay step $d$ = 32]{
\includegraphics[width=0.21\textwidth]{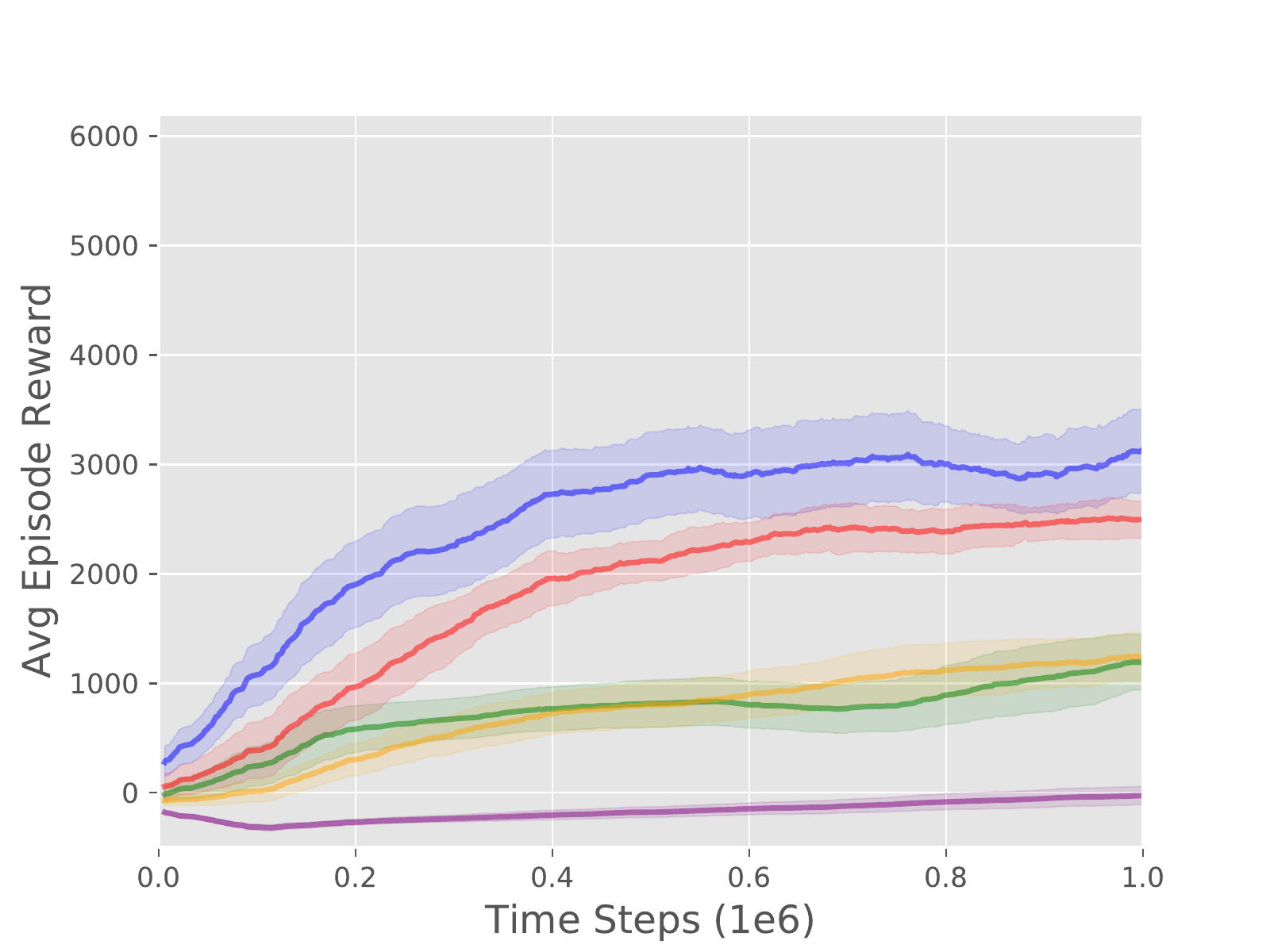}
}
\hspace{-0.6cm}
\subfigure[delay step $d$ = 64]{
\includegraphics[width=0.21\textwidth]{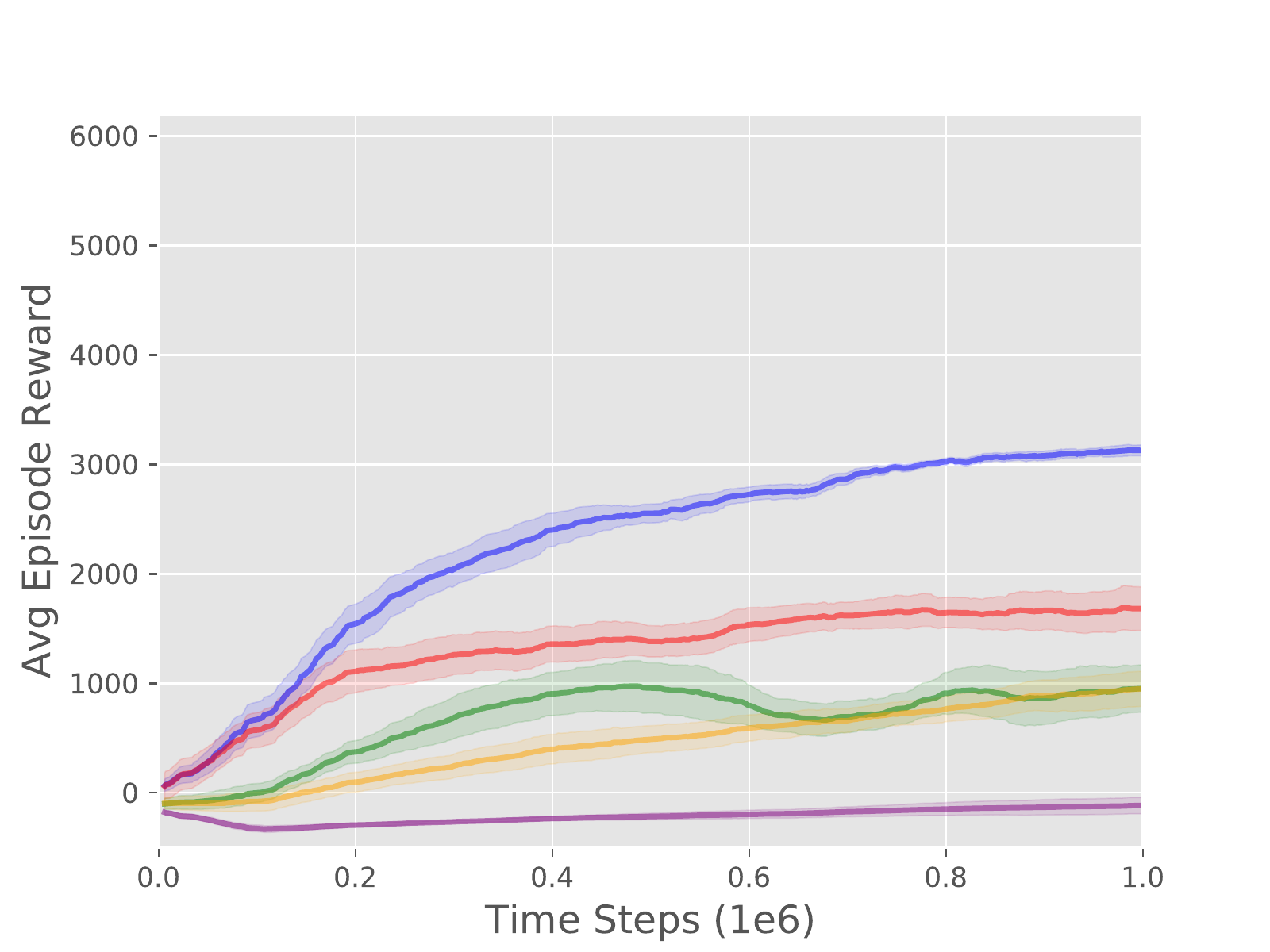}
}
\hspace{-0.6cm}
\subfigure[delay step $d$ = 128]{
\includegraphics[width=0.21\textwidth]{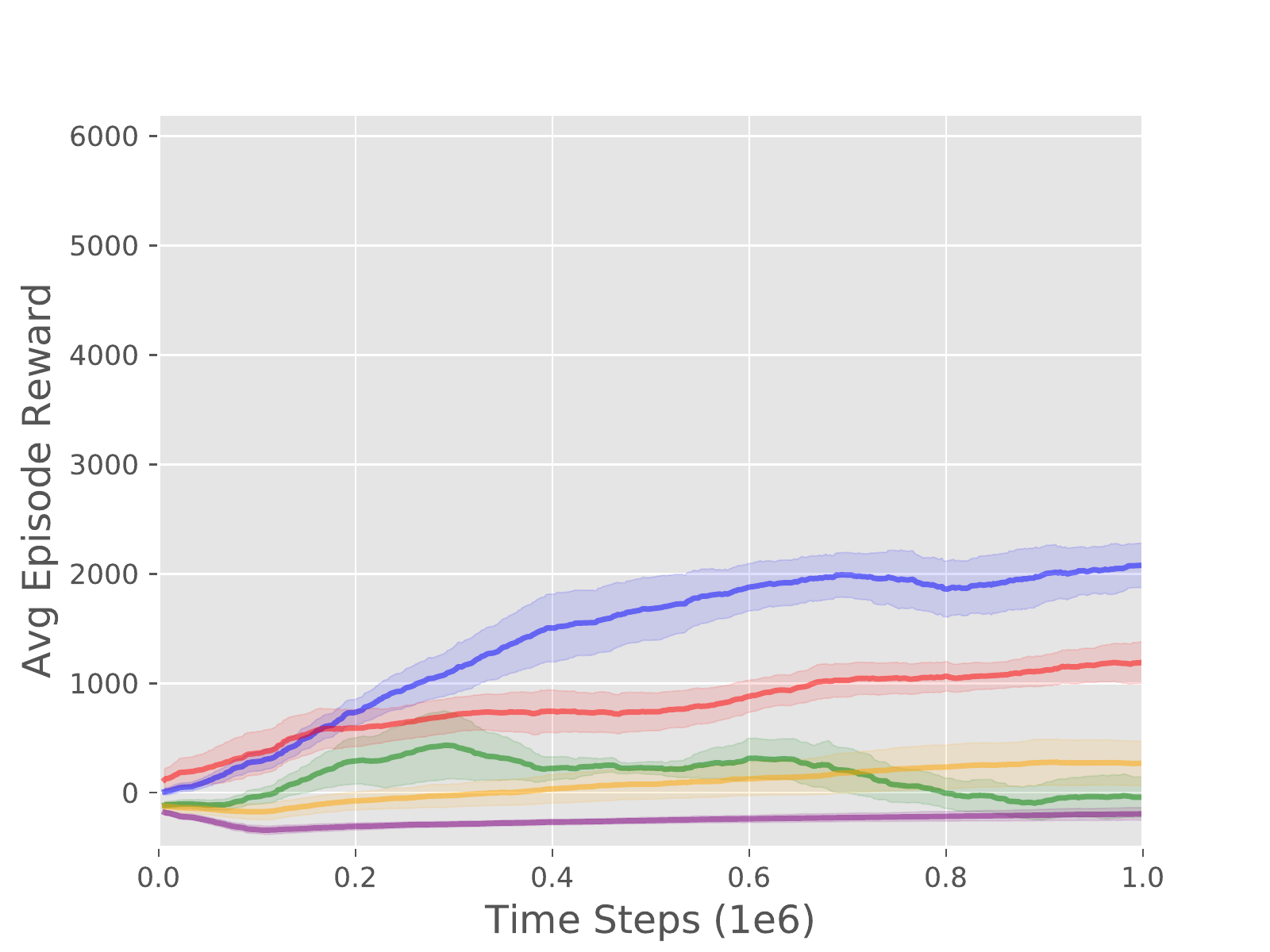}
}

\caption{Learning curves of algorithms in HalfCheetah-v1 under \emph{the second (2nd) delayed reward setting}. 
Different delay steps are listed from left to right.
The shaded region denotes half a standard deviation of average evaluation over 5 trials. 
Results are smoothed over recent 100 episodes.}
\label{figure:ds2-HalfCheetah}
\end{figure}

\begin{figure}
\centering
\hspace{-0.1cm}
\subfigure[delay step $d$ = 0]{
\includegraphics[width=0.21\textwidth]{appendix_figs/supp-Evaluation-Walker2d-v1.pdf}
}
\hspace{-0.6cm}
\subfigure[delay step $d$ = 16]{
\includegraphics[width=0.21\textwidth]{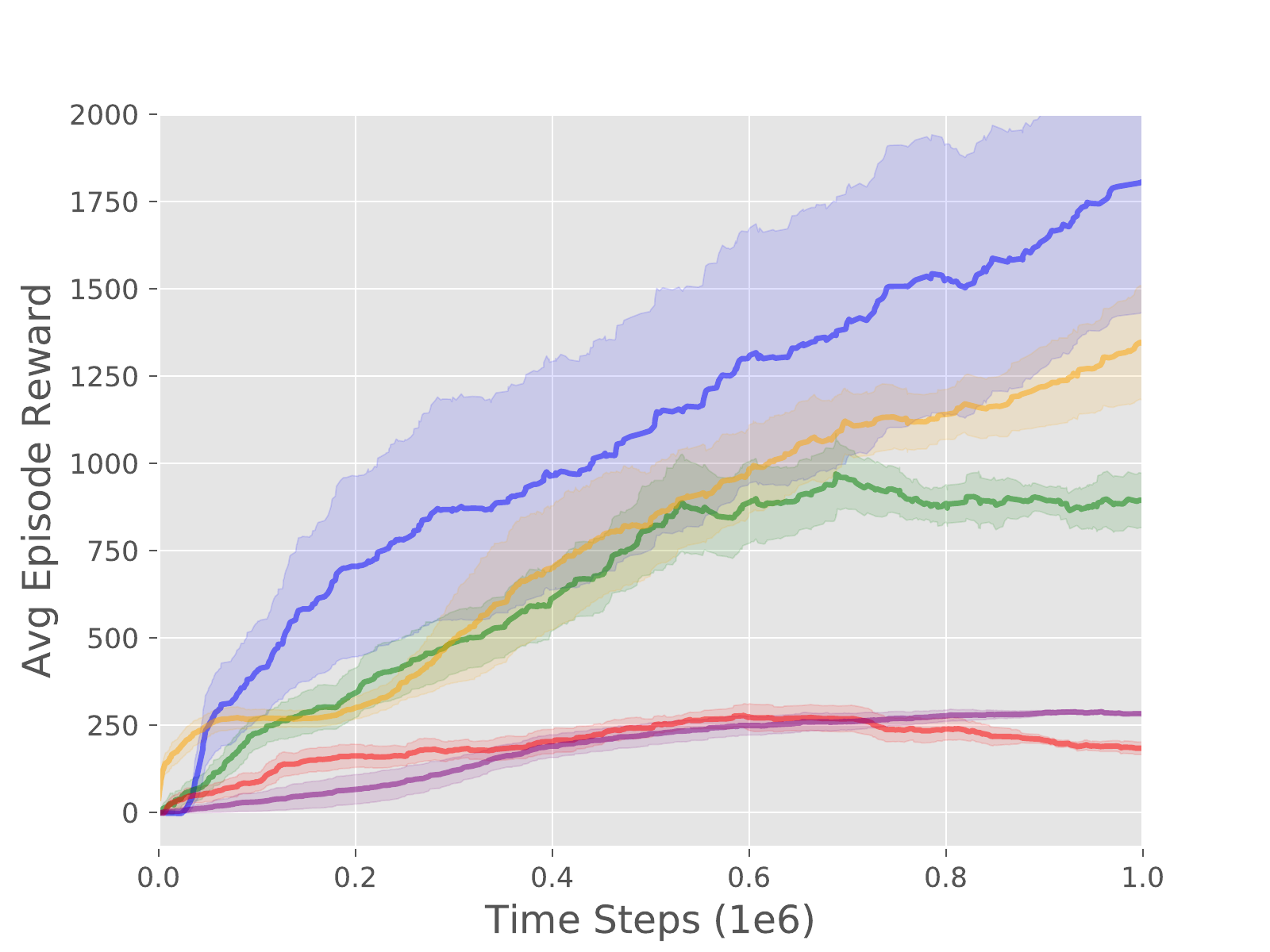}
}
\hspace{-0.6cm}
\subfigure[delay step $d$ = 32]{
\includegraphics[width=0.21\textwidth]{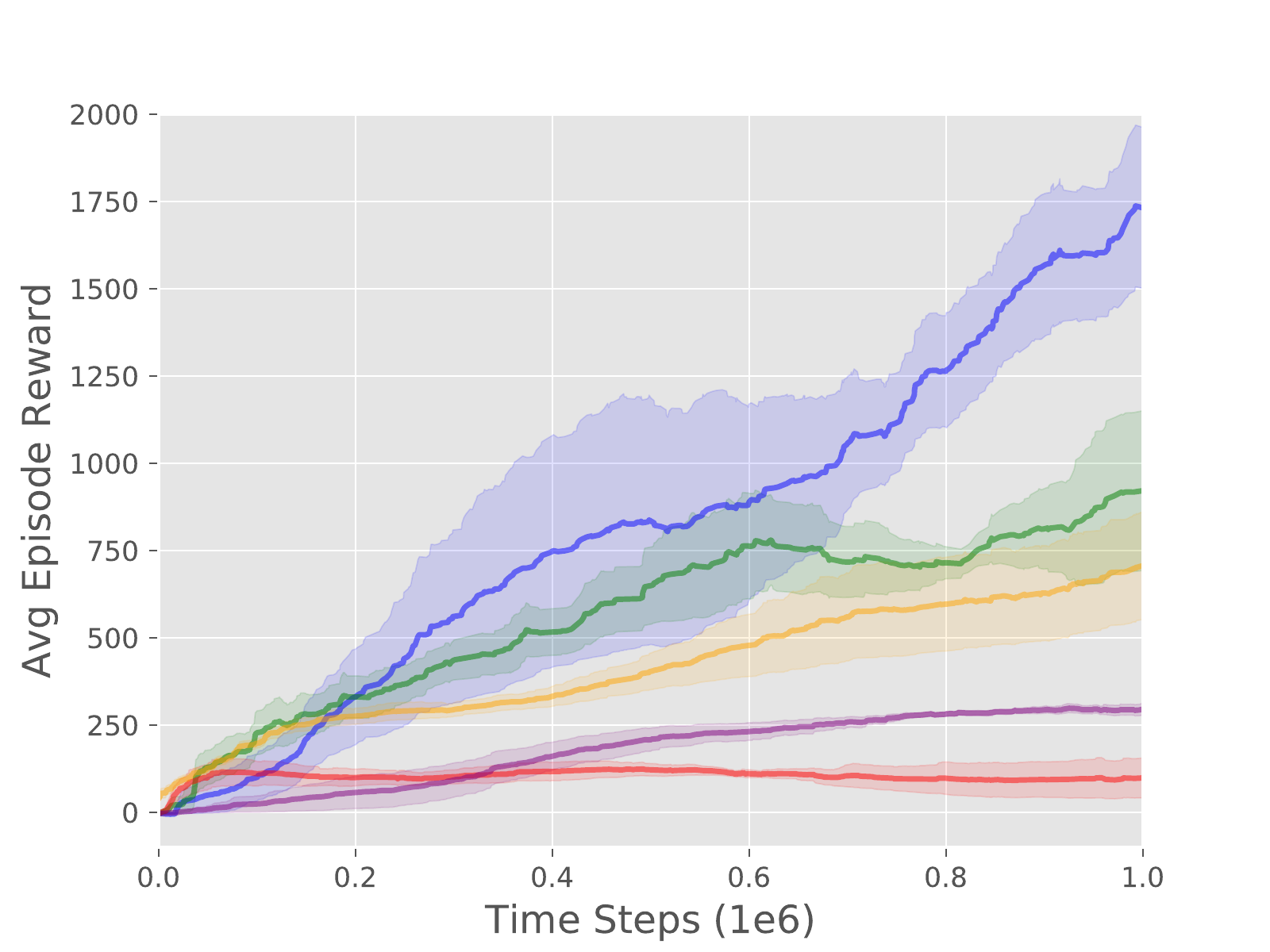}
}
\hspace{-0.6cm}
\subfigure[delay step $d$ = 64]{
\includegraphics[width=0.21\textwidth]{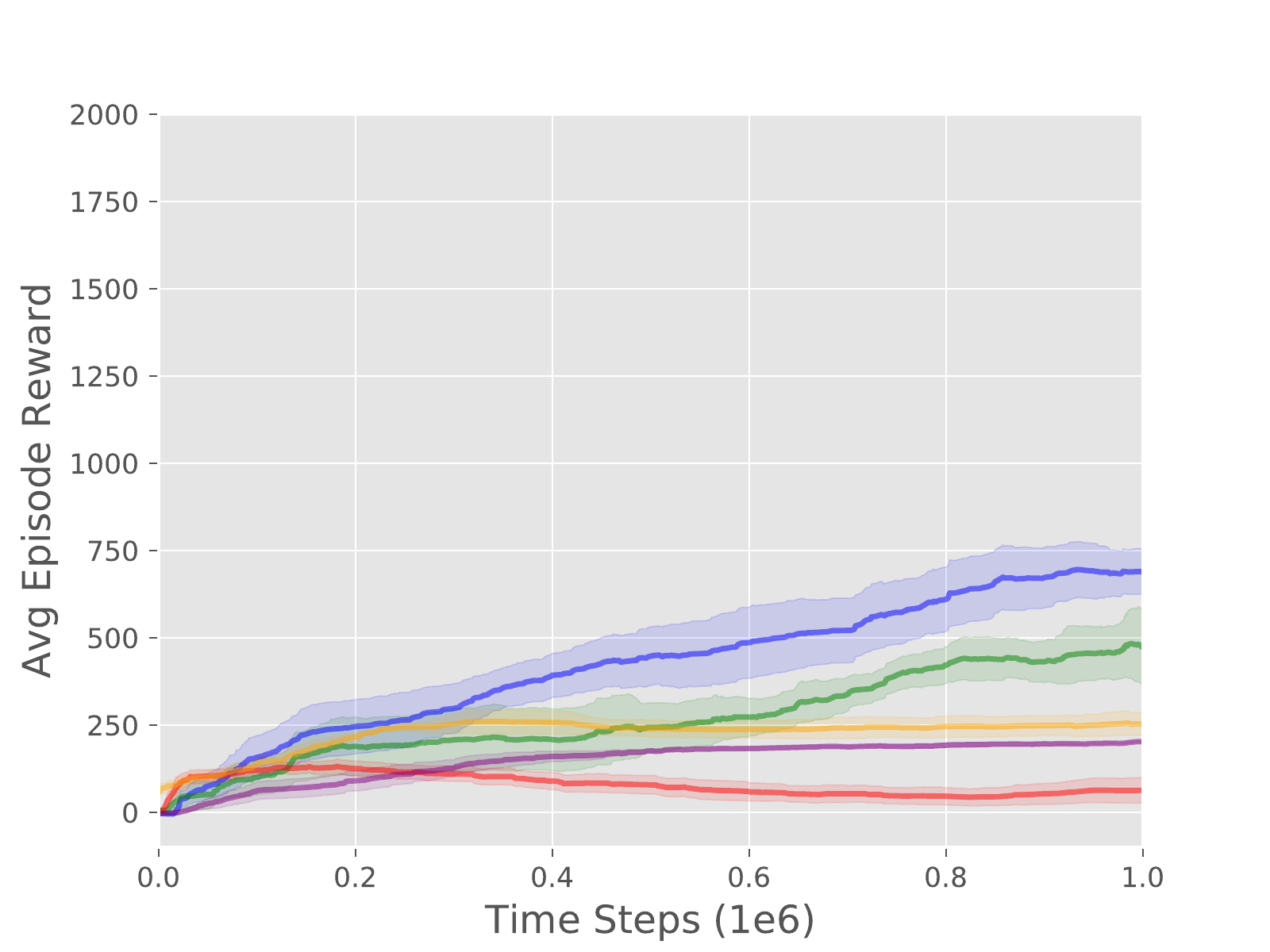}
}
\hspace{-0.6cm}
\subfigure[delay step $d$ = 128]{
\includegraphics[width=0.21\textwidth]{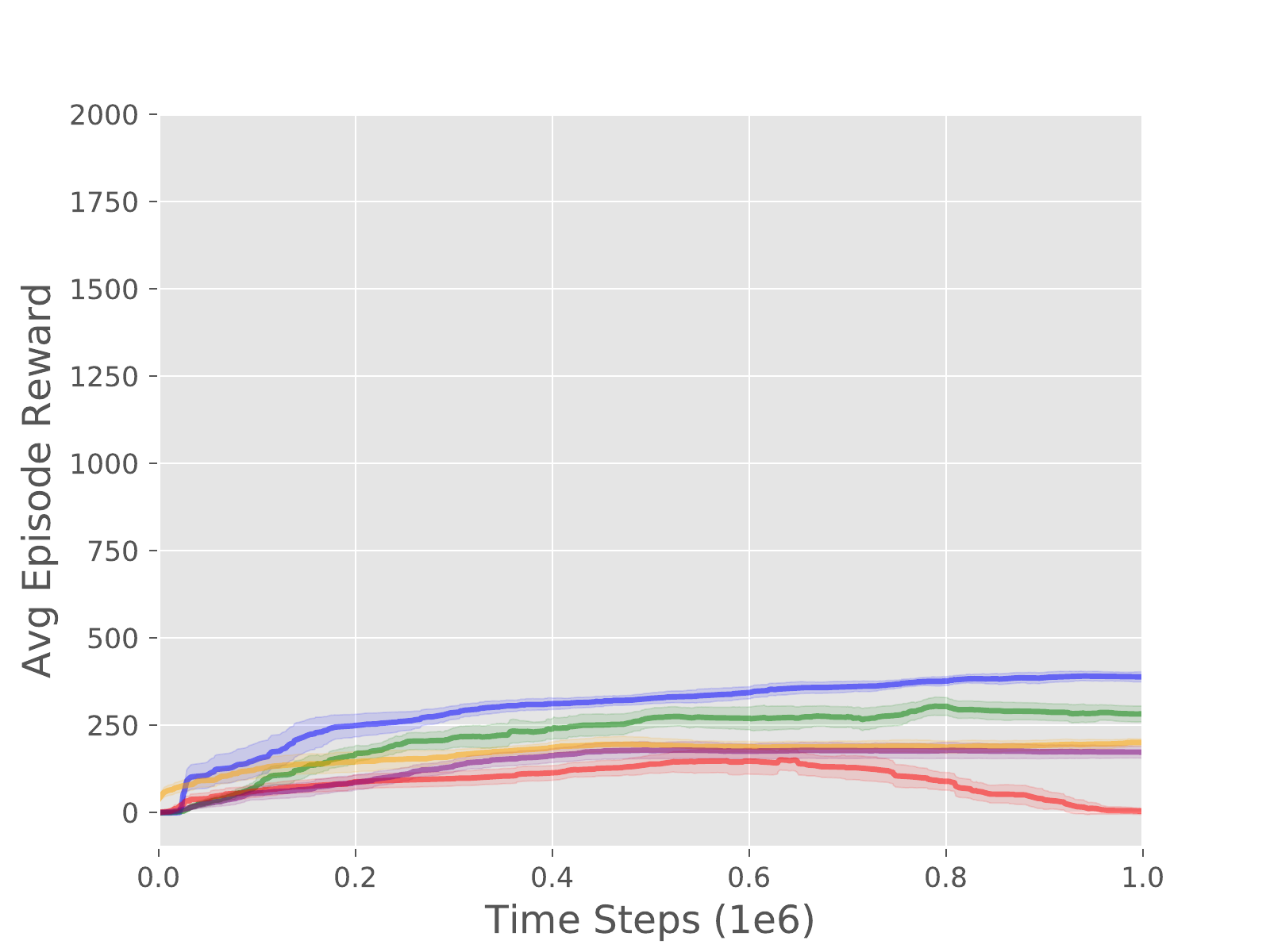}
}

\caption{Learning curves of algorithms in Walker2d-v1 under \emph{the second (2nd) delayed reward setting}. 
Different delay steps are listed from left to right.
The shaded region denotes half a standard deviation of average evaluation over 5 trials. 
Results are smoothed over recent 100 episodes.}
\label{figure:ds2-Walker2d}
\end{figure}

The complete learning curves of algorithms under the two delay reward settings are shown below.
Figure \ref{figure:ds1-HalfCheetah} and \ref{figure:ds1-Walker2d} shows the results for the first delay reward setting, and
Figure \ref{figure:ds2-HalfCheetah} and \ref{figure:ds2-Walker2d} shows the results for the second delay reward setting.
All algorithms gradually degenerate with the increase of delay step $d$.
VD-DDPG consistently outperforms others under all settings, and shows good robustness with delay step $d \le 64$.

In addition, we also found DDPG with n-step TD ($n = 2d$ to cover delay rewards) outperforms DDPG in delayed reward settings but is still much worse than VD-DDPG.

\begin{figure}
\centering
\hspace{-0.1cm}
\subfigure[delay step $d$ = 0]{
\label{subfig:vae_mlp}
\includegraphics[width=0.21\textwidth]{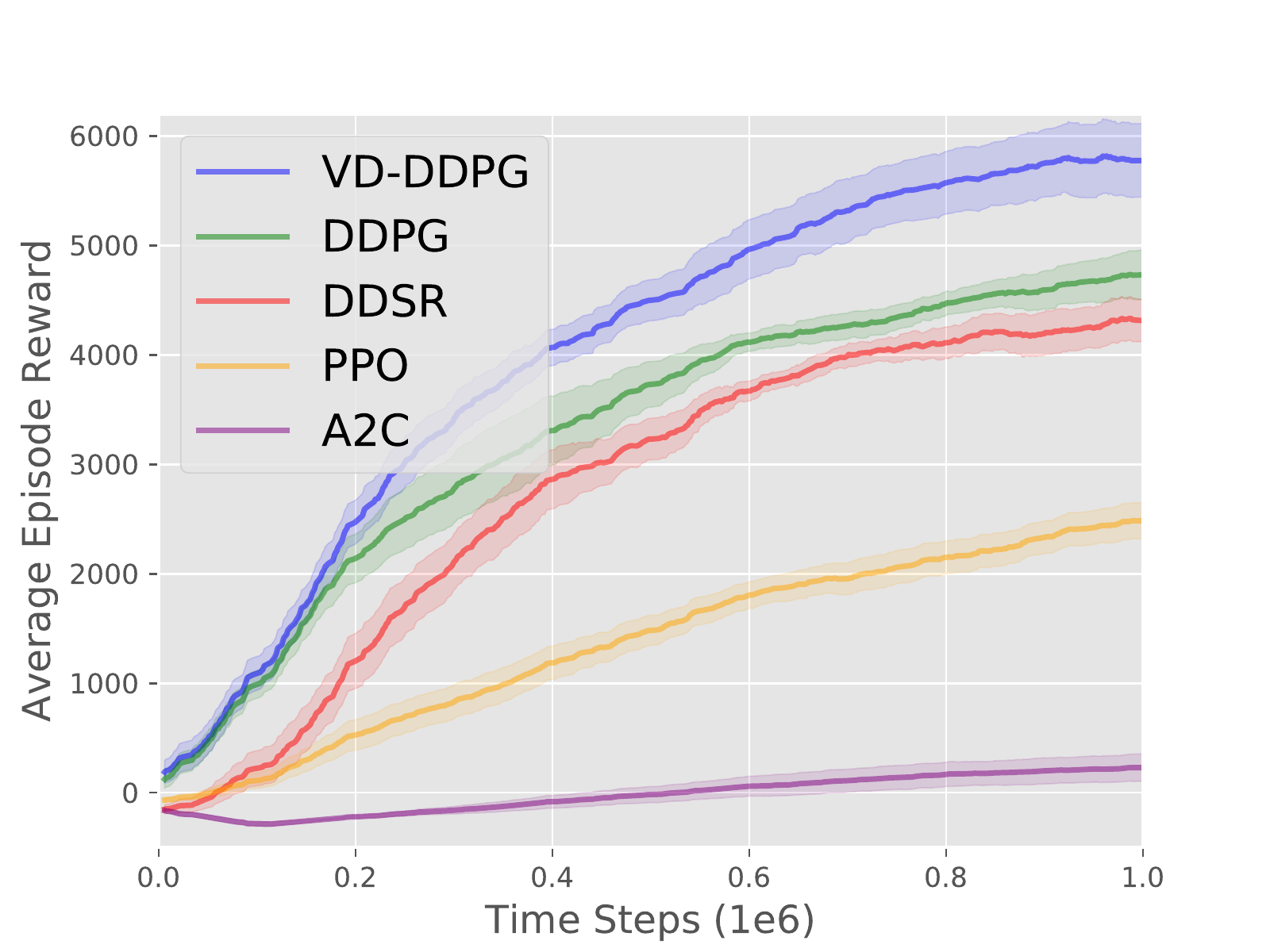}
}
\hspace{-0.6cm}
\subfigure[delay step $d$ = 16]{
\includegraphics[width=0.21\textwidth]{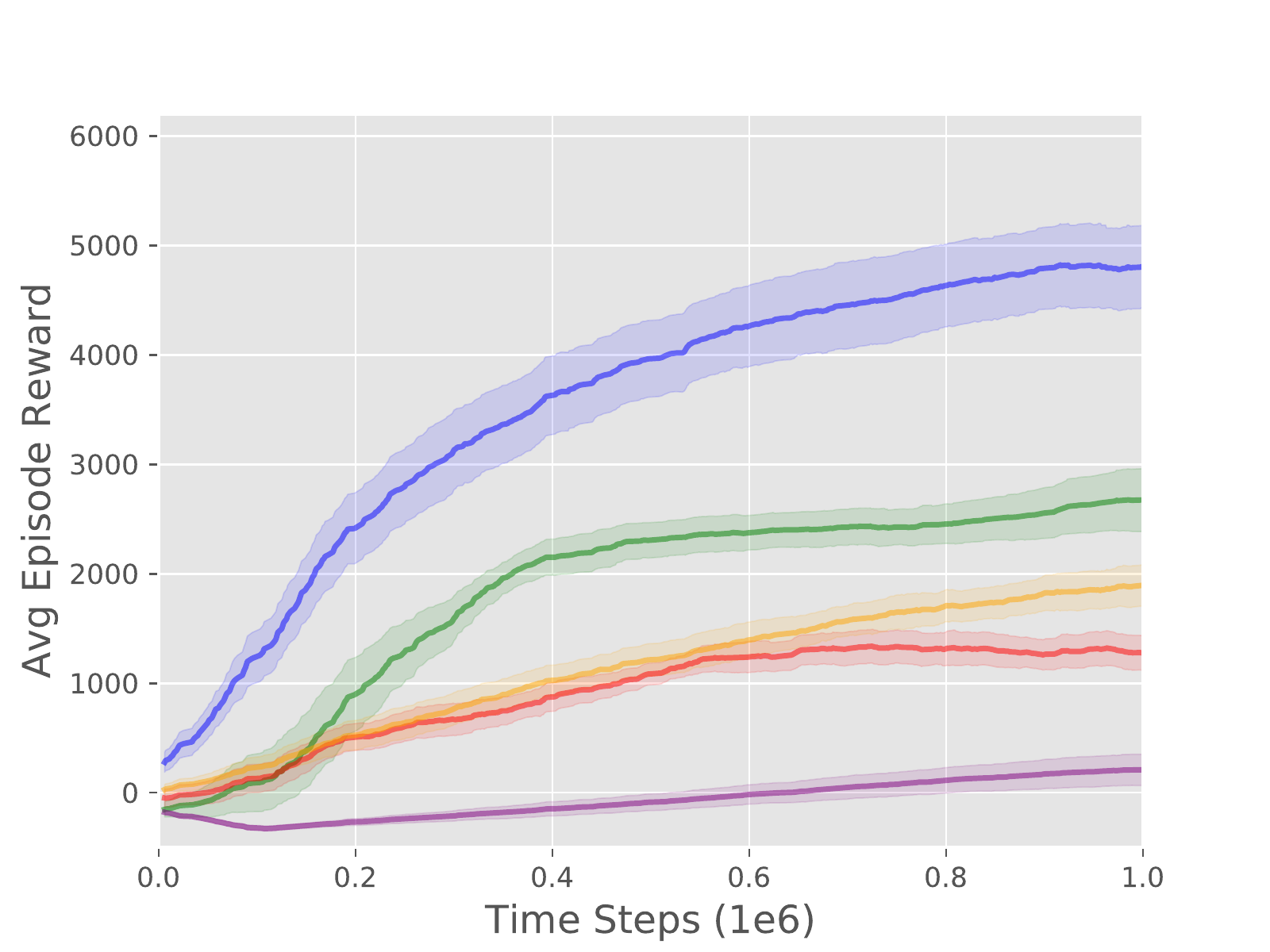}
}
\hspace{-0.6cm}
\subfigure[delay step $d$ = 32]{
\includegraphics[width=0.21\textwidth]{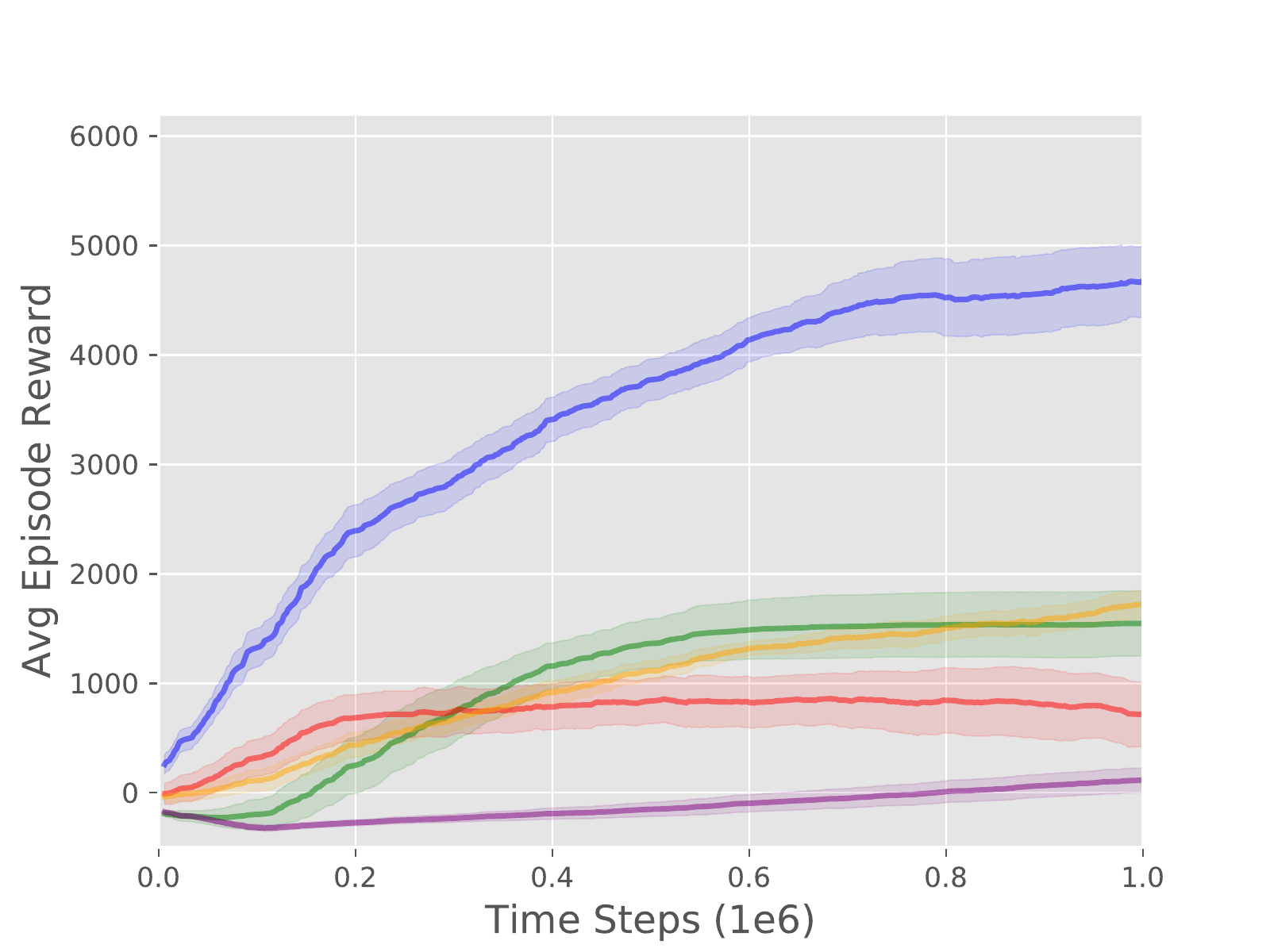}
}
\hspace{-0.6cm}
\subfigure[delay step $d$ = 64]{
\includegraphics[width=0.21\textwidth]{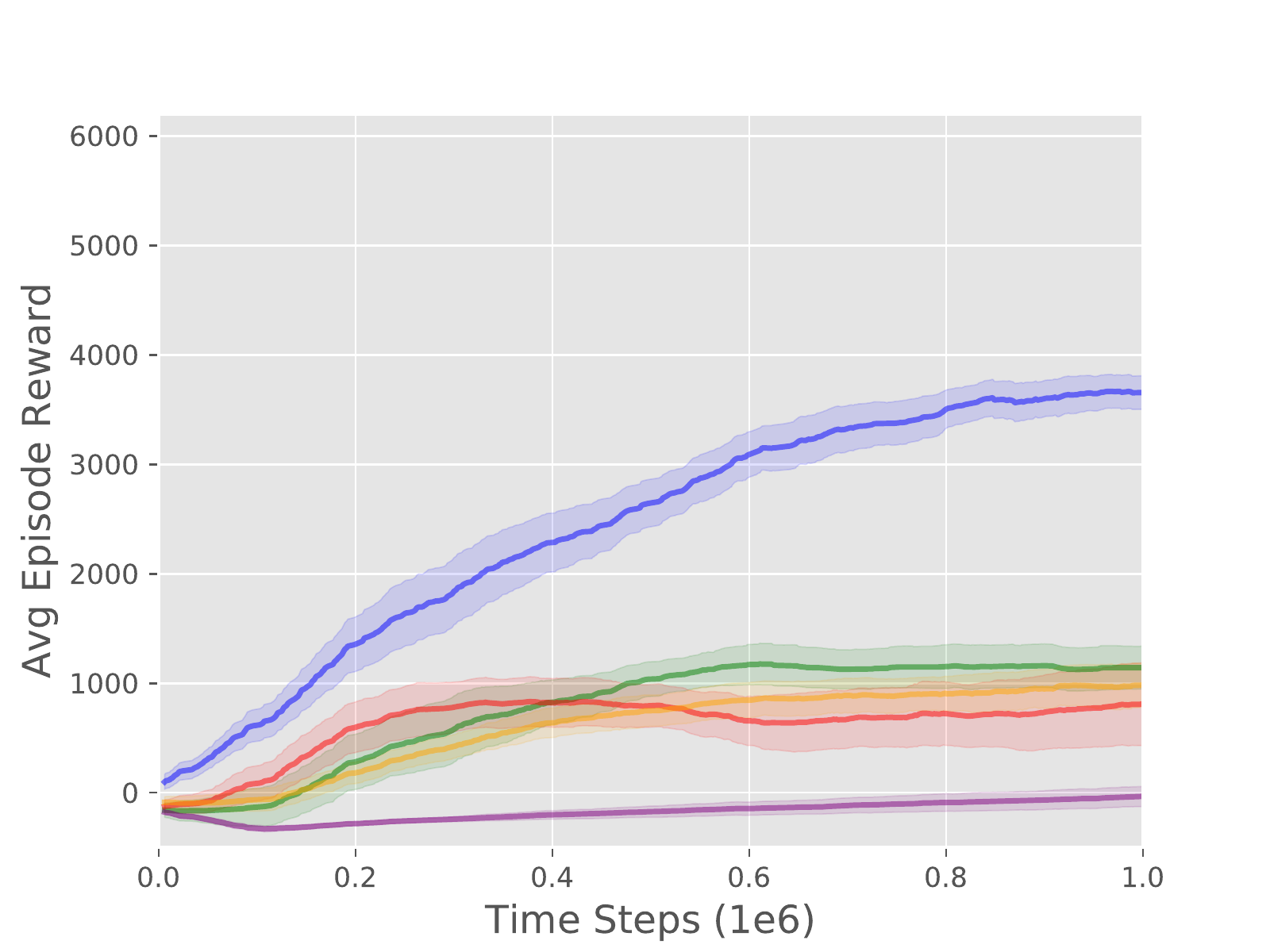}
}
\hspace{-0.6cm}
\subfigure[delay step $d$ = 128]{
\includegraphics[width=0.21\textwidth]{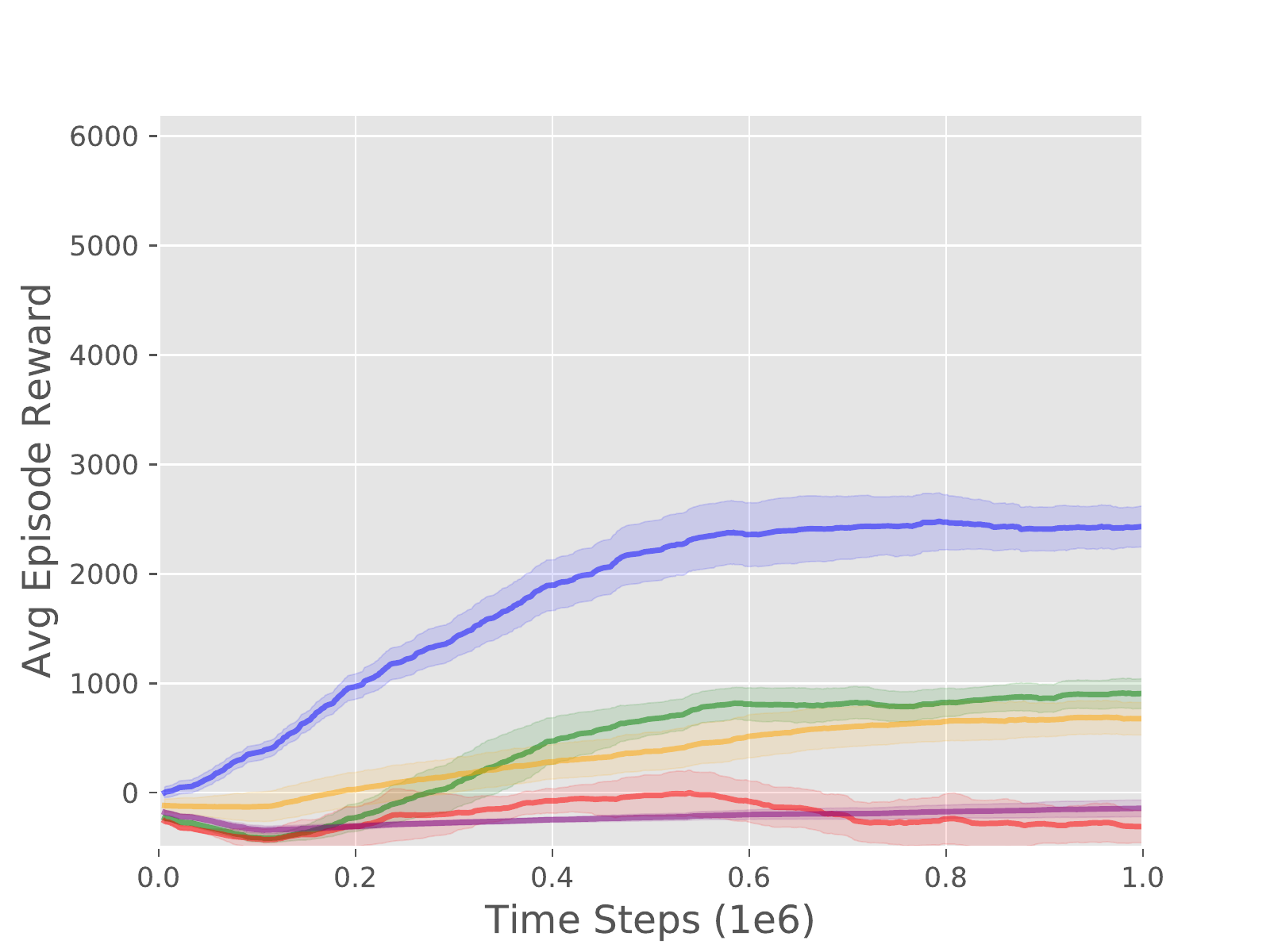}
}

\caption{Learning curves of algorithms in HalfCheetah-v1 under \emph{the first (1st) delayed reward setting}. 
Different delay steps are listed from left to right.
The shaded region denotes half a standard deviation of average evaluation over 5 trials. 
Results are smoothed over recent 100 episodes.}
\label{figure:ds1-HalfCheetah}
\end{figure}

\begin{figure}
\centering
\hspace{-0.1cm}
\subfigure[delay step $d$ = 0]{
\includegraphics[width=0.21\textwidth]{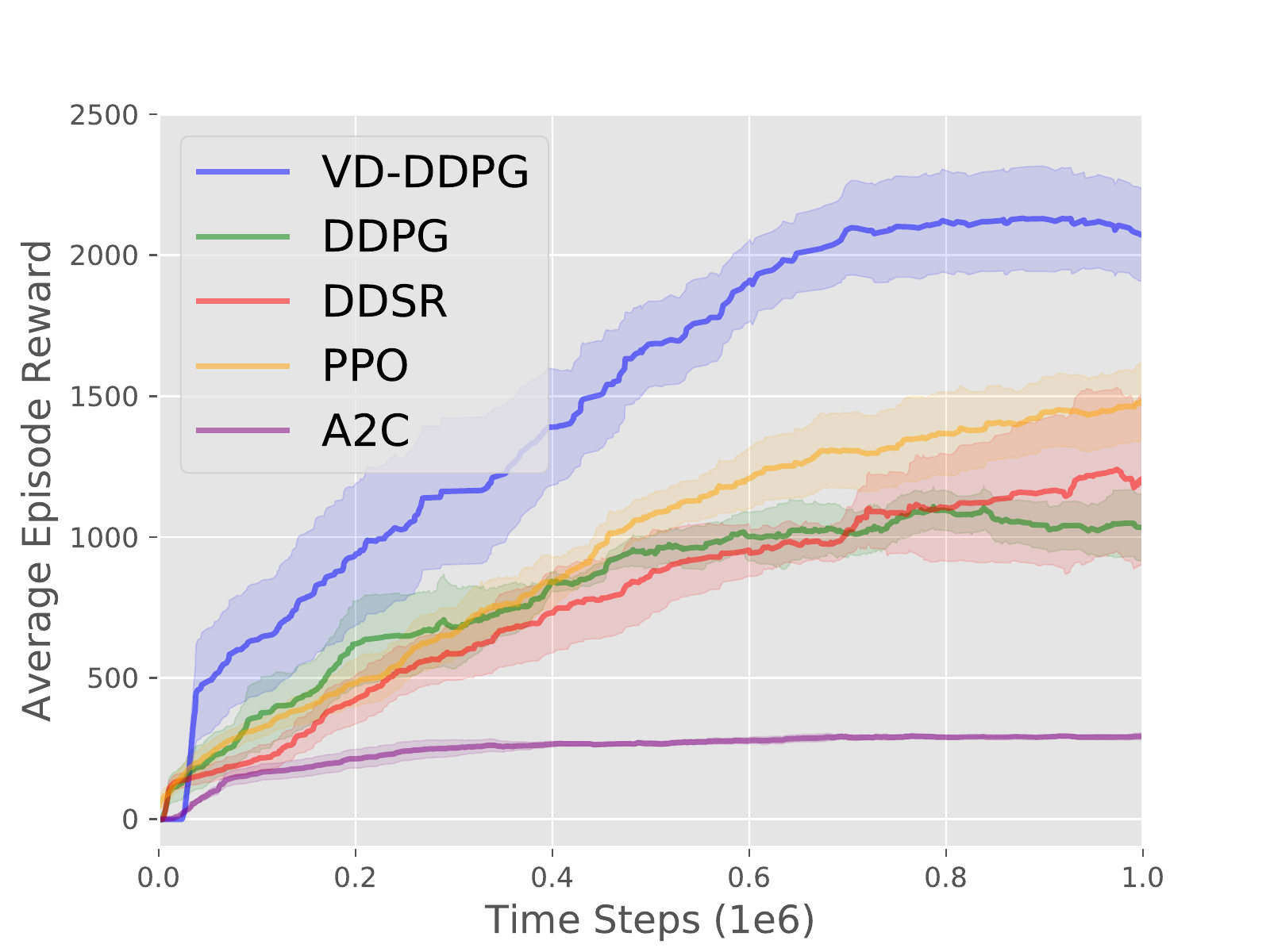}
}
\hspace{-0.6cm}
\subfigure[delay step $d$ = 16]{
\includegraphics[width=0.21\textwidth]{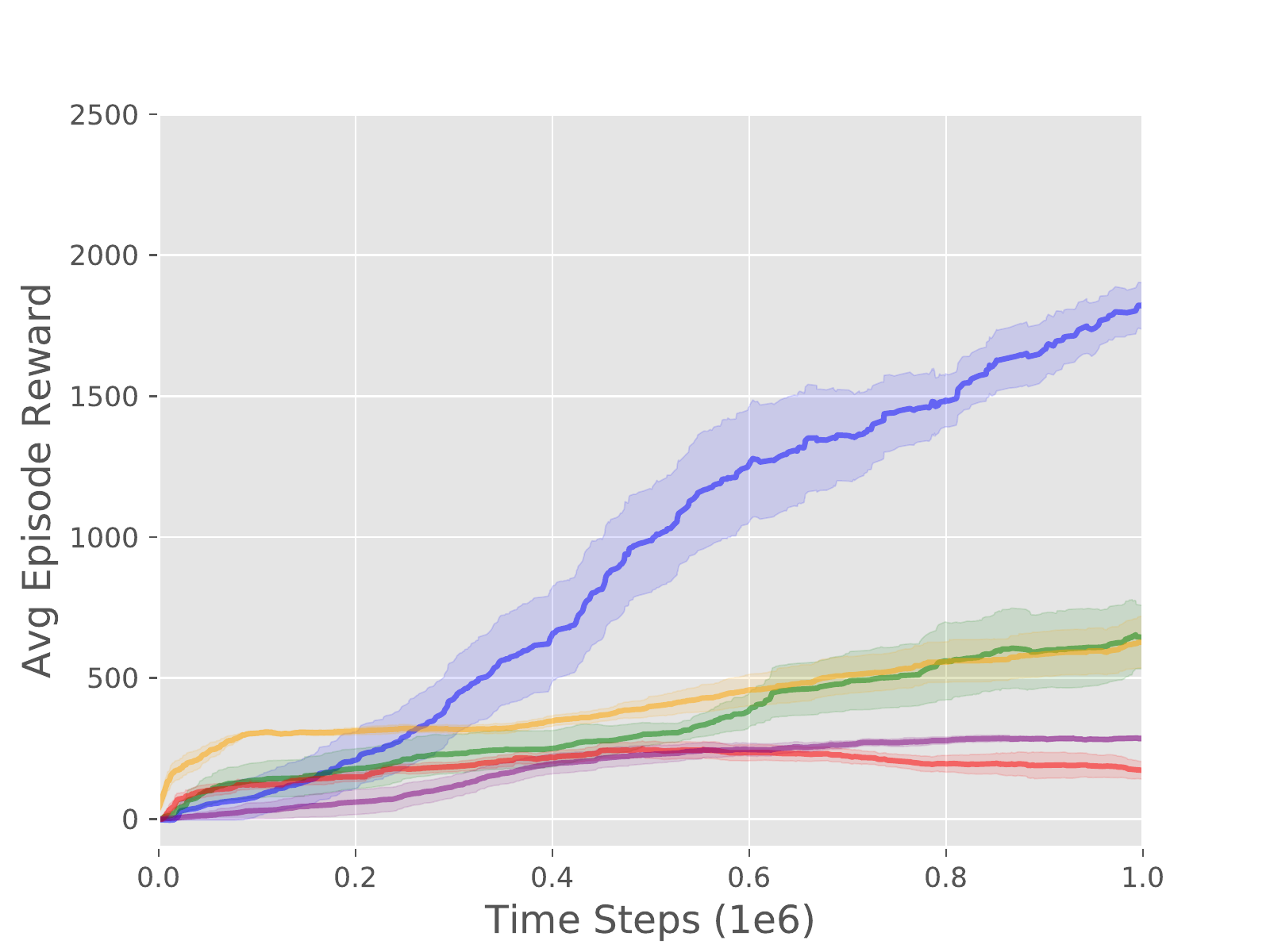}
}
\hspace{-0.6cm}
\subfigure[delay step $d$ = 32]{
\includegraphics[width=0.21\textwidth]{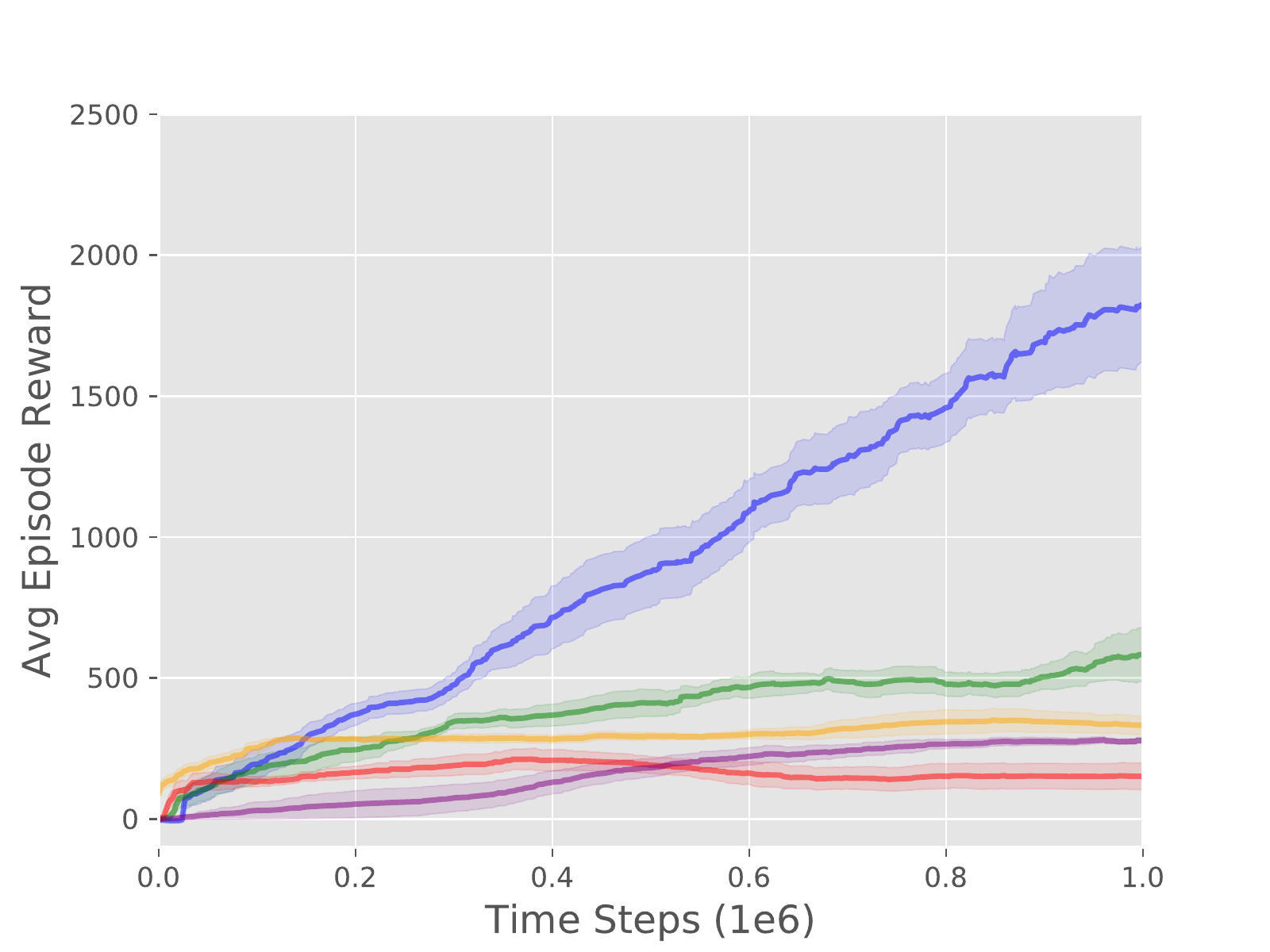}
}
\hspace{-0.6cm}
\subfigure[delay step $d$ = 64]{
\includegraphics[width=0.21\textwidth]{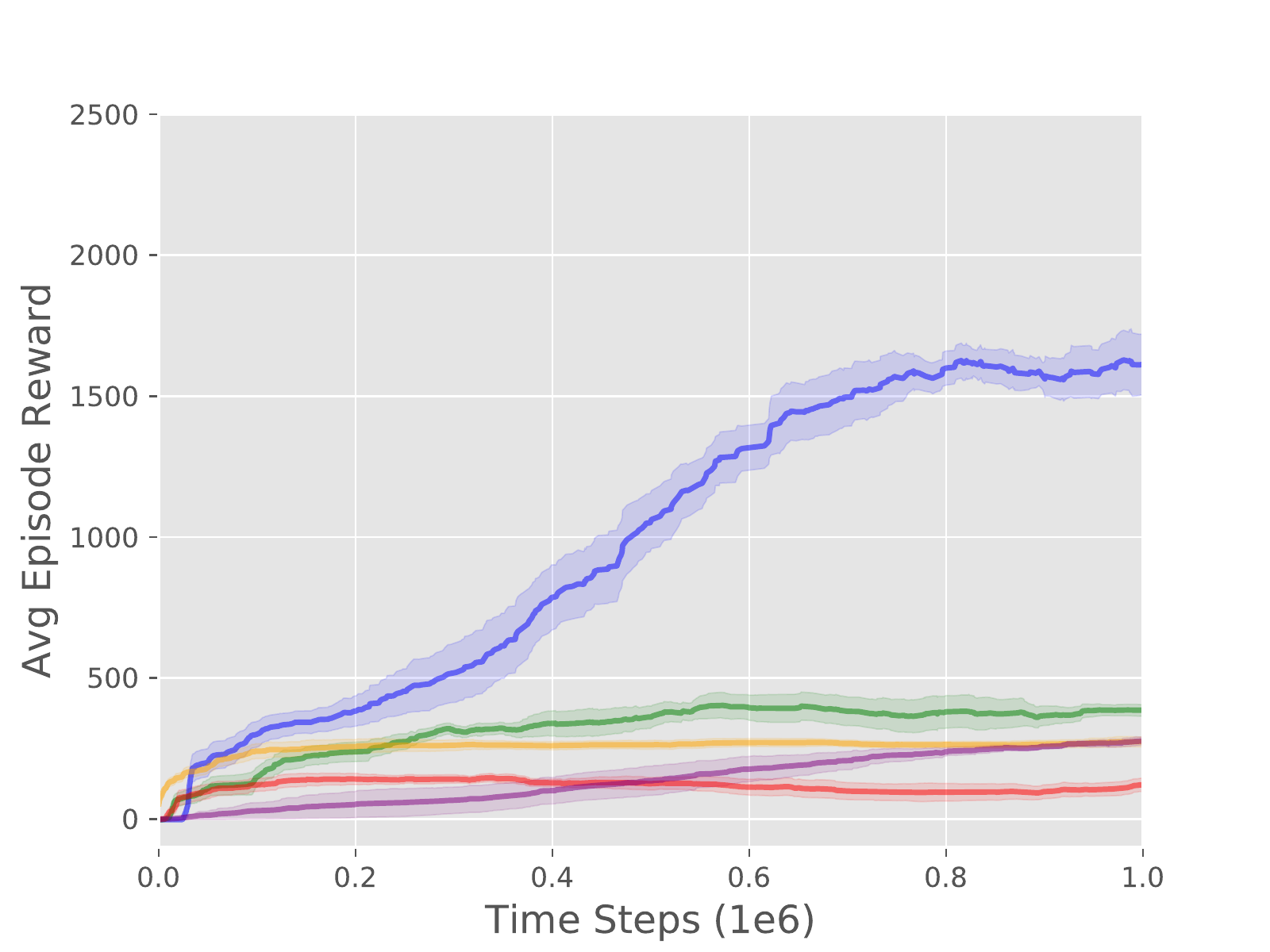}
}
\hspace{-0.6cm}
\subfigure[delay step $d$ = 128]{
\includegraphics[width=0.21\textwidth]{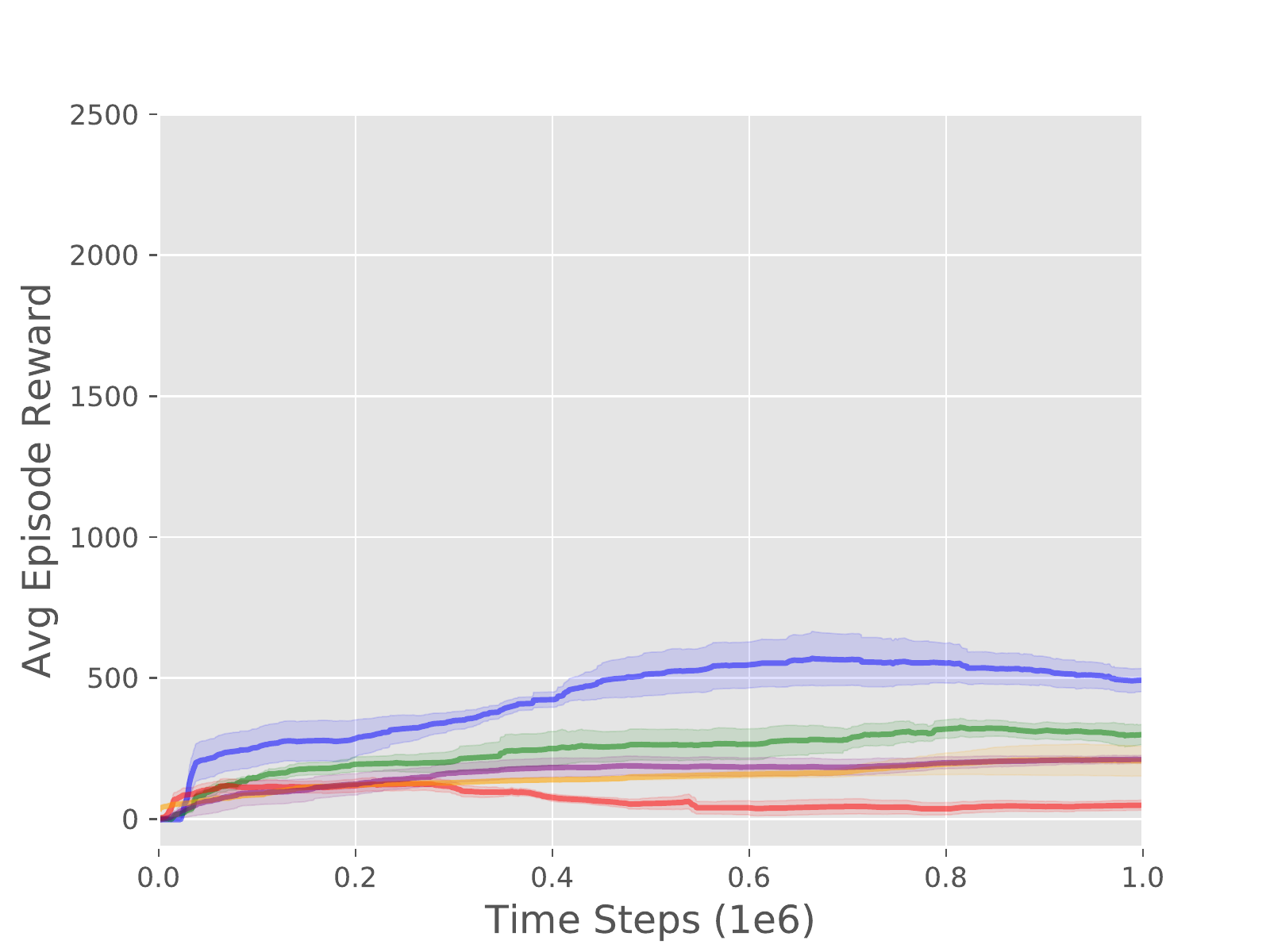}
}

\caption{Learning curves of algorithms in Walker2d-v1 under \emph{the first (1st) delayed reward setting}. 
Different delay steps are listed from left to right.
The shaded region denotes half a standard deviation of average evaluation over 5 trials. 
Results are smoothed over recent 100 episodes.}
\label{figure:ds1-Walker2d}
\end{figure}

\end{document}